\documentclass[11pt]{article}

\usepackage{multirow,booktabs,threeparttable,color}
\usepackage{setspace,graphicx,epstopdf,amsmath,amsfonts,amssymb,amsthm,subcaption}
\usepackage{verbatim}
\usepackage{marginnote,datetime,rotating,fancyvrb}
\usepackage{hyperref}
\usepackage{adjustbox}
\usepackage{multicol}
\usepackage{siunitx}
\usepackage{float}
\usepackage{natbib}
\usepackage[shortlabels]{enumitem}
\usepackage{amssymb}
\usepackage{amsmath}
\usepackage{bm}
\usepackage{bbm}
\usepackage{float}
\usepackage{fancyhdr}
\usepackage{booktabs}
\usepackage{graphicx,rotating}
\usepackage{color}
\usepackage{setspace}
\usepackage{enumitem}
\usepackage{pdfpages}

\usepackage{caption}
\usepackage{subcaption}

\usepackage[margin=1in]{geometry}%

\usepackage[verbose]{placeins}

\usepackage[normalem]{ulem}
\useunder{\uline}{\ul}{}

\usepackage{indentfirst} 
\usepackage{endnotes}    
\usepackage{jf}          
\usepackage[labelfont=bf,labelsep=period]{caption}   
\newcommand{\R}{\mathbb R}

\usepackage{ragged2e}
\usepackage{booktabs, ltablex, makecell, tabularx}

\newcolumntype{R}{>{\RaggedRight}X}

\newcommand{\E}{\mathbbm E}
\newcommand{\sfunc}{\bm{\theta}}
\newcommand{\Sspace}{\bm{\Theta}}
\newcommand{\s}{\theta}
\newcommand{\wfunc}{\bm{w^{\epsilon}}}
\newcommand{\w}{w^{\epsilon}}
\newcommand{\wret}{w^{R}}
\newcommand{\W}{\bm{W}}

\newcommand{\ffn}{\bm{g^{\text{FFN}}}}

\newcommand{\ds}{D_{\text{size}}}

\newcommand{\proj}{\Phi}

\newcommand*{\thisdraft}{This draft: September 25, 2022} 
\newcommand*{\firstdraft}{First draft: March 15, 2019}  

\usepackage[margin=10pt,labelfont=bf]{caption}
\newcommand\tcaptab[1]{\captionsetup{position=top, font=normalsize, labelfont=bf, textfont=normalfont, justification=centering, margin=0mm, aboveskip=1mm, belowskip=0mm, labelsep=colon, singlelinecheck=false}\caption{#1}}
\newcommand\bnotetab[1]{\captionsetup{position=bottom, font=scriptsize,  textfont=normalfont, margin=1mm, skip=2mm, justification=justified, singlelinecheck=false}\caption*{#1}}
\newcommand\tcapfig[1]{\captionsetup{position=top, font=normalsize, labelfont=bf, textfont=normalfont, justification=centering, margin=0mm, aboveskip=2mm, belowskip=0mm, labelsep=colon, singlelinecheck=false}\caption{#1}}
\newcommand\bnotefig[1]{\captionsetup{position=bottom, font=scriptsize,  textfont=normalfont, margin=1mm, skip=2mm, justification=justified, singlelinecheck=false}\caption*{#1}}

\setlist{noitemsep} 

\begin{document}

\title{Deep Learning Statistical Arbitrage\thanks{\scriptsize We thank Robert Anderson, Jose Blanchet, Marcelo Fernandes, Kay Giesecke, Lisa Goldberg, Amit Goyal (discussant), Bryan Kelly, Robert Korajczyk, Martin Lettau, Sophia Li, Marcelo Medeiros, Scott Murray, George Papanicolaou, Guofu Zhou (discussant) and seminar and conference participants at Stanford, UC Berkeley, Rutgers University, GSU CEAR Finance Conference, NBER-NSF Time-Series Conference, AI and Big Data in Finance Research Seminar, AI in Fintech Forum, World Congress of the Bachelier Finance Society, Society of Financial Econometrics Annual Conference, the Econometric Research in Finance Workshop, the Meeting of the Brazilian Finance Society, World Online Seminars on Machine Learning in Finance, the Machine Learning and Quantitative Finance Workshop at Oxford, Annual Bloomberg-Columbia Machine Learning in Finance Workshop, NVIDIA AI Webinar, Vanguard Academic Seminar, INFORMS and the Western Conference on Mathematical Finance for helpful comments. We thank MSCI for generous research support.}}
\date{\thisdraft \\ \firstdraft}
\author{Jorge Guijarro-Ordonez\thanks{\scriptsize Stanford University, Department of Mathematics, Email: jguiord@stanford.edu.}
\and
Markus Pelger\thanks{\scriptsize Stanford University, Department of Management Science \& Engineering, Email: mpelger@stanford.edu.}
\and Greg Zanotti\thanks{\scriptsize Stanford University, Department of Management Science \& Engineering, Email: gzanotti@stanford.edu.}}

\onehalfspacing

\begin{titlepage}
\maketitle
\thispagestyle{empty}
\begin{abstract}
\vspace{1cm}


Statistical arbitrage exploits temporal price differences between similar assets. We develop a unifying conceptual framework for statistical arbitrage and a novel data driven solution. First, we construct arbitrage portfolios of similar assets as residual portfolios from conditional latent asset pricing factors. Second, we extract their time series signals with a powerful machine-learning time-series solution, a convolutional transformer. Lastly, we use these signals to form an optimal trading policy, that maximizes risk-adjusted returns under constraints. Our comprehensive empirical study on daily US equities shows a high compensation for arbitrageurs to enforce the law of one price. Our arbitrage strategies obtain consistently high out-of-sample mean returns and Sharpe ratios, and substantially outperform all benchmark approaches.

\vspace{1cm}

\end{abstract}
\end{titlepage}
\section{ Introduction}


Statistical arbitrage is one of the pillars of quantitative trading, and has long been used by hedge funds and investment banks. The term statistical arbitrage encompasses a wide variety of investment strategies, which identify and exploit temporal price differences between similar assets using statistical methods. Conceptually, they are based on relative trades between a stock and a mimicking portfolio. The mimicking portfolio is constructed to be ``similar'' to the target stock, usually based on historical co-movements in the price time-series or similar firm characteristics. When the spread between the prices of the two comparison assets widens, the arbitrageur sells the winner and buys the loser. If their prices move back together, the arbitrageur will profit. While Wall Street has developed a plethora of proprietary tools for sophisticated arbitrage trading, there is still a lack of understanding of how much arbitrage opportunity is actually left in financial markets.  In this paper we answer the two key questions around statistical arbitrage: What are the important elements of a successful arbitrage strategy and how much realistic arbitrage is in financial markets?

Every statistical arbitrage strategy needs to solve the following three fundamental problems: Given a large universe of assets, what are long-short portfolios of similar assets? Given these portfolios, what are time series signals that indicate the presence of temporary price deviations? Last, but not least, given these signals, how should an arbitrageur trade them to optimize a trading objective while taking into account possible constraints and market frictions? Each of these three questions poses substantial challenges, that prior work has only partly addressed. First, it is a hard problem to find long-short portfolios for all stocks as it is a priori unknown what constitutes ``similarity''. This problem requires considering all the big data available for a large number of assets and times, including not just conventional return data but also exogenous information like asset characteristics. Second, extracting the right signals requires detecting flexibly all the relevant patterns in the noisy, complex, low-sample-size time series of the portfolio prices. Last but not least, optimal trading rules on a multitude of signals and assets are complicated and depend on the trading objective. All of these challenges fundamentally require flexible estimation tools that can deal with many variables. It is a natural idea to use machine learning techniques like deep neural networks to deal with the high dimensionality and complex functional dependencies of the problem. However, our problem is different from the usual prediction task, where machine learning tools excel. We show how to optimally design a machine learning solution to our problem that leverages the economic structure and objective.

In this paper, we propose a unifying conceptual framework that generalizes common approaches to statistical arbitrage. Statistical arbitrage can be decomposed into three fundamental elements: (1) arbitrage portfolio generation, (2) arbitrage signal extraction and (3) the arbitrage allocation decision given the signal. By decomposing different methods into their arbitrage portfolio, signal and allocation element, we can compare different methods and study which components are the most relevant for successful trading. For each step we develop a novel machine learning implementation, which we compare with conventional methods. As a result, we construct a new deep learning statistical arbitrage approach. Our new approach constructs arbitrage portfolios with a conditional latent factor model, extracts the signals with the currently most successful machine learning time-series method and maps them into a trading allocation with a flexible neural network. These components are integrated and optimized over a global economic objective, which maximizes the risk-adjusted return under constraints. Empirically, our general model outperforms out-of-sample the leading benchmark approaches and provides a clear insight into the structure of statistical arbitrage.

To construct arbitrage portfolios, we introduce the economically motivated asset pricing perspective to create them as residuals relative to asset pricing models. This perspective allows us to take advantage of the recent developments in asset pricing and to also include a large set of firm characteristics in the construction of the arbitrage portfolios. We use fundamental risk factors and conditional and unconditional statistical factors for our asset pricing models. Similarity between assets is captured by similar exposure to those factors.  Hence, arbitrage portfolios are trades relative to mimicking stock portfolios, which are well-diversified portfolios with the same exposure to risk factors. In the case of statistical principal component factors, the mimicking portfolios correspond to assets with the highest correlation with the target stocks. Arbitrage Pricing Theory provides a justification for mean reversion patterns. It implies that, with an appropriate model, the corresponding factor portfolios represent the ``fair price'' of each of the assets. Therefore, the residual portfolios relative to the asset pricing factors capture the temporary deviations from the fair price of each of the assets and should only temporally deviate from their long-term mean. Importantly, the residuals are tradeable portfolios, which are only weakly cross-sectionally correlated, and close to orthogonal to firm characteristics and systematic factors. These properties allow us to extract a stationary time-series model for the signal.

To detect time series patterns and signals in the residual portfolios, we introduce a filter perspective and estimate them with a flexible data-driven filter based on convolutional networks combined with transformers. In this way, we do not prescribe a potentially misspecified function to extract the time series structure, for example, by estimating the parameters of a given parametric time-series model, or the coefficients of a decomposition into given basis functions, as in conventional methods. Instead, we directly learn in a data-driven way what the optimal pattern extraction function is for our trading objective. The convolutional transformer is the ideal method for this purpose. Convolutional neural networks are the state-of-the-art AI method for pattern recognition, in particular in computer vision. In our case they identify the local patterns in the data and may be thought as a nonlinear and learnable generalization of conventional kernel-based data filters. Transformer networks are the most successful AI model for time series in natural language processing. In our model, they combine the local patterns to global time-series patterns. Their combination results in a data-driven flexible time-series filter that can essentially extract any complex time-series signal, while providing an interpretable model.

To find the optimal trading allocation, we propose neural networks to map the arbitrage signals into a complex trading allocation. This generalizes conventional parametric rules, for example fixed rules based on thresholds, which are only valid under strong model assumptions and a small signal dimension. Importantly, these components are integrated and optimized over a global economic objective, which maximizes the risk-adjusted return under constraints. This allows our model to learn the optimal signals and allocation for the actual trading objective, which is different from a prediction objective. The trading objective can maximize the Sharpe ratio or expected return subject to a risk penalty, while taking into account constraints important to real investment managers, such as restricting turnover, leverage, or proportion of short trades.

Our comprehensive empirical out-of-sample analysis is based on the daily returns of roughly the 550 largest and most liquid stocks in the U.S. from 1998 to 2016. We estimate the out-of-sample residuals on a rolling window relative to the empirically most important factor models. These are observed fundamental factors, for example the Fama-French 5 factors and price trend factors, locally estimated latent factors based on principal component analysis (PCA) or locally estimated conditional latent factors that include the information in 46 firm-specific characteristics and are based on the Instrumented PCA (IPCA) of \cite{IPCA}. We extract the trading signal with one of the most successful parametric models, based on the mean-reverting Ornstein-Uhlenbeck process, a frequency decomposition of the time-series with a Fourier transformation and our novel convolutional network with transformer. Finally, we compare the trading allocations based on parametric or nonparametric rules estimated with different risk-adjusted trading objectives.

Our empirical main findings are five-fold. First, our deep learning statistical arbitrage model substantially outperforms all benchmark approaches out-of-sample. In fact, our model can achieve an impressive annual Sharpe ratio larger than four. While respecting short-selling constraints we can obtain annual out-of-sample mean returns of 20\%. This performance is four times better than one of the best parametric arbitrage models, and twice as good as an alternative deep learning model without the convolutional transformer filter. These results are particularly impressive as we only trade the largest and most liquid stocks. Hence, our model establishes a new standard for arbitrage trading. 

Second, the performance of our deep learning model suggests that there is a substantial amount of short-term arbitrage in financial markets. The profitability of our strategies is orthogonal to market movements and conventional risk factors including momentum and reversal factors and does not constitute a risk-premium. Our strategy performs consistently well over the full time horizon. The model is extremely robust to the choice of tuning parameters, and the period when it is estimated. Importantly, our arbitrage strategy remains profitable in the presence of realistic transaction and holdings costs. Arbitrage signals are persistent over short time horizons. While arbitrageurs correct most mispricing over the horizon of a month, around half of the Sharpe ratio can persist for a holding period of one week. Assessing the amount of arbitrage in financial markets with unconditional pricing errors relative to factor models or with parametric statistical arbitrage models, severely underestimates the amount of statistical arbitrage.

Third, the trading signal extraction is the most challenging and separating element among different arbitrage models. Surprisingly, the choice of asset pricing factors has only a minor effect on the overall performance. Residuals relative to the five Fama-French factors and five locally estimated principal component factors perform very well with out-of-sample Sharpe ratios above 3.2 for our deep learning model. Five conditional IPCA factors increase the out-of-sample Sharpe ratio to 4.2, which suggests that asset characteristics provide additional useful information. Increasing the number of risk factors beyond five has only a marginal effect. Similarly, the other benchmark models are robust to the choice of factor model as long as it contains sufficiently many factors. The distinguishing element is the time-series model to extract the arbitrage signal. The convolutional transformer doubles the performance relative to an identical deep learning model with a pre-specified frequency filter. Importantly, we highlight that time-series modeling requires a time-series machine learning approach, which takes temporal dependency into account. An off-the-shelf nonparametric machine learning method like conventional neural networks, that estimates an arbitrage allocation directly from residuals, performs substantially worse. 

Fourth, successful arbitrage trading is based on local asymmetric trend and reversion patterns. Our convolutional transformer framework provides an interpretable representation of the underlying patterns, based on local basic patterns and global ``dependency factors''. The building blocks of arbitrage trading are smooth trend and reversion patterns. The arbitrage trading is short-term and the last 30 trading days seem to capture the relevant information. Interestingly, the direction of policies is asymmetric. The model reacts quickly on downturn movements, but more cautiously on uptrends. More specifically, the ``dependency factors'' which are the most active in downturn movements focus only on the most recent 10 days, while those for upward movements focus on the first 20 days in a 30-day window.  

Fifth, time-series-based trading patterns should be extracted from residuals and not directly from returns. For an appropriate factor model, the residuals are only weakly correlated and close to stationary in both, the time and cross-sectional dimension. Hence, it is meaningful to extract a uniform trading pattern, that is based only on the past time-series information, from the residuals. In contrast stock returns are dominated by a few factors, which severely limits the actual independent time-series information, and are strongly heterogenous due to their variation in firm characteristics. While the level of stock returns is extremely hard to predict, even with flexible machine learning methods, residuals capture relative movements and remove the level component. These properties make residuals analyzable from a purely time-series based perspective and, unlike the existing literature, they allow us to incorporate alternative data into the portfolio construction process. This also highlights a fundamental difference with most of the existing financial machine learning literature: We do not use characteristics to get features for prediction, but rather to obtain the data orthogonal to these features.

\subsection*{Related Literature}

Our paper builds on the classical statistical arbitrage literature, in which the three main problems of portfolio generation, pattern extraction, and allocation decision have traditionally been considered independently. Classical statistical methods of generating arbitrage portfolios have mostly focused on obtaining multiple pairs or small portfolios of assets, using techniques like the distance method of \cite{Gatev}, the cointegration approach of \cite{Vidyamurthy}, or copulas as in \cite{Copula}. In contrast, more general methods that exploit large panels of stock returns include the use of PCA factor models, as in \cite{avelee} and its extension in \cite{papayeo}, and the maximization of mean-reversion and sparsity statistics as in \cite{dAspremont}. We include the model of \cite{papayeo} as the parametric benchmark model in our study as it has one of the best empirical performances among the class of parametric models. Our paper paper contributes to this literature by introducing a general asset pricing perspective to obtain the arbitrage portfolios as residuals. This allows us to take advantage of conditional asset pricing models, that include time-varying firm characteristics in addition to the return time-series, and provides a more disciplined, economically motivated approach. The signal extraction step for these models assumes parametric time series models for the arbitrage portfolios, whereas the allocations are often decided from the estimated parameters by using stochastic control methods or given threshold rules and one-period optimizations. Some representative papers of the first approach include \cite{jurek-yang}, \cite{orig-stoch-opt}, \cite{Cartea2016}, \cite{Tourin2016} and \cite{leung-stoploss}, whereas the second one is illustrated by \cite{Elliott} and \cite{ papayeo}. Both approaches are special cases of our more general framework. \cite{Mulvey} and \cite{StatArbRl} are examples of including machine learning elements within the parametric statistical arbitrage framework, by either solving a stochastic control problem with neural networks or estimating a time-varying threshold rule with reinforcement learning. 

Our paper is complementary to the emerging literature that uses machine learning methods for asset pricing. While the asset pricing literature aims to explain the risk premia of assets, our focus is on the residual component which is not explained by the asset pricing models. \cite{DeepMarkus}, \cite{bryzgalova2019} and \cite{kozak2017} estimate the stochastic discount factor (SDF), which explains the risk premia of assets, with deep neural networks, decision trees or elastic net regularization. These papers employ advanced statistical methods to solve a conditional method of moment problem in the presence of many variables. The workhorse models in equity asset pricing are based on linear factor models exemplified by \cite{fama1993,fama2015}. Recently, new methods have been developed to extract statistical asset pricing factors from large panels with various versions of principal component analysis (PCA). The Risk-Premium PCA in \cite{LettauPelger2018Theory,lettaupelger2018} includes a pricing error penalty to detect weak factors that explain the cross-section of returns. The high-frequency PCA in \cite{pelger2019} uses high-frequency data to estimate local time-varying latent risk factors and the Instrumented PCA (IPCA) of \cite{IPCA} estimates conditional latent factors by allowing the loadings to be functions of time-varying asset characteristics. \cite{kelly2019} generalize IPCA to allow the loadings to be nonlinear functions of characteristics. \cite{giglio2021} use PCA factors to account for missing priced factors in risk premia estimation. \cite{he_et_al2022} highlight the important insight that the time-series of residuals can be informative for trading. They test asset pricing models by studying the profitability of trading residuals based on high-minus-low sorting strategies of prior residual returns.

Our paper is related to the growing literature on return prediction with machine learning methods, which has shown the benefits of regularized flexible methods. In their pioneering work \cite{gu2018} conduct a comparison of machine learning methods for predicting the panel of individual U.S. stock returns based on the asset-specific characteristics and economic conditions in the previous period. \cite{Freyberger} use different methods for predicting stock returns. In a similar spirit, \cite{bianchi2019} predict bond returns, Li and Rossi (2020) predict mutual fund returns, and \cite{Turan_et_al2022} predict option returns. A related stream extends cross-sectional machine learning prediction to higher moments and residuals, which are used for investment decisions. \cite{LiTang2022} use machine learning to predict risk measures and estimate conditional volatilities. \cite{kaniel_et_al2022} and \cite{DeMiguel_et_al2022} predict the skill of mutual fund managers based on residuals from benchmark asset pricing factors. The return prediction literature is fundamentally estimating the risk premia of assets, while our focus is on understanding and exploiting the temporal deviations thereof. This different goal is reflected in the different methods that are needed. These return predictions estimate a nonparametric cross-sectional model between current returns and large set of covariates from the last period, but do not estimate a time-series model. In contrast, the important challenge that we solve is to extract a complex time-series pattern. 

A related stream of the return prediction literature forecasts returns using past returns, generally followed by some long-short investment policy based on the prediction. For example, \cite{Krauss} use various machine learning methods for this type of prediction.\footnote{Similar studies include \cite{krauss2019}, \cite{StatArbCNN}, \cite{Huck}, and \cite{Dunis}.} However, they use general nonparametric function estimates, which are not specifically designed for time-series data. \cite{deepTimeSeries} show that it is important for machine learning solutions to explicitly account for temporal dependence when they are applied to time-series data. \cite{murray2021} and \cite{kelly2022} build on this insight and use machine learning to learn price trends for return forecasting. Forecasting returns and building a long-short portfolio based on the prediction is different from statistical arbitrage trading as it combines a risk premium and potential arbitrage component. It is not based on temporary price differences and also in general not orthogonal to common risk factors and market movements. In this paper we highlight the challenge of inferring complex time-series information and argue that using returns directly as an input to a time-series machine learning method, can be suboptimal as returns are dominated by a few factor time-series and are heterogeneous due to cross-sectionally and time-varying characteristics. In contrast, appropriate residuals are close to uncorrelated and locally cross-sectionally stationary. Hence, appropriate residuals allow the extraction of a complex time-series pattern.

Naturally, our work overlaps with the literature on using machine learning tools for investment. The SDF estimated by asset pricing models, like in \cite{DeepMarkus} and \cite{bryzgalova2019}, directly maps into a conditionally mean-variance efficient portfolio and hence an attractive investment opportunity. However, by construction this investment portfolio is not orthogonal but fully exposed to systematic risk, which is exactly the opposite for an arbitrage portfolio. Prediction approaches also imply investment strategies, typically long-short portfolios based on the prediction signal. However, estimating a signal with a prediction objective, is not necessarily providing an optimal signal for investment. \cite{bryzgalova2019} and \cite{DeepMarkus} illustrate that machine learning models that use a trading objective can result in a substantially more profitable investment than models that estimate a signal with a prediction objective, while using the same information as input and having the same flexibility. This is also confirmed in \cite{Cong}, who use an investment objective and reinforcement learning to construct machine learning investment portfolios. Our paper contributes to this literature by estimating investment strategies, that are orthogonal to systematic risk and are based on a trading objective with constraints.

Finally, our approach is also informed by the recent deep learning for time series literature. The transformer method was first introduced in the groundbreaking paper by \cite{attention}. We are the first to bring this idea into the context of statistical arbitrage and adopt it to the economic problem.


\section{Model}\label{sec:theory}

The fundamental problem of statistical arbitrage consists of three elements: (1) The identification of similar assets to generate arbitrage portfolios, (2) the extraction of time-series signals for the temporary deviations of the similarity between assets and (3) a trading policy in the arbitrage portfolios based on the time-series signals. We discuss each element separately.

\subsection{Arbitrage portfolios}\label{sec:arbitrageportfolios}

We consider a panel of excess returns $R_{n,t}$, that is the return minus risk free rate of stock $n=1,...,N_t$ at time $t=1,...,T$. The number of available assets at time $t$ can be time-varying. The excess return vector of all assets at time $t$ is denoted as $R_t= \begin{pmatrix} R_{1,t} & \cdots & R_{N_t,t} \end{pmatrix}^{\top}$.

We use a general asset pricing model to identify similar assets. In this context, similarity is defined as the same exposure to systematic risk, which implies that assets with the same risk exposure should have the same fundamental value. We assume that asset returns can be modeled by a conditional factor model:
\begin{align*}
R_{n,t}= \beta_{n,t-1}^{\top} F_t + \epsilon_{n,t}.
\end{align*}
The $K$ factors $F \in \mathbbm R^{T \times K}$ capture the systematic risk, while the risk loadings $\beta_{t-1} \in \mathbbm R^{N_t \times K}$ can be general functions of the information set at time $t-1$ and hence can be time-varying. This general formulation includes the empirically most successful factor models. In our empirical analysis we will include observed traded factors, e.g. the Fama-French 5 factor model, latent factors based on the principal components analysis (PCA) of stock returns and conditional latent factors estimated with Instrumented Principal Component Analysis (IPCA).

Without loss of generality, we can treat the factors as excess returns of traded assets. Either the factors are traded, for example a market factor, in which case we include them in the returns $R_t$. Otherwise, we can generate factor mimicking portfolios by projecting them on the asset space, as for example with latent factors:
\begin{align*}
F_t= {w_{t-1}^F}^{\top} R_t.
\end{align*}
We define {\it arbitrage portfolios} as residual portfolios $\epsilon_{n,t}=R_{n,t} - \beta_{n,t-1}^{\top} F_t$. As factors are traded assets, the arbitrage portfolios are themselves traded portfolios.
Hence, the vector of residual portfolios equals 
\begin{align}\label{eq:transitionMatrix}
\epsilon_t &= R_t - \beta_{t-1} {w_{t-1}^F}^{\top} R_t = \underbrace{\left(I_{N_t} - \beta_{t-1} {w_{t-1}^F}^{\top}  \right )}_{ \proj_{t-1}} R_{t} = \proj_{t-1} R_t.
\end{align}

Arbitrage portfolios are trades relative to mimicking stock portfolios. The factor component in the residuals corresponds to a well-diversified mimicking portfolio, that is is ``close'' to the specific stock in the relative trade. More specifically, in the case of PCA factors, we construct well-diversified portfolios of stocks that have the highest correlation with each target stock in the relative trades. This is conceptually the same idea as using some form of clustering algorithm to construct a portfolio of highly correlated stocks.\footnote{In fact, under appropriate assumptions a clustering problem can be solved by a latent factor estimation.} Hence, residuals of PCA factors construct relative trades based on correlations in past return time-series. In the case of IPCA factors, we intuitively construct for each stock a well-diversified mimicking portfolio that is as similar as possible in terms of the underlying firm characteristics. Hence, the mimicking portfolio represents an asset with very similar firm fundamentals. Lastly, for Fama-French factors, we construct mimicking portfolios with the same exposure to those fundamental factors. In all cases, the choice of factor model implies a notation of similarity. By construction the mimicking portfolio and the target stock have identical loadings to the selected factors.

Arbitrage trading relative to mimicking portfolios takes advantage of all assets in the market. In practice and modern statistical arbitrage theory, it has replaced the more restrictive idea of pairs trading, which considers relative trades between only two stocks. On an intuitive level, the mimicking portfolio represents the reference asset in the relative trade. It can be constructed even if there does not exist an individual stock that is highly correlated with a target stock.

Our arbitrage portfolios based on residuals also allow us to benefit from an asset pricing perspective. Arbitrage portfolios are projections on the return space that annihilate systematic asset risk. For an appropriate asset pricing model, the residual portfolios should not earn a risk premium. This is the fundamental assumption behind any arbitrage argument. As deviations from a mean of zero have to be temporary, arbitrage trading bets on the mean revision of the residuals. In particular, for an appropriate factor model the residuals are expected to have the following properties: First, the unconditional mean of the arbitrage portfolios is zero, $\E [\epsilon_{n,t}]=0$. Second, the arbitrage portfolios are only weakly cross-sectionally dependent. We confirm empirically that once we use sufficiently many statistical or fundamental factors the residuals are indeed close to uncorrelated and have an average mean close to zero. However, while the asset pricing perspective provides a theoretical argument for mean reversion in residuals, our approach is still valid even if we omit some relevant risk factors. The relative trades would then only exploit the temporal time-series deviations from the risk factors captured by the mimicking portfolios.

We denote by $\mathcal F_{t-1}$ the filtration generated by the returns $R_{t-1}$, which include the factors, and the information set that captures the risk exposure $\beta_{t-1}$, which is typically based on asset specific characteristics or past returns.

\subsection{Arbitrage signal}
The {\it arbitrage signal} is extracted from the time-series of the arbitrage portfolios. These time-series signals are the input for a trading policy. An example for an arbitrage signal would be a parametric model for mean reversion that is estimated for each arbitrage portfolio. The trading strategy for each arbitrage portfolio would depend on its speed of mean reversion and its deviation from the long run mean. More generally, the arbitrage signal is the estimation of a time-series model, which can be parametric or nonparametric. An important class of models are filtering approaches. Conceptually, time-series models are multivariate functional mappings between sequences, which take into account the temporal order of the elements and potentially complex dependencies between the elements of the input sequence. 

We apply the signal extraction to the time-series of the last $L$ lagged residuals, which we denote in vector notation as
\begin{align*}
\epsilon_{n,t-1}^L := \begin{pmatrix} \epsilon_{n,t-L} & \cdots & \epsilon_{n,t-1} \end{pmatrix}.
\end{align*}
The arbitrage signal function is a mapping $\sfunc \in \Sspace$ from $\mathbbm R^L$ to $\mathbbm R^p$, where $\Sspace$ defines an appropriate function space: 
\begin{align*}
\sfunc(\cdot) : \epsilon_{n,t-1}^L \rightarrow \s_{n,t-1}.
\end{align*}
The signals $\s_{n,t-1} \in \mathbbm R^p$ for the arbitrage portfolio $n$ at time $t$ only depend on the time-series of lagged returns $ \epsilon_{n,t-1}^L$. Note that the dimensionality of the signal can be the same as for the input sequence. Formally, the function $\sfunc$ is a mapping from the filtration $\mathcal F_{n,t-1}^{\epsilon,L}$ generated by $ \epsilon_{n,t-1}^L$ into the filtration $\mathcal F_{n,t-1}^{\theta}$ generated by $\s_{n,t-1}$ and $\mathcal F_{n,t-1}^{\theta} \subseteq \mathcal F_{n,t-1}^{\epsilon,L}$. We use the notation of evaluating functions elementwise, that is $\sfunc(\epsilon_{t-1}^L) = \begin{pmatrix} \s_{1,t-1} & \cdots & \s_{N_{t},t-1} \end{pmatrix} = \s_{t-1}  \in \mathbbm R^{N_t}$ with $\epsilon_{t-1}^L= \begin{pmatrix} \epsilon_{1,t-1} & \cdots & \epsilon_{N_{t},t-1} \end{pmatrix}$.

The arbitrage signal $\s_{n,t-1}$ is a sufficient statistic for the trading policy; that is, all relevant information for trading decisions is summarized in it. This also implies that two arbitrage portfolios with the same signal get the same weight in the trading strategy. More formally, this means that the arbitrage signal defines equivalence classes for the arbitrage portfolios. The most relevant signals summarize reversal patterns and their direction with a small number of parameters. A potential trading policy could be to hold long positions in residuals with a predicted upward movement and go short in residuals that are in a downward cycle. 

This problem formulation makes two implicit assumptions. First, the residual time-series follow a stationary distribution conditioned on its lagged returns. This is a very general framework that includes the most important models for financial time-series. Second, the first $L$ lagged returns are a sufficient statistic to obtain the arbitrage signal $\s_{n,t-1}$. This reflects the motivation that arbitrage is a temporary deviation of the fair price. The lookback window can be chosen to be arbitrarily large, but in practice it is limited by the availability of lagged returns.

\subsection{Arbitrage trading}\label{sec:arbitragetrading}

The trading policy assigns an investment weight to each arbitrage portfolio based on its signal. The allocation weight is the solution to an optimization problem, which models a general risk-return tradeoff and can also include trading frictions and  constraints. An important case are mean-variance efficient portfolios with transaction costs and short sale constraints.

An arbitrage allocation is a mapping from $\mathbbm R^p$ to $\mathbbm R$ in a function class $\W$, that assigns a weight $\w_{n,t-1}$ for the arbitrage portfolio $\epsilon_{n,t-1}$ in the investment strategy using only the arbitrage signal $\s_{n,t}$:
\begin{align*}
\wfunc: \s_{n,t-1} \rightarrow \w_{n,t-1}.
\end{align*} 
Given a concave utility function $U(\cdot)$, the allocation function is the solution to
\begin{align}
&\max_{\wfunc \in \W, \sfunc \in \Sspace}  \E_{t-1} \left [ U \left(  \wret_{t-1} R_{t}  \right)   \right]   \\
&\text{s.t. } \qquad \wret_{t-1} = \frac{{\w_{t-1}}^{\top}\proj_{t-1} }{\| {\w_{t-1}}^{\top}\proj_{t-1}  \|_1} \qquad \text{and}  \qquad \w_{t-1}= \wfunc(\sfunc(\epsilon_{t-1}^L)).
\end{align}
In the presence of trading costs, we calculate the expected utility of the portfolio net return by subtracting from the portfolio return $\wret_{t-1} R_{t}$ the trading costs that are associated with the stock allocation $\wret_{t-1}$. The trading costs can capture the transaction costs from frequent rebalancing and the higher costs of short selling compared to long positions. The stock weights $ \wret_{t-1} $ are normalized to add up to one in absolute value, which implicitly imposes a leverage constraint. The conditional expectation uses the general filtration $\mathcal{F}_{t-1}$.

This is a combined optimization problem, which simultaneously solves for the optimal allocation function and arbitrage signal function. As the weight is a composition of the two functions, i.e. $\w_{t-1}= \wfunc(\sfunc(\epsilon_{t-1}^L))$, the decomposition into a signal and allocation function is in general not uniquely identified. This means there can be multiple representations of $\sfunc$ and $\wfunc$, that will result in the same trading policy. We use a decomposition that allows us to compare the problem to classical arbitrage approaches, for which this separation is uniquely identified. The key feature of the signal function $\sfunc$ is that it models a time-series, that means it is a mapping that explicitly models the temporal order and the dependency between the elements of $\epsilon_{t-1}^L$. The allocation function $\wfunc$ can be a complex nonlinear function, but does not explicitly model time-series behavior. This means that $\wfunc$ is implicitly limited in the dependency patterns of its input elements that it can capture.

We will consider arbitrage trading that maximizes the Sharpe ratio or achieves the highest average return for a given level of variance risk. More specifically we will solve for 
\begin{align}\label{doubleOptimization}
&\max_{\wfunc \in \W, \sfunc \in \Sspace}  \frac{\E \left[ {w_{t-1}^R}^{\top} R_t  \right]}{\sqrt{\text{Var}({w_{t-1}^R}^{\top} R_t ) }}  \qquad \text{or} \qquad \max_{\wfunc \in \W, \sfunc \in \Sspace} \E [{w_{t-1}^R}^{\top} R_t ] - \gamma \text{Var}({w_{t-1}^R}^{\top} R_t )\\
&\text{s.t. } \qquad \wret_{t-1} =  \frac{{\w_{t-1}}^{\top}\proj_{t-1} }{\| {\w_{t-1}}^{\top}\proj_{t-1}  \|_1}\qquad \text{and} \qquad  \qquad \w_{t-1}= \wfunc(\sfunc(\epsilon_{t-1}^L)).
\end{align}
for some risk aversion parameter $\gamma$. We will consider this formulation with and without trading costs.\footnote{While the conventional mean-variance optimization applies a penalty to the variance, we found that it is numerically beneficial to use the standard deviation instead. Both formulations describe the same fundamental trade-off.} 

Many relevant models estimate the signal and allocation function separately. The arbitrage signals can be estimated as the parameters of a parametric time-series model, the serial moments for a given stationary distribution or a time-series filter. In these cases, the signal estimation solves a separate optimization problem as part of the estimation. Given the signals, the allocation function is the solution of
\begin{align}\label{eq:opt-problem}
\max_{\wfunc \in \W}   \E_{t-1} \left [ U \left(  \wret_{t-1} R_t  \right)   \right]    
 \qquad \text{s.t.} \qquad \wret_{t-1}  =  \frac{w^{\epsilon}(\theta_{t-1} )^{\top} \Phi_{t-1} }{\| w^{\epsilon}(\theta_{t-1})^{\top} \Phi_{t-1}  \|_1} .
\end{align}

We provide an extensive study of the importance of the different elements in statistical arbitrage trading. We find that the most important driver for profitable portfolios is the arbitrage signal function; that is, a good model to extract time-series behavior and to time the predictable mean reversion patterns is essential. The arbitrage portfolios of asset pricing models that are sufficiently rich result in roughly the same performance. Once an informative signal is extracted, parametric and nonparametric allocation functions can take advantage of it. We find that the key element is to consider a sufficiently general class of functions $\Sspace$ for the arbitrage signal and to estimate the signal that is the most relevant for trading. In other words, the largest gains in statistical arbitrage come from flexible time-series signals $\sfunc$ and a joint optimization problem.



\subsection{Models for Arbitrage Signal and Allocation Functions}

\begin{figure}[t!]
\centering
\tcaptab{Conceptual Arbitrage Model}\label{fig:model}\vspace{4pt}
\includegraphics[width=0.95\textwidth]{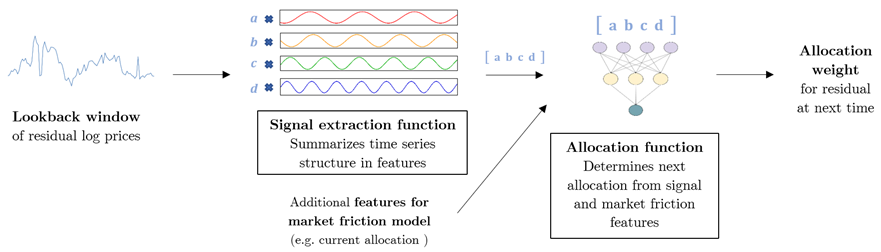}
\bnotetab{This figure illustrates the conceptual structure of our statistical arbitrage framework. The model takes as input the last $L$ cumulative returns of a residual portfolio on a lookback window at a given time and outputs the predicted optimal allocation weight for that residual for the next time. The model is composed of a signal extraction function and an allocation function.}
\label{fig:arbitrage-model}
\end{figure}

In this section we introduce different functional models for the signal and allocation functions. They range from the most restrictive assumptions for simple parametric models to the most flexible model, which is our sophisticated deep neural network architecture. The general problem is the estimation of a signal and allocation function given the residual time-series. Here, we take the residual returns as given, i.e. we have selected an asset pricing model. In order to illustrate the key elements of the allocation functions, we consider trading the residuals directly. Projecting the residuals back into the original return space is identical for the different methods and discussed in the empirical part. The conceptual steps are illustrated in Figure \ref{fig:model}.


The input to the signal extraction functions are the last $L$ cumulative residuals. We simplify the notation by dropping the time index $t-1$ and the asset index $n$ and define the generic input vector
\begin{align*}
x := \text{\bf Int} \left( \epsilon_{n,t-1}^L \right) = \begin{pmatrix}  \epsilon_{n,t-L} & \sum_{l=1}^2 \epsilon_{n,t-L-1+l} & \cdots &  \sum_{l=1}^L \epsilon_{n,t-L-1+l}  \end{pmatrix}.
\end{align*}
Here the operation $\text{\bf Int}(\cdot)$ simply integrates a discrete time-series. We can view the cumulative residuals as the residual ``price'' process. We discuss three different classes of models for the signal function $\sfunc$ that vary in the degree of flexibility of the type of patterns that they can capture. Similarly, we consider different classes of models for the allocation function $\wfunc$.

\subsubsection{Parametric Models}

Our first benchmark method is a parametric model and corresponds to classical mean-reversion trading. In this framework, the cumulative residuals $x$ are assumed to be the discrete realizations of continuous time model:
 \begin{align*}
 x = \begin{pmatrix}  X_1 & \cdots & X_L \end{pmatrix}.
 \end{align*}
Following the influential papers by \cite{avelee} and \cite{papayeo} we model $X_t$ as an  Ornstein-Uhlenbeck (OU) process  
 \begin{align*}
 d X_t = \kappa \left( \mu  - X_t \right) dt + \sigma d B_t
 \end{align*}   
 for a Brownian motion $B_t$. These are the standard models for mean-reversion trading and \cite{avelee} among others have shown their good empirical performance.

The parameters of this model are estimated from the moments of the discretized time-series, as described in detail along with the other implementation details in Appendix \ref{app:OU}. The parameters for each residual process, the last cumulative sum and a goodness of fit measure form the signals for the Ornstein-Uhlenbeck model:
 \begin{align*}
 \theta^{\text{OU}} = \begin{pmatrix} \hat \kappa & \hat \mu & \hat \sigma & X_L & R^2  \end{pmatrix}.
 \end{align*}
 Following \cite{papayeo} we also include the goodness of fit parameter $R^2$ as part of the signal. $R^2$ has the conventional definition of the ratio of squared values explained by the model normalized by total squared values. If the $R^2$ value is too low, the predictions of the model seem to be unreliable, which can be taken into account in a trading policy.
Hence, for each cumulative residual vector $\epsilon_{n,t-1}^L$ we obtain the signal
  \begin{align*}
 \theta^{\text{OU}}_{n,t-1} = \begin{pmatrix} \hat \kappa_{n,t-1} & \hat \mu_{n,t-1} & \hat \sigma_{n,t-1} & \sum_{l=1}^L \epsilon_{n,t-1+l} & R_{n,t-1}^2 \end{pmatrix}.
 \end{align*}   
\cite{avelee} and \cite{papayeo} advocate a classical mean-reversion thresholding rule, which implies the following allocation function\footnote{The allocation function is derived by maximizing an expected trading profit. This deviates slightly from our objective of either maximizing the Sharpe ratio or the expected return subject to a variance penalty. As this is the most common arbitrage trading rule, we include it as a natural benchmark.}:
\begin{align*}
{\bm w}^{\epsilon |\text{OU}}\left( \theta^{\text{OU}} \right)  = \left\{\begin{array}{ll}
        -1 & \text{if } \frac{X_L - \mu}{\sigma /\sqrt{2 \kappa}} > c_{\text{thresh}} \text{ and $R^2>c_{\text{crit}}$}  \\
        1 & \text{if } \frac{X_L - \mu}{\sigma /\sqrt{2 \kappa}} < -c_{\text{thresh}} \text{ and $R^2>c_{\text{crit}}$} \\
        0 & \text{otherwise } 
        \end{array} \right.
\end{align*}
The threshold parameters $c_{\text{thresh}}$ and $c_{\text{crit}}$ are tuning parameters. The strategy suggests to buy or sell residuals based on the ratio $\frac{X_L - \mu}{\sigma / \sqrt{2 \kappa}}$. If this ratio exceeds a threshold, it is likely that the process reverts back to its long term mean, which starts the trading. If the $R^2$ value is too low, the predictions of the model seem to be unreliable, which stops the trading. This will be our parametric benchmark model. It has a parametric model for both the signal and allocation function.

Figure \ref{fig:weights-and-signals} in the Appendix illustrates this model with an empirical example. In this figure we show the allocation weights and signals of the Ornstein-Uhlenbeck with threshold model as well as the more flexible models that we are going to discuss next. The models are estimated on the empirical data, and the residual is a representative empirical example. In more detail, we consider the residuals from five IPCA factors and estimate the benchmark models as explained in the empirical Section \ref{sec:interpretation}. The left subplots display the cumulative residual process along with the out-of-sample allocation weights $w_l^{\epsilon}$ that each model assigns to this specific residual. The evaluation of this illustrative example is a simplification of the general model that we use in our empirical main analysis. In this example, we consider trading only this specific residual and hence normalize the weights to $\{-1,0,1\}$. In our empirical analysis we trade all residuals and map them back into the original stock returns. The middle column shows the time-series of estimated out-of-sample signals for each model, by applying the $\theta_l$ arbitrage signal function to the previous $L=30$ cumulative returns of the residual. The right column displays the out-of-sample cumulative returns of trading this particular residual based on the corresponding allocation weights.

The last row in Figure \ref{fig:weights-and-signals} shows the results for the OU+Threshold model. The cumulative return of trading this residual is negative, suggesting that the parametric model fails. The residual time-series with the corresponding allocation weights in subplot (g) explain why. The trading allocation does not assign a positive weight during the uptrend and wrongly assigns a constant negative weight, when the residual price process follows a mean-reversion pattern with positive and negative returns. A parametric model can break down if it is misspecified. This is not only the case for trend patterns, but also if there are multiple mean reversion patterns of different frequencies. Subplot (h) shows the signal.\footnote{For better readability we have scaled the parameters of the OU process by a factor of five, but this still represents the same model as the scaling cancels out in the allocation function. As a minor modification, we use the ratio $\sigma/\sqrt{2 \kappa}$ as a signal instead of two individual parameters, as the conventional regression estimator of the OU process directly provides the ratio, but requires additional moments for the individual parameters. However, this results in an equivalent presentation of the model as only the ratio enters the allocation function.} We see that changes in the allocation function are related to sharp changes in at least one of the signals, but overall, the signal does not seem to represent the complex price patterns of the residual.

A natural generalization is to allow for a more flexible allocation function given the same time-series signals. 
We will consider for all our models also a general feedforward neural network (FFN) to map the signal into an allocation weight. FFNs are nonparametric estimators that can capture very general functional relationships.\footnote{Appendix \ref{app:ffn} provides the details for estimating a FFN as a functional mapping $\ffn: \mathbbm R^p  \rightarrow \mathbbm R$.} Hence, we also consider the additional model that restricts the signal function, but allows for a flexible allocation function:
 \begin{align*}
 {\bm w}^{\epsilon|\text{OU-FFN}} \left(  \theta^{\text{OU}}  \right) = \ffn \left( \theta^{\text{OU}}  \right).
 \end{align*}
 We will show empirically that the gains of a flexible allocation function are minor relative to the very simple parametric model.  
    
%

\subsubsection{Pre-Specified Filters}

As a generalization of the restrictive parametric model of the last subsection, we consider more general time-series models. Many relevant time-series models can be formulated as filtering problems. Filters are transformations of time-series that provide an alternative representation of the original time-series which emphasizes certain dynamic patterns.

A time-invariant linear filter can be formulated as 
\begin{align*}
\theta_l = \sum_{j=1}^L W_{j}^{\text{filter}} x_j,
\end{align*}
which is a linear mapping from $\mathbbm R^L$ into $\mathbbm R^L$ with the matrix $W^{\text{filter}} \in \mathbbm R^{ L \times L}$. The estimation of causal ARMA processes is an example for such filters. A spectral decomposition based on a frequency filter is the most relevant filter for our problem of finding mean reversion patterns.

A Fast Fourier Transform (FFT) provides a frequency decomposition of the original time-series and separates the movements into mean reverting processes of different frequencies. FFT applies the filter $W_j^{\text{FFT}} = e^{\frac{2 \pi i}{L} j}$ in the complex plane, but for real-valued time-series it is equivalent to fitting the following model:
\begin{align*}
x_l = a_0 + \sum_{j=1}^{L/2-1} \left( a_j \cdot \text{cos} \left( \frac{2 \pi j}{L} l \right) + b_j \cdot \text{sin} \left( \frac{2 \pi j}{L} l \right) \right) + a_{L/2} \text{cos} \left( \pi l \right).
\end{align*}  
The FFT representation is given by coefficients of the trigonometric representation
\begin{align*}
\theta^{\text{FFT}} = \begin{pmatrix}  a_0 & \cdots & a_{L/2} & b_1 & \cdots & b_{L/2 -1}  \end{pmatrix} \in \mathbbm R^L.
\end{align*}
The coefficients $a_l$ and $b_l$ can be interpreted as ``loadings'' or exposure to long or short-term reversal patterns. Note that the FFT is an invertible transformation. Hence, it simply represents the original time-series in a different form without losing any information. It is based on the insight that not the magnitude of the original data but the relative relationship in a time-series matters.

We use a flexible feedforward neural network for the allocation function
\begin{align*}
 {\bm w}^{\epsilon |\text{FFT}} \left(  \theta^{\text{FFT}}  \right) = \ffn \left( \theta^{\text{FFT}}  \right).
\end{align*}
The usual intuition behind filtering is to use the frequency representation to cut off frequencies that have low coefficients and therefore remove noise in the representation. The FFN is essentially implementing this filtering step of removing less important frequencies. 

We illustrate the model within our running example in Figure \ref{fig:weights-and-signals}. The middle row shows the results for the FFT+FFN model. The cumulative residual in subplot (d) seems to be a combination of low and high-frequency movements with an initial trend component. The signal in subplot (e) suggests that the FFT filter seems to capture the low frequency reversal pattern. However, it misses the high-frequency components as indicated by the simplistic allocation function. The trading strategy takes a long position for the first half and a short position for the second part. While this simple allocation results in a positive cumulative return, in this example it neglects the more complex local reversal patterns. 


While the FFT framework is an improvement over the simple OU model as it can deal with multiple combined mean-reversion patterns of different frequencies, it fails if the data follows a pattern that cannot be well approximated by a small number of the pre-specified basis functions.

For completeness, our empirical analysis will also report the case of a trivial filter, which simply takes the residuals as signals, and  combines them with a general allocation function:
\begin{align*}
 \sfunc^{\text{ident}} (x) &=x= \theta^{\text{ident}}   \\
  {\bm w}^{\epsilon |\text{FFN}} \left(  \theta^{\text{ident}}  \right) &= \ffn \left(x \right).
\end{align*}
This is a good example to emphasize the importance of a time-series model. While FFNs are flexible in learning low dimensional functional relationships, they are limited in learning a complex dependency model. For example, the FFN architecture we consider is not sufficiently flexible to learn the FFT transformation and hence has a worse performance on the original time-series compared to frequency-transformed time-series. While \cite{chen1999} have shown that FFNs are ``universal approximators'' of low-dimensional functional relationships, they also show that FFN can suffer from a curse of dimensionality when capturing complex dependencies between the input. Although the time domain and frequency domain representations of the input are equivalent under the Fourier transform, clearly the time-series model implied by the frequency domain representation allows for a more effective learning of an arbitrage policy. However, the choice of the pre-specified filter limits the time-series patterns that can be exploited. The solution is our data driven filter presented in the next section.

\subsubsection{Convolutional Neural Network with Transformer}\label{sec:theoryCNN}

Our benchmark machine learning model is a Convolutional Neural Network (CNN) combined with a Transformer. It uses the most advanced state-of-the art machine learning tools tailored to our problem. Convolutional networks are in fact the most successful networks for computer vision, i.e. for pattern detection. Transformers have rapidly become the model of choice for sequence modeling such as Natural Language Processing (NLP) problems, replacing older recurrent neural network models such as the Long Short-Term Memory (LSTM) network.

The CNN and transformer framework has two key elements: (1) Local filters and (2) the temporal combination of these local filters. The CNN can be interpreted as a set of data driven flexible local filters. A transformer can be viewed as a data driven flexible time-series model to capture complex dependencies between local patterns. We use the CNN+Transformer to generate the time-series signal. The allocation function is then modeled as a flexible data driven allocation with an FFN.

\begin{figure}[t!]
  \tcapfig{Examples of Local Filters}\label{fig:examples-basic-patterns}
  \begin{subfigure}[t]{.245\textwidth}
  \centering
    \includegraphics[width=1\linewidth]{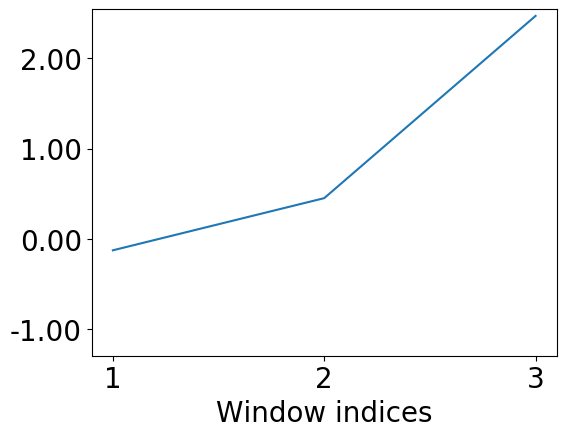}
    \caption{Upward trend}
   \end{subfigure}
  \begin{subfigure}[t]{.245\textwidth}
  \centering
    \includegraphics[width=1\linewidth]{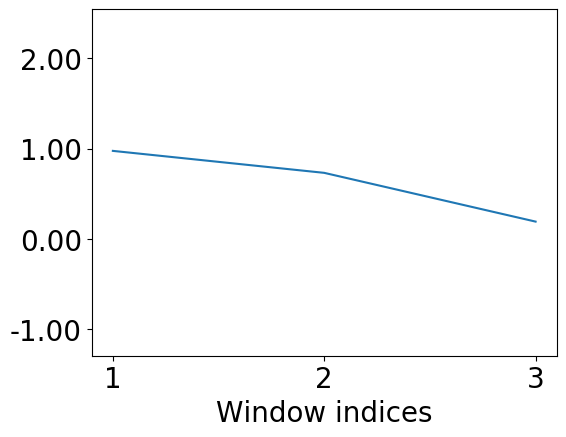}
    \caption{Downward trend}
   \end{subfigure}
  \begin{subfigure}[t]{.245\textwidth}
  \centering
    \includegraphics[width=1\linewidth]{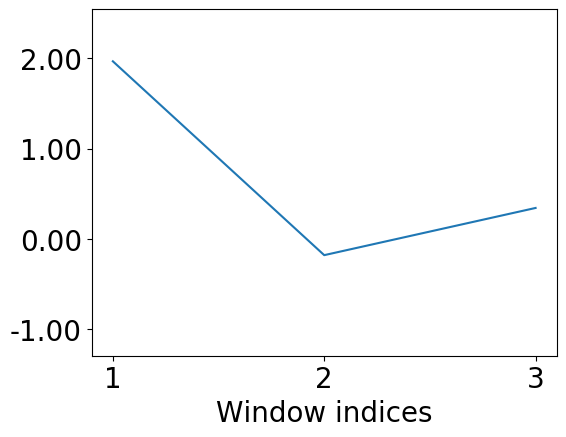}
    \caption{Up reversal}
   \end{subfigure}
  \begin{subfigure}[t]{.245\textwidth}
  \centering
    \includegraphics[width=1\linewidth]{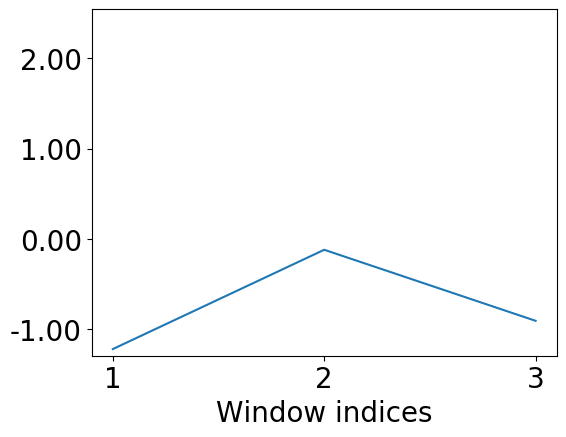}
    \caption{Down reversal}
   \end{subfigure}
\bnotefig{These figures show the most important local filters estimated for the benchmark model in our empirical analysis. These are projections of our higher dimensional nonlinear filter from a 2-layer CNN into two-dimensional linear filters.}
\end{figure}

The CNN estimates $D$ local filters of size $\ds$:
\begin{align*}
y_l^{(0)} =\sum_{m=1}^{\ds} W_m^{(0)} x_{l-m+1}
\end{align*} 
for a matrix $W^{(0)} \in \mathbbm R^{\ds \times D}$. The local filters are a mapping from $x \in \mathbbm R^L$ to $y^{(0)} \in \mathbbm R^{L \times D}$ given by the convolution $y^{(0)} =W^{(0)} * x$. Figure \ref{fig:examples-basic-patterns} shows examples of these local filters for $\ds=3$. The values of $y^{(0)}$ can be interpreted as the ``loadings'' or exposure to local basis patterns. For example, if $x$ represents a global upward trend, its filtered representation should have mainly large values for the local upward trend filter.

\begin{figure}[t!]
\centering
\tcaptab{Convolutional Network Architecture }\label{fig:convolutional}
\includegraphics[width=1\textwidth]{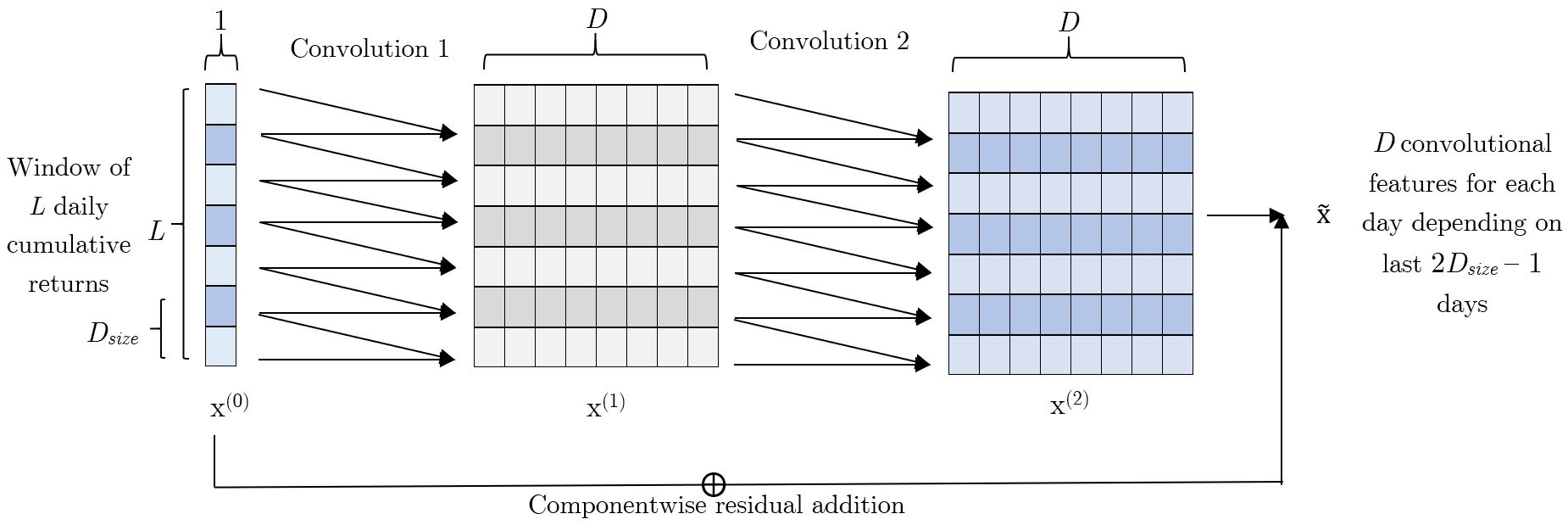}
\bnotetab{This figure shows the structure of our convolutional network. The network takes as input a window of $L$ consecutive daily cumulative returns a residual, and outputs $D$ features for each block of $\ds$ days. Each of the features is a nonlinear function of the observations in the block, and captures a common pattern.}
\end{figure}

The convolutional mapping can be repeated in multiple layers to obtain a multi-layer CNN. First, the output of the first layer of the CNN is transformed nonlinearly by applying the $\text{ReLU}(\cdot)$ function: 
\begin{align*}
x_{l,d}^{(1)} =\text{ReLU} \left( y_{l,d}^{(0)}\right):=\max(y_{l,d}^{(0)},0).
\end{align*}
The second layer is given by a higher dimensional filtering projection:
\begin{align*}
y_{l,d}^{(2)} &= \sum_{m=1}^{\ds} \sum_{j=1}^D W_{d,j,m}^{(1)} x_{l-m+1,j}^{(1)}, \\
x_{l,d}^{(2)} &= \text{ReLU} \left( y_{l,d}^{(1)} \right).
\end{align*}
The final output of the CNN is $\tilde x \in \mathbbm R^{L \times D}$. Our benchmark model is a 2-layered convolutional neural network. The number of layers is a hyperparameter selected on the validation data. Figure \ref{fig:convolutional} illustrates the structure of the 2-layer CNN. While this description captures all the conceptual elements, the actual implementation includes additional details, such as bias terms, instance normalization and residual connection to improve the implementation as explained in Appendix \ref{app:CNNtrans}.


For a 1-layer CNN without the final nonlinear transformation, i.e. for a simple local linear filter, the patterns can be visualized by the vectors $W_m^{(0)}$. In our case of a 2-layer CNN the local filter can capture more complex patterns as it applies a 3-dimensional weighting scheme in the array $W^{(1)}$ and nonlinear transformations. In order to visualize the type of patterns, we project the local filter into a simple local linear filter. We want to find the basic patterns that activate only one of the $D$ filters, but none of the others, i.e. we are looking for an orthogonal representation of the projection.\footnote{A local filter can be formalized as a mapping from the local $\ds$ points of a sequence to the activation of the $D$ filters: $\phi: \mathbbm R^{\ds} \rightarrow \mathbbm R^D$.
Denote by $e_d \in \mathbbm R^D$ a vector that is 0 everywhere except for the value 1 at position $d$, i.e. $e_d= \begin{pmatrix} 0 & \cdots &1 & 0 & \cdots 0  \end{pmatrix}$. Fundamentally, we want to invert the local filter to obtain $\phi^{-1} (e_d)$ to find the local sequences that only activates filter $d$. In general, the inverse is a set and not unique. Our example basic patterns in Figure \ref{fig:examples-basic-patterns} solve
\begin{align*}
\text{argmin}_{x_{\text{loc},d} \in \mathbbm R^{\ds}}  \| \phi(x_{\text{loc},d} ) - e_d \|_2 \qquad \text{for $d=1,...,D$.}
\end{align*}}

The example plots for local filters in Figure \ref{fig:examples-basic-patterns} are projections of our higher dimensional nonlinear filter into two-dimensional linear filters. The examples show some of the most important local filters for our empirical benchmark model. While these projections are of course not complete representations of the nonlinear filters of the CNN, they provide an intuition for the type of patterns which are activated by specific filters. Our 2-layer CNN network has a local window size of $\ds=2$, but because of the 2-layer structure it captures information from two neighboring points. Hence, the projection on a one-dimensional linear filter has a local window size of three as depicted in Figure \ref{fig:examples-basic-patterns}.   

The output of the CNN $\tilde x \in \mathbbm R^{L \times D}$ is used as an input to the transformer. The CNN projection provides a more informative representation of the dynamics than the original time-series as it captures the relative local dependencies between data points. However, by construction the CNN is only a local representation, and we need the transformer network to detect the global patterns. A transformer network is a model of temporal dependencies between local filters. Given the local structure $\tilde x$ the transformer estimates the temporal interactions between the $L$ different blocks by computing a ``global pattern projection'':

%

Assume there are $H$ different global patterns. The transformer will calculate projections on these $H$ patterns with the ``attention weights''. We first introduce a simplified linear projection model before extending it to the actual transformer. For each of the $i=1,...,H$ patterns we have projections defined by $\alpha_i \in \mathbbm R^{L \times L}$:
\begin{align*}
h^{\text{simple}}_{i} = \sum_{j=1}^L  \alpha_{i,j} \tilde x_j \qquad \text{ for $i=1,..,H$} .
\end{align*} 
The ``attention function'' ${\bm \alpha_i}(.,.) \in [0,1]$ captures dependencies between the last local patterns $\tilde x_L$ and the prior local patterns $\tilde x_j$:
\begin{align*}
\alpha_{i,j} = {\bm \alpha_i} \left( \tilde x_L, \tilde x_j \right) \qquad \text{for $i=1,...,H$}.
\end{align*}
Each projection $h^{\text{simple}}_{i} $ is called an ``attention head''. These attention heads $h^{\text{simple}}_{i} $ could be interpreted as ``loadings'' or ``exposure'' for a specific ``pattern factor'' $\alpha_i$. For example, a global upward trend can be captured by an attention function that puts weight on subsequent local upward trends. Another example would be sinusoidal mean reversion patterns which would put weights on alternating ``curved'' local basis patterns. The projection on these weights captures how much a specific time-series $\tilde x$ is exposed to this global pattern. Hence, $h^{\text{simple}}_{i}$ measures the exposure to the global pattern $i$ of the time-series $\tilde x$. Each attention head can focus on a specific global pattern, which we then combine to obtain our signal. 

The fundamental challenge is to learn attention functions that can model complex dependencies. The crucial innovation in transformers is their modeling of the attention functions ${\bm \alpha_i}$ and attention heads $h_{i}$. In order to deal with the high dimensionality of the problem, transformers consider lower dimensional projections of $\tilde x$ into $\mathbbm R^{D/H}$ and use the lower-dimensional scaled dot product attention mechanism for ${\bm \alpha_i}$ as explained in Appendix \ref{app:CNNtrans}. More specifically, each attention head $h_i \in \mathbbm R^{L \times D/H}$ is based on\footnote{The actual implementation also includes bias terms which we neglect here for simplicity. The Appendix provides the implementation details.} the projected input $\tilde x W_i^V$ with $W_i^V \in \mathbbm R^{D \times D/H}$ and $\alpha_i \in \mathbbm R^{L \times L}$:
\begin{align*}
h_{i} =  \alpha_{i} {\tilde x} W_i^V  \qquad \text{ for $i=1,..,H$}. 
\end{align*} 
The projection on all global basis patterns $h \in \mathbbm R^{L \times D/H}$ is given by a weighted linear combination of the different attention heads
\begin{align*}
h^{\text{proj}} = \begin{pmatrix} h_1 & \cdots & h_H  \end{pmatrix} W^O
\end{align*}
with $W^O \in \mathbbm R^{D \times D}$. This final projection can, for example, model a combination of a global trend and mean reversion patterns. In conclusion, $h^{\text{proj}}$ represents the time-series in terms of the $H$ global patterns. This is analogous to a Fourier filter, but without pre-specifying the global patterns a priori. All parts of the CNN+Transformer network, i.e. the local patterns, the attention functions and the projections on global patterns, are estimated from the data. 

\begin{figure}[t!]
\centering
\tcaptab{Transformer Network Architecture}\label{fig:transformer}
\includegraphics[width=0.95\textwidth]{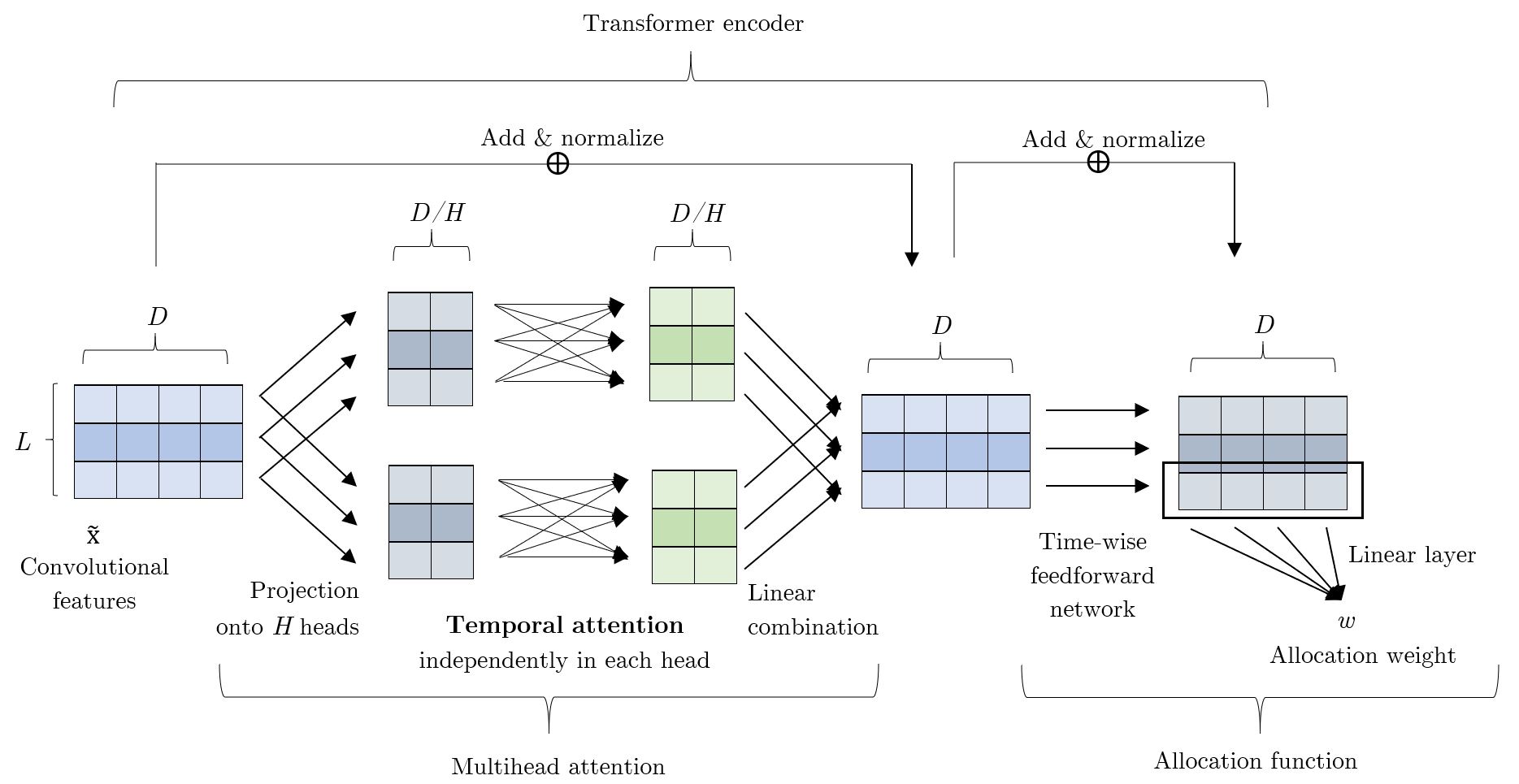} 
\bnotetab{This figure shows the structure of our transformer network. The model takes as input the matrix $\tilde{x}\in\R^{L\times D}$ that we obtain as output of the convolutional network depicted in Figure \ref{fig:convolutional}, which contains $D$ features for each of the $L$ blocks of the original time series. These features are projected onto $H$ attention heads, which capture the global temporal dependency patterns. The projections on these attention heads represent our arbitrage signals, which are the input to the feedforward network that estimates the allocation weight for the residual on the next day.}
\end{figure}

The trading signal $\theta^{\text{CNN+Trans}}$ equals the global pattern projection for the final cumulative return projection\footnote{In principle, we can use the complete matrix $h \in \mathbbm R^{L \times D}$ as the signal. However, conceptually the global pattern at the end of the time period should be the most relevant for the next realization of the process. We have also implemented a transformer that uses the full matrix, with similar results and the variable importance rankings suggest that only $h_L$ is selected in the allocation function.} $h^{\text{proj}}_L$:
\begin{align*}
\theta^{\text{CNN+Trans}}= h^{\text{proj}}_L \in \mathbbm R^H,
\end{align*} 
which is then used as input to a time-wise feedforward network allocation function
\begin{align*}
 {\bm w}^{\epsilon |\text{CNN+Trans}} \left(  \theta^{\text{CNN+Trans}}  \right) = \ffn \left( \theta^{\text{CNN+Trans}}  \right).
 \end{align*}
The separation between signal and allocation is not uniquely identified as we use a joint optimization problem. We have chosen a separation that maps naturally into the classical examples considered in the previous subsections. Figure \ref{fig:transformer} illustrates the transformer network architecture. We have presented a 1-layer transformer network, which is our benchmark model. The transformed data can be used as input in more iterations of the transformer to obtain a multi-layer transformer.

We illustrate the CNN+Transformer model in the first row of Figure \ref{fig:weights-and-signals} in the Appendix for an empirical residual example. First, it is apparent that the cumulative returns of the strategy in subplot (c) outperforms the previous two models. This is because the allocation weights in subplot (a) capture not only the low frequency reversal patterns, but also the high-frequency cycles and trend components. This also implies that the allocation weights change more frequently to capture the higher frequency components. This more sophisticated allocation function requires a more complex signal as illustrated in subplot (b). Each change in the allocation can be traced back to changes in at least one of the signals. While the signals themselves are hard to interpret, we will leverage the transformer structure to extract interpretable ``global dependency factors'' in our main analysis. Figure \ref{fig:example-trading-residuals-appendix} in the Appendix provides another example to illustrate the differences between the three models. This example has a strong negative trend component with a superimposed mean-reversion. Only the CNN+Transformer captures both type of patterns.

\section{Empirical Analysis}\label{sec:empirical}

\subsection{Data}\label{sec:data}
We collect daily equity return data for the securities on CRSP from January 1978 through December 2016. We use the first part of the sample to estimate the various factor models, which gives us the residuals for the time period from January 1998 to December 2016 for the arbitrage trading. The arbitrage strategies trade on a daily frequency at the close of each day. We use the daily adjusted returns to account for dividends and splits and the one-month Treasury bill rates from the Kenneth French Data Library as the risk-free rate. In addition, we complement the stock returns with the 46 firm-specific characteristics from \cite{DeepMarkus}, which are listed in Table \ref{tab:category}. All these variables are constructed either from accounting variables from the CRSP/Compustat database or from past returns from CRSP. The full details on the construction of these variables are in the Internet Appendix of \cite{DeepMarkus}.

Our analysis uses only the most liquid stocks in order to avoid trading and market friction issues. More specifically, we consider only the stocks whose market capitalization at the previous month was larger than 0.01\% of the total market capitalization at that previous month, which is the same selection criterion as in \cite{kozak2017}. On average this leaves us with approximately the largest 550 stocks, which correspond roughly to the S\&P 500 index. This is an unbalanced dataset, as the stocks that we consider each month need not be the same as in the next month, but it is essentially balanced on a daily frequency in rolling windows of up to one year in our trading period from 1998 through 2016. For each stock we have its cross-sectionally centered and rank-transformed characteristics of the previous month. This is a standard transformation to deal with the different scales which is robust to outliers and time-variation, and has also been used in \cite{DeepMarkus}, \cite{kozak2017}, \cite{IPCA}, and \cite{Freyberger}.

Our daily residual time-series start in 1998 as we have a large number of missing values in daily individual stock returns prior to this date, but almost no missing daily values in our sample.\footnote{Of all the stocks that have daily returns observed in a the local lockback window of $L=30$ days, only 0.1\% have a missing return the next day for the out-of-sample trading, in which case we do not trade this stock. Hence, our data set of stocks with market capitalization higher than 0.01\% of the total market capitalization, has essentially no missing daily values on a local window for the time period after 1998.} We want to point out that the time period after 1998 also seems to be more challenging for arbitrage trading or factor trading, and hence our results can be viewed as conservative lower bounds.

\subsection{Factor model estimation}\label{sec:factor-model-estimation} 

As discussed in Section \ref{sec:arbitrageportfolios}, we construct the statistical arbitrage portfolios by using the residuals of a general factor model for the daily excess returns of a collection of stocks. 
In particular, we consider the three empirically most successful families of factor models in our implementation. For each family, we conduct a rolling window estimation to obtain daily residuals out of sample from 1998 through 2016. This means that the residual composition matrix $\Phi_{t-1}$ of equation (\ref{eq:transitionMatrix}) depends only on the information up to time $t-1$, and hence there is no look-ahead bias in trading the residuals. The rolling window estimation is necessary because of the time-variation in risk exposure of individual stocks and the unbalanced nature of a panel of individuals stock returns.

The three classes of factor models consists of pre-specified factors, latent unconditional factors and latent conditional factors:
\begin{enumerate}
    \item \textbf{Fama-French factors:} We consider 1, 3, 5 and 8 factors based on various versions and extensions of the Fama-French factor models and downloaded from the Kenneth French Data Library. We consider them as tradeable assets in our universe. Each model includes the previous one  and adds additional characteristic-based risk factors:
    \begin{enumerate}
    \item $K=1$: CAPM model with the excess return of a market factor
    \item $K=3$: Fama-French 3 factor model includes a market, size and value factor
    \item $K=5$: Fama-French 3 factor model + investment and profitability factors
    \item $K=8$: Fama-French 5 factor model + momentum, short-term reversal and long-term reversal factors.
    \end{enumerate}
We estimate the loadings of the individual stock returns daily with a linear regression on the factors with a rolling window on the previous 60 days and compute the residual for the current day out-of-sample. This is the same procedure as in \cite{carhart}. At each day we only consider the stocks with no missing observations in the daily returns within the rolling window, which in any window removes at most 2\% of the stocks given our market capitalization filter.
    \item \textbf{PCA factors:} We consider 1, 3, 5, 8, 10, and 15 latent factors, which are estimated daily on a rolling window. At each time $t-1$, we use the last 252 days, or roughly one trading year, to estimate the correlation matrix from which we extract the PCA factors.\footnote{This is the same procedure as in \cite{avelee}.} Then, we use the last 60 days to estimate the loadings on the latent factors using linear regressions, and compute residuals for the current day out-of-sample. At each day we only consider the stocks with no missing observations in the daily returns during the rolling window, which in any window removes at most 2\% of the stocks given our market capitalization filter.
\item \textbf{IPCA factors:} We consider 1, 3, 5, 8, 10, and 15 factors in the Instrumented PCA (IPCA) model of \cite{IPCA}. This is a conditional latent factor model, in which the loadings $\beta_{t-1}$ are a linear function of the asset characteristics at time $t-1$. As the characteristics change at most each month, we reestimate the IPCA model on rolling window every year using the monthly returns and characteristics of the last 240 months. The IPCA provides the factor weights and loadings for each stock as a function of the stock characteristics. Hence, we do not need to estimate the loadings for individual stocks with an additional time-series regression, but use the loading function and the characteristics at time $t-1$ to obtain the out-of-sample residuals at time $t$. The other details of the estimation process are carried out in the way outlined in \cite{IPCA}.    
\end{enumerate}
In addition to the factor models above, we also include the excess returns of the individual stocks without projecting out the factors. This ``zero-factor model'' simply consists of the original excess returns of stocks in our universe and is denoted as $K=0$. For each factor model, in our empirical analysis we observe that the cumulative residuals exhibit consistent and relatively regular mean-reverting behavior. After taking out sufficiently many factors, the residuals of different stocks are only weakly correlated. 

\subsection{Implementation}\label{sec:implementation}
Given the daily out-of-sample residuals from 1998 through 2016 we estimate the trading signal and policy on a rolling window to obtain the out-of-sample returns of the strategy. For each strategy we calculate the 
annualized sample mean $\mu$, annualized volatility $\sigma$ and annualized Sharpe ratio\footnote{We obtain the annualized metrics from the daily returns using the standard calculations $ \mu= \frac{252}{T}\sum_{t=1}^TR_t$ and $\sigma = \sqrt{ \frac{252}{T}\sum_{t=1}^T(R_t- \mu)^2}$.} $\text{SR} =\frac{\mu}{\sigma}$. The Sharpe ratio represents a risk-adjusted average return. Our main models estimate arbitrage strategies to maximize the Sharpe ratio without transaction costs. In Section \ref{sec:MVO} we also consider a mean-variance objective and in Section \ref{sec:market-frictions-results} we include transaction costs in the estimation and evaluation.

Our strategies trade the residuals of all stocks, which are mapped back into positions of the original stocks. We use the standard normalization that the absolute values of the individual stock portfolio weights sums up to one, i.e. we use the normalization $\| \omega^R_{t-1} \|_1=1$. This normalization implies a leverage constraint as short positions are bounded by one. The trading signal is based on a local lookback window of $L=30$ days. We show in Section \ref{sec:time-stability}, that the results are robust to this choice and are very comparable for a lookback window of $L=60$ days. Our main results use a rolling window of 1,000 days to estimate the deep learning models. For computational reasons we re-estimate the network only every 125 days using the previous 1,000 days. Section \ref{sec:time-stability} shows that our results are robust to this choice. Our main results show the out-of-sample trading performance from January 2002 to December 2016 as we use the first four years to estimate the signal and allocation function. 

The hyperparameters for the deep learning models are based on the validation results summarized in Appendix \ref{sec:hyperparameters}. Our benchmark model is a 2-layer CNN with $D=8$ local convolutional filters and local window size of $D_{size}=2$ days. The transformer has $H=4$ attention heads, which can be interpreted as capturing four different global patterns. The results are extremely robust to the choice of hyperparameters. Appendix \ref{sec:learning} includes all the technical details for implementing the deep learning models.

\subsection{Main Results}

Table \ref{tab:deep-results} displays the main results for various arbitrage models. It reports the annualized Sharpe ratio, mean return and volatility for our principal deep trading strategy CNN+Transformer and the two benchmark models, Fourier+FFN and OU+Threshold, for every factor model described in Section \ref{sec:factor-model-estimation}. The CNN+Transformer model and Fourier+FFN model are estimated with a Sharpe ratio objective. We obtain the daily out-of-sample residuals for different number of factors $K$ for the time period January 1998 to December 2016. The daily returns of the out-of-sample arbitrage trading is then evaluated from January 2002 to December 2016, as we use a rolling window of four years to estimate the deep learning models.

\begin{table}[t!]
\begin{center}
    \setlength{\tabcolsep}{4pt}
    \small
    \tcaptab{OOS Annualized Performance Based on Sharpe Ratio Objective}
    \label{tab:deep-results}
    \scalebox{1.0}{
    \begin{tabular}{c||c||ccc||ccc||ccc}
    \multicolumn{11}{c}{\vspace{-5pt}} \\
    \toprule
     & Factors
     & \multicolumn{3}{c||}{Fama-French} 
     & \multicolumn{3}{c||}{PCA} 
     & \multicolumn{3}{c}{IPCA}\\
    \cmidrule{1-11}
     Model
     & K  
     & SR & $\mu$ & $\sigma$ 
     & SR & $\mu$ & $\sigma$ 
         & SR & $\mu$ & $\sigma$   \\
    \midrule
    \midrule
    &
    0  
    & {1.64} & 13.7\% & 8.4\%
    & {1.64} & 13.7\% & 8.4\%
    & {1.64} & 13.7\% & 8.4\%
    \\ 
    &
    1  
    & {3.68} & 7.2\% & 2.0\%
    & {2.74} & 15.2\% & 5.5\%
    & {3.22} & 8.7\% & 2.7\%
    \\ 
    CNN &
    3  
    & {3.13} & 5.5\% & 1.8\%
    & {3.56} & 16.0\% & 4.5\%
    & {3.93} & 8.6\% & 2.2\%
    \\ 
    + &
    5  
    & {3.21} & 4.6\% & 1.4\%
    & {3.36} & 14.3\% & 4.2\%
    & {4.16} & 8.7\% & 2.1\%
    \\ 
    Trans &
    8
    & {2.49} & 3.4\% & 1.4\%
    & 3.02 & 12.2\% & 4.0\%
    & 3.95 & 8.2\% & 2.1\%
    \\
    &
    10  
    & - & - & -
    & {2.81} & 10.7\% & 3.8\%
    & {3.97} & 8.0\% & 2.0\%
    \\ 
     &
    15  
    & - & - & -
    & {2.30} & 7.6\% & 3.3\%
    & {4.17} & 8.4\% & 2.0\%
    \\
    \midrule
    \midrule
    &
    0  
    & 0.36 & 4.9\% & 13.6\%
     & 0.36 & 4.9\% & 13.6\%
     & 0.36 & 4.9\% & 13.6\%
    \\ 
    &
    1  
    & 0.89 &  3.2\% &  3.5\%
    & 0.80 & 8.4\% & 10.6\%
    & 1.24 & 6.3\% & 5.0\%
    \\ 
    Fourier &
    3  
    & 1.32 &  3.5\% & 2.7\%
    & 1.66 & 11.2\% & 6.7\%
    & 1.77 & 7.8\% & 4.4\%
    \\ 
    + &
    5  
    & 1.66 &  3.1\% & 1.8\%
    & 1.98 & 12.4\% & 6.3\%
    & 1.90 & 7.7\% & 4.1\%
    \\ 
    FFN &
    8
    & 1.90 &  3.1\% & 1.6\%
    & 1.95 & 10.1\% & 5.2\%
    & 1.94 &  7.8\% & 4.0\%
    \\
    &
    10  
    & - & - & -
    & 1.71 & 8.2\% & 4.8\%
    & 1.93 & 7.6\% & 3.9\%
    \\ 
    &
    15  
    & - & - & -
    & 1.14 & 4.8\% & 4.2\%
    & 2.06 & 7.9\% & 3.8\%
    \\
    \midrule
    \midrule
    &
    0  
    & -0.18 & -2.4\% & 13.3\%
    & -0.18 & -2.4\% & 13.3\%
    & -0.18 & -2.4\% & 13.3\%
    \\ 
    &
    1  
    & 0.16 & 0.6\% & 3.8\%
    & 0.21 & 2.1\% & 10.4\%
    & 0.60 & 3.0\% & 5.1\%
    \\ 
    OU &
    3  
    & 0.54 & 1.6\% & 3.0\%
    & 0.77 & 5.2\% & 6.8\%
    & 0.88 & 3.8\% & 4.3\%
    \\ 
    + &
    5  
    & 0.38 & 0.9\% & 2.3\%
    & 0.73 & 4.4\% & 6.1\%
    & 0.97 & 3.8\% & 4.0\%
    \\ 
    Thresh &
    8
    & {1.16} & 2.8\% & 2.4\%
    & 0.87 & 4.4\%   & 5.1\%
    & 0.91 & 3.5\%   & 3.8\%
    \\
    &
    10  
     & - & - & -
    & 0.63 & 2.9\% & 4.6\%
    & 0.86 & 3.1\% & 3.6\%
    \\ 
    &
    15  
     & - & - & -
    & 0.62 & 2.4\% & 3.8\%
    & 0.93 & 3.2\% & 3.5\%
    \\ 
    \bottomrule
    \end{tabular}}
    \bnotetab{This table shows the out-of-sample annualized Sharpe ratio (SR), mean return ($\mu$), and volatility ($\sigma$) of our three statistical arbitrage models for different numbers of risk factors $K$, that we use to obtain the residuals. We use the daily out-of-sample residuals from January 1998 to December 2016 and evaluate the out-of-sample arbitrage trading from January 2002 to December 2016. CNN+Trans denotes the convolutional network with transformer model, Fourier+FFN estimates the signal with a FFT and the policy with a feedforward neural network and lastly, OU+Thres is the parametric Ornstein-Uhlenbeck model with thresholding trading policy. The two deep learning models are calibrated on a rolling window of four years and use the Sharpe ratio objective function. The signals are extracted from a rolling window of $L=30$ days. The $K=0$ factor model corresponds to directly using stock returns instead of residuals for the signal and trading policy. 
}
   \end{center} \end{table}

First, we confirm that it is crucial to apply arbitrage trading to residuals and not individual stock returns. The stock returns, denoted as the $K=0$ model, perform substantially worse than any type of residual within the same model and factor family. This is not surprising as residuals for an appropriate factor model are expected to be better described by a model that captures mean reversion. Importantly, individual stock returns are highly correlated and a substantial part of the returns is driven by the low dimensional factor component.\footnote{\cite{pelger2019} shows that around one third of the individual stock returns is explained by a latent four-factor model.} Hence, the complex nonparametric models are actually not estimated on many weakly dependent residual time-series, but most time-series have redundant information. In other words, the models are essentially calibrated on only a few factor time-series, which severely limits the structure that can be estimated. However, once we extract around $K=5$ factors with any of the different factor models, the performance does not substantially increase by adding more factors. This suggests that most of commonality is explained by a small number of factors. 

Second, the CNN+Transformer model strongly dominates the other benchmark models in terms of Sharpe ratio and average return. The Sharpe ratio is approximately twice as large as for a comparable Fourier+FFN model and four times higher for the corresponding parametric OU+Threshold model. Using IPCA residuals, the CNN+Transformer achieves the impressive out-of-sample Sharpe ratio of around 4, in spite of trading only the most liquid large cap stocks and the time period after 2002. The mean returns of the CNN+Transformer are similar to the Fourier+FFN model, but have substantially smaller volatilities, which results in the higher Sharpe ratios. The parametric mean-reversion model achieves positive mean returns with Sharpe ratios close to one for the IPCA residuals, but as expected is too restrictive relative to the flexible models. The Fourier+FFN has the same flexibility as the CNN+Transformer in its allocation function, but is restricted to a pre-specified signal structure. The difference in performance quantifies the importance of extracting the complex time-series signals.

Third, the average return of the arbitrage strategies is large in spite of the leverage constraints. Normalizing the individual stock weights $ \wret_{t-1}$ so sum up in absolute value to one limits the short-selling. The CNN+Transformer with a five-factor PCA residual achieves an attractive annual mean return of around 14\%. This means that the strategies do not require an infeasible amount of leverage to yield an average return that might be required by investors. In other words, the high Sharpe ratios are not the results of vanishing volatility but a combination of high average returns with moderate volatility.

Fourth, the arbitrage strategies are qualitatively robust to the choice of factor models to obtain residuals. The Fama-French and PCA factor lead to very similar Sharpe ratio results, suggesting that they explain a similar amount of co-movement in the data. However, as the mean returns of PCA factors are usually higher than the mean returns of the Fama-French factors, the risk factors are different. This confirms the findings of \cite{pelger2019} and \cite{lettaupelger2018}, who show that PCA factors do not coincide with Fama-French type factors and explain different mean returns. The IPCA factors use the additional firm-specific characteristic information. The resulting residuals achieve the highest Sharpe ratios, which illustrates that conditional factor models can capture more information than unconditional models. Including the momentum and reversal factors in the Fama-French 8 factor models to obtain residuals still results in profitable arbitrage strategies. Hence, the arbitrage strategies are not simply capturing a conventional price trend risk premium. 

\begin{table}[t!]
    \centering
    {\footnotesize
    \tcaptab{Significance of Arbitrage Alphas based on Sharpe Ratio Objective}
    \setlength{\tabcolsep}{4pt} 
    \scalebox{0.9}{
    \begin{tabular}{c||ccccc||ccccc||ccccc}
    \toprule
    \multicolumn{16}{c}{CNN+Trans model}\\
    \midrule
     & \multicolumn{5}{c||}{Fama-French} & \multicolumn{5}{c||}{PCA} & \multicolumn{5}{c}{IPCA} \\
    \cmidrule{2-16}
    K & $\alpha$ & $t_\alpha$ & $R^2$ & $\mu$ & $t_\mu$ & $\alpha$ & $t_\alpha$ & $R^2$ & $\mu$ & $t_\mu$ & $\alpha$ & $t_\alpha$ & $R^2$ & $\mu$ & $t_\mu$ \\
    \midrule
    0       &      11.6\% &      6.4$^{***}$ & 30.3\% & 13.7\% &     6.3$^{***}$ & 11.6\% &      6.4$^{***}$ & 30.3\% & 13.7\% &     6.3$^{***}$ & 11.6\% &      6.4$^{***}$ & 30.3\% & 13.7\% &     6.3$^{***}$ \\
    1       &       7.0\% &       14$^{***}$ &  2.4\% &  7.2\% &      14$^{***}$ & 14.9\% &       10$^{***}$ &  0.6\% & 15.2\% &      11$^{***}$ &  8.1\% &       12$^{***}$ &  9.5\% &  8.7\% &      12$^{***}$ \\
    3       &       5.5\% &       12$^{***}$ &  1.2\% &  5.5\% &      12$^{***}$ & 15.8\% &       14$^{***}$ &  1.7\% & 16.0\% &      14$^{***}$ &  8.2\% &       15$^{***}$ &  6.0\% &  8.6\% &      15$^{***}$ \\
    5       &       4.5\% &       12$^{***}$ &  2.3\% &  4.6\% &      12$^{***}$ & 14.1\% &       13$^{***}$ &  1.3\% & 14.3\% &      13$^{***}$ &  8.3\% &       16$^{***}$ &  3.9\% &  8.7\% &      16$^{***}$ \\
    8       &       3.3\% &      9.4$^{***}$ &  2.1\% &  3.4\% &     9.6$^{***}$ & 12.0\% &       12$^{***}$ &  0.9\% & 12.2\% &      12$^{***}$ &  7.8\% &       15$^{***}$ &  5.0\% &  8.2\% &      15$^{***}$ \\
    10      &          - &           - &     - &     - &          - & 10.5\% &       11$^{***}$ &  0.7\% & 10.7\% &      11$^{***}$ &  7.7\% &       15$^{***}$ &  4.0\% &  8.0\% &      15$^{***}$ \\
    15      &          - &           - &     - &     - &          - &  7.5\% &      8.8$^{***}$ &  0.5\% &  7.6\% &     8.9$^{***}$ &  8.1\% &       16$^{***}$ &  4.2\% &  8.4\% &      16$^{***}$ \\
    \midrule
    \midrule
    \multicolumn{16}{c}{Fourier+FFN model}\\
    \midrule
     & \multicolumn{5}{c||}{Fama-French} & \multicolumn{5}{c||}{PCA} & \multicolumn{5}{c}{IPCA} \\
    \cmidrule{2-16}
    K & $\alpha$ & $t_\alpha$ & $R^2$ & $\mu$ & $t_\mu$ & $\alpha$ & $t_\alpha$ & $R^2$ & $\mu$ & $t_\mu$ & $\alpha$ & $t_\alpha$ & $R^2$ & $\mu$ & $t_\mu$ \\
    \midrule
    0       &       2.7\% &         0.8 & 8.6\% & 4.9\% &        1.4 &  2.7\% &         0.8 & 8.6\% &  4.9\% &        1.4 &  2.7\% &         0.8 &  8.6\% & 4.9\% &        1.4 \\
    1       &       3.0\% &       3.3$^{**}$ & 3.3\% & 3.2\% &     3.5$^{***}$ &  7.4\% &       2.7$^{**}$ & 3.3\% &  8.4\% &      3.1$^{**}$ &  4.8\% &      4.0$^{***}$ & 16.4\% & 6.3\% &     4.8$^{***}$ \\
    3       &       3.2\% &      4.7$^{***}$ & 4.2\% & 3.5\% &     5.1$^{***}$ & 10.9\% &      6.3$^{***}$ & 2.2\% & 11.2\% &     6.4$^{***}$ &  6.8\% &      6.4$^{***}$ & 13.0\% & 7.8\% &     6.9$^{***}$ \\
    5       &       2.9\% &      6.1$^{***}$ & 3.5\% & 3.1\% &     6.4$^{***}$ & 12.1\% &      7.5$^{***}$ & 1.5\% & 12.4\% &     7.6$^{***}$ &  6.7\% &      6.9$^{***}$ & 13.3\% & 7.7\% &     7.4$^{***}$ \\
    8       &       3.0\% &      7.2$^{***}$ & 3.2\% & 3.1\% &     7.4$^{***}$ & 10.0\% &      7.5$^{***}$ & 0.9\% & 10.1\% &     7.6$^{***}$ &  6.8\% &      7.0$^{***}$ & 13.3\% & 7.8\% &     7.5$^{***}$ \\
    10      &          - &           - &    - &    - &          - &  8.0\% &      6.5$^{***}$ & 1.0\% &  8.2\% &     6.6$^{***}$ &  6.8\% &      7.1$^{***}$ & 12.7\% & 7.6\% &     7.5$^{***}$ \\
    15      &          - &           - &    - &    - &          - &  4.7\% &      4.3$^{***}$ & 0.4\% &  4.8\% &     4.4$^{***}$ &  7.1\% &      7.6$^{***}$ & 12.2\% & 7.9\% &     8.0$^{***}$ \\
    \midrule
    \midrule
    \multicolumn{16}{c}{OU+Thresh model}\\
    \midrule
     & \multicolumn{5}{c||}{Fama-French} & \multicolumn{5}{c||}{PCA} & \multicolumn{5}{c}{IPCA} \\
    \cmidrule{2-16}
    K & $\alpha$ & $t_\alpha$ & $R^2$ & $\mu$ & $t_\mu$ & $\alpha$ & $t_\alpha$ & $R^2$ & $\mu$ & $t_\mu$ & $\alpha$ & $t_\alpha$ & $R^2$ & $\mu$ & $t_\mu$ \\
    \midrule
    0       &      -4.5\% &        -1.4 & 13.4\% & -2.4\% &       -0.7 & -4.5\% &        -1.4 & 13.4\% & -2.4\% &       -0.7 & -4.5\% &        -1.4 & 13.4\% & -2.4\% &       -0.7 \\
    1       &      -0.2\% &        -0.2 & 13.5\% &  0.6\% &        0.6 &  0.7\% &         0.3 &  6.3\% &  2.1\% &        0.8 &  1.7\% &         1.4 & 18.9\% &  3.0\% &       2.3$^{*}$ \\
    3       &       0.9\% &         1.2 & 10.4\% &  1.6\% &       2.1$^{*}$ &  4.3\% &        2.5$^{*}$ &  4.3\% &  5.2\% &      3.0$^{**}$ &  2.6\% &       2.6$^{**}$ & 18.8\% &  3.8\% &     3.4$^{***}$ \\
    5       &       0.5\% &         0.9 &  6.8\% &  0.9\% &        1.5 &  3.7\% &        2.4$^{*}$ &  3.2\% &  4.4\% &      2.8$^{**}$ &  2.8\% &       3.0$^{**}$ & 17.7\% &  3.8\% &     3.8$^{***}$ \\
    8       &       0.6\% &         1.2 &  5.5\% &  1.0\% &        1.9 &  3.9\% &       3.0$^{**}$ &  1.9\% &  4.4\% &     3.4$^{***}$ &  2.3\% &       2.6$^{**}$ & 17.6\% &  3.5\% &     3.6$^{***}$ \\
    10      &          - &           - &     - &     - &          - &  2.6\% &        2.2$^{*}$ &  1.4\% &  2.9\% &       2.4$^{*}$ &  2.1\% &        2.5$^{*}$ & 17.6\% &  3.1\% &     3.3$^{***}$ \\
    15      &          - &           - &     - &     - &          - &  2.1\% &        2.1$^{*}$ &  0.7\% &  2.4\% &       2.4$^{*}$ &  2.3\% &       2.8$^{**}$ & 18.1\% &  3.2\% &     3.6$^{***}$ \\
    \bottomrule
    \end{tabular}
    \label{tab:deep-results-tests}
    }}
    \bnotetab{This table shows the out-of-sample pricing errors $\alpha$ of the arbitrage strategies relative of the Fama-French 8 factor model and their mean returns $\mu$ for the different arbitrage models and different number of factors $K$ that we use to obtain the residuals. We run a time-series regression of the out-of-sample returns of the arbitrage strategies on the 8-factor model (Fama-French 5 factors + momentum + short-term reversal + long-term reversal) and report the annualized $\alpha$, accompanying t-statistic value $t_\alpha$, and the $R^2$ of the regression. In addition, we report the annualized mean return $\mu$ along with its accompanying t-statistic $t_\mu$. The hypothesis test are two-sided and stars indicate p-values of 5\% ($^{*}$), 1\% ($^{**}$), and 0.1\% ($^{***}$). All results use the out-of-sample daily returns from January 2002 to December 2016 and the deep learning models are based on a Sharpe ratio objective.
    }
\end{table}

 \begin{figure}[t!]
 \caption{Cumulative OOS Returns of Different Arbitrage Strategies}
    \centering
    \begin{subfigure}[t]{.32\textwidth}
    \centering
    \includegraphics[width=1\linewidth]{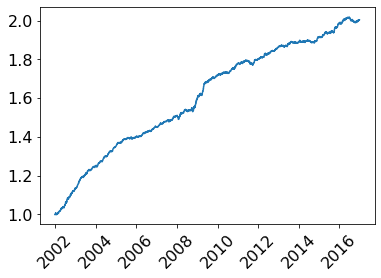}
    \vspace{-18pt}
    \caption{CNN+Trans, Fama-French 5}
    \end{subfigure}
    \vspace{-1pt}
    \begin{subfigure}[t]{.32\textwidth}
    \centering
    \includegraphics[width=1\linewidth]{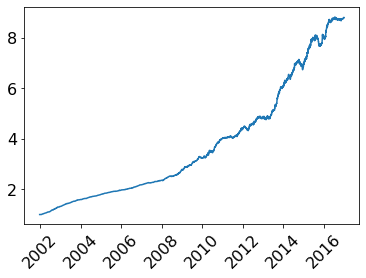} \vspace{-18pt}
    \caption{CNN+Trans, PCA 5}
    \end{subfigure}
    \vspace{-1pt}
    \begin{subfigure}[t]{.32\textwidth}
    \centering
    \includegraphics[width=1\linewidth]{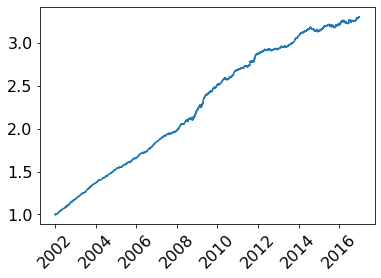} \vspace{-18pt}
    \caption{CNN+Trans, IPCA 5}
    \end{subfigure}
    \vspace{-1pt}
    
    \begin{subfigure}[t]{.32\textwidth}
    \centering
    \includegraphics[width=1\linewidth]{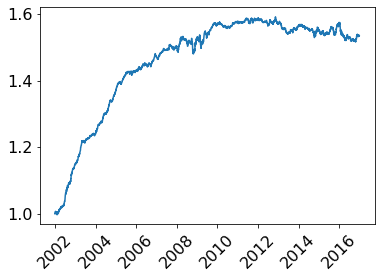} \vspace{-18pt}
    \caption{Fourier+FFN, Fama-French 5}
    \end{subfigure}
    \vspace{-1pt}
    \begin{subfigure}[t]{.32\textwidth}
    \centering
    \includegraphics[width=1\linewidth]{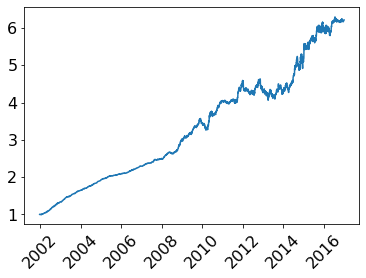} \vspace{-18pt}
    \caption{Fourier+FFN, PCA 5}
    \end{subfigure}
    \vspace{-1pt}
    \begin{subfigure}[t]{.32\textwidth}
    \centering
    \includegraphics[width=1\linewidth]{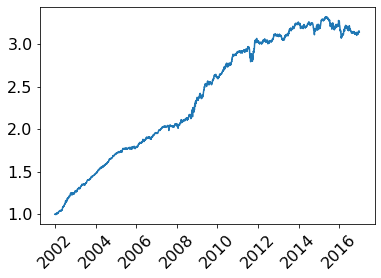} \vspace{-18pt}
    \caption{Fourier+FFN, IPCA 5}
    \end{subfigure}
    \vspace{-1pt}
    \begin{subfigure}[t]{.32\textwidth}
    \centering
    \includegraphics[width=1\linewidth]{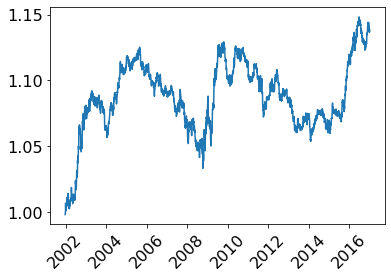}
    \vspace{-18pt}
    \caption{OU+Thresh Fama-French 5}
    \end{subfigure}
    \vspace{-1pt}
    \begin{subfigure}[t]{.32\textwidth}
    \centering
    \includegraphics[width=1\linewidth]{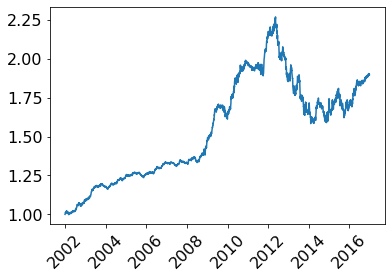}
    \vspace{-18pt}
    \caption{OU+Thresh PCA 5}
    \end{subfigure}
    \vspace{-1pt}
    \begin{subfigure}[t]{.32\textwidth}
    \centering
    \includegraphics[width=1\linewidth]{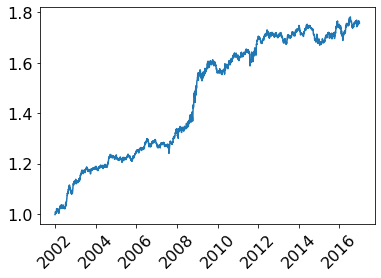}
    \vspace{-18pt}
    \caption{OU+Thresh IPCA 5}
    \end{subfigure}
    \vspace{-1pt}
    \label{fig:deep-cum-rets}
    \bnotefig{These figures show the cumulative daily returns of the arbitrage strategies for our representative models on the out-of-sample trading period between January 2002 and December 2016. We estimate the optimal arbitrage trading strategies for our three benchmark models based on the out-of-sample residuals of the Fama-French, PCA and IPCA 5-factor models. The deep learning models use the Sharpe ratio objective. 
}
\end{figure}

The returns of the CNN+Transformer arbitrage strategies are statistically significant and not subsumed by conventional risk factors. Table \ref{tab:deep-results-tests} reports the out-of-sample pricing errors $\alpha$ of the arbitrage strategies relative of the Fama-French 8 factor model and their mean returns $\mu$. We run a time-series regression of the out-of-sample returns of the arbitrage strategies on the 8-factor model (Fama-French 5 factors + momentum + short-term reversal + long-term reversal) and report the annualized $\alpha$, accompanying t-statistic value $t_\alpha$, and the $R^2$ of the regression. In addition, we report the annualized mean return $\mu$ along with its accompanying t-statistic $t_\mu$. The arbitrage strategies for the CNN+Transformer model for $K \geq 1$ are all statistically significant and not explained by the Fama-French 5 factors or price trend factors. Importantly, the pricing errors are essentially as large as the mean returns, which implies that the returns of the CNN+Transformer arbitrage strategies do not carry any risk premium of these eight factors. This is supported by the $R^2$ values, which are close to zero for the Fama-French or PCA residuals, and hence confirm that these arbitrage portfolios are essentially orthogonal to the Fama-French 8 factors. In contrast, one third of the individual return variation for $K=0$ is explained by those risk factors. However, even in that case the pricing errors are significant. The residuals of IPCA factors have a higher correlation with the Fama-French 8 factors, suggesting that the conditional IPCA factor model extracts factors that are inherently different from the conventional risk factors. Note that the residuals of a Fama-French 8 factor model are not mechanically orthogonal in the time-series regression on the Fama-French 8 factors, as we construct out-of-sample residuals based on rolling window estimates. The parametric arbitrage strategies are largely explained by conventional risk factors.  The third subtable in Table \ref{tab:deep-results-tests} shows that the residuals of the OU+Threshold model for Fama-French of PCA residuals do not have statistically significant pricing errors on a 1\% level.

The CNN+Transformer has a consistent out-of-sample performance and is not affected by various negative events. Figure \ref{fig:deep-cum-rets} shows the cumulative out-of-sample returns of the arbitrage strategies for our representative models. We select the residuals of the various five-factor models as including additional factors has only minor effects on the performance. Note that the CNN+Transformer model has consistently almost always positive returns, while maintaining a low volatility and avoiding any large losses. Importantly, the performance of the CNN+Transformer is nearly completely immune to both the ``quant quake'' which affected quantitative trading groups and funds engaging in statistical arbitrage in August 2007 (\cite{quant-quake-2007}), and the period of poor performance in quant funds during 2011--2012 (\cite{quant-quake-2011}). The Fourier+FFN model also performs similarly well until the financial crisis, but its risk increases afterwards as displayed by the larger volatility and larger drawdowns. The performance of the parametric model is visibly inferior. This illustrates that although all strategies trade the same residuals, which should be orthogonal to common market risk, profitable arbitrage trading requires an appropriate signal and allocation policy.

\subsection{Mean-Variance Objective}\label{sec:MVO}

\begin{table}[t!]
    \centering
    {\small
    \tcaptab{OOS Annualized Performance Based on Mean-Variance Objective}
    \scalebox{1.0}{
    \begin{tabular}{c||ccc||ccc||ccc}
    \toprule
    \multicolumn{10}{c}{CNN+Trans strategy, mean-variance objective function}\\
    \midrule
     & \multicolumn{3}{c||}{Fama-French} & \multicolumn{3}{c||}{PCA} & \multicolumn{3}{c}{IPCA}\\
    \cmidrule{2-10}
    
     K & SR &$\mu$ & $\sigma$ & SR & $\mu$& $\sigma$ & SR & $\mu$ &   $\sigma$ \\
    \midrule
    0  
    & 0.83 & 9.5\% & 11.4\%
    & 0.83 & 9.5\% & 11.4\%
    & 0.83 & 9.5\% & 11.4\%
    \\ 
    1  
    & 3.15 & 10.5\% & 3.3\%
    & 2.21 & 27.3\% & 12.3\%
    & 2.83 & 15.9\% & 5.6\%
    \\ 
    3  
    & 2.95 & 7.8\% & 2.6\%
    & 2.38 & 22.6\% & 9.5\%
    & 3.13 & 17.9\% & 5.7\%
    \\ 
    5  
    & 3.03 & 5.9\% & 2.0\%
    & 2.75 & 19.6\% & 7.1\%
    & 3.21 & 18.2\% & 5.7\%
    \\ 
    8
    & 2.96 &  4.2\% & 1.4\% 
    & 2.68 & 16.6\% & 6.2\%
    & 3.18 & 17.0\% & 5.4\%
    \\
    10  
     & - & - & -
    & 2.67 & 15.3\% & 5.7\%
    & 3.21 & 16.6\% & 5.2\%
    \\ 
    15  
     & - & - & -
    & 2.20 & 8.7\% & 4.0\%
    & 3.34 & 16.3\% & 4.9\%
    \\ 
    \midrule
    \midrule
    \multicolumn{10}{c}{Fourier+FFN strategy, mean-variance objective function}\\
    \midrule
    & \multicolumn{3}{c||}{Fama-French} & \multicolumn{3}{c||}{PCA} & \multicolumn{3}{c}{IPCA}\\
    \cmidrule{2-10}
     K &  SR & $\mu$ & $\sigma$ & SR &  $\mu$ & $\sigma$ & SR & $\mu$ & $\sigma$   \\
    \midrule
    0	
    & 0.28 & 5.5\% & 19.3\%	
    & 0.28 & 5.5\% & 19.3\%			
    & 0.28 & 5.5\% & 19.3\%			
    \\
    1	
    & 0.38 & 2.5\% & 6.7\%	
    & 0.48 & 16.6\% & 34.8\%
    & 0.56 & 9.7\% & 17.2\%		
    \\
    3		
    & 1.16 & 4.3\% & 3.7\%	
    & 0.34 & 32.1\% & 93.1\%
    & 1.06 & 17.6\% & 16.7\%		
    \\
    5	
    & 1.30 & 3.1\% & 2.4\%	
    & 0.37 & 22.5\% & 61.2\%
    & 1.17 & 17.0\% & 14.5\%		
    \\
    8 
    & 1.73 & 3.6\% & 2.0\%
    & 0.67 & 17.4\% & 25.9\%
    & 1.21 & 14.4\% & 11.9\%
    \\
    10	
    &	- & - & -	
    & 0.45 & 7.4\% & 16.4\%
    & 1.06 & 12.6\% & 11.9\%		
    \\
    15	
    &	- & - & -	
    & 0.56 & 5.7\% & 10.2\%
    & 1.17 & 12.1\% & 10.4\%		
    \\
    \bottomrule
    \end{tabular}}
    \label{tab:deep-mean-variance-results}
    }
    \bnotetab{This table shows the out-of-sample annualized Sharpe ratio (SR), mean return ($\mu$), and volatility ($\sigma$) of our CNN+Transformer and Fourier+FFN models for different numbers of risk factors $K$, that we use to obtain the residuals. We use a mean-variance objective function with risk aversion $\gamma=1$. We use the daily out-of-sample residuals from January 1998 to December 2016 and evaluate the out-of-sample arbitrage trading from January 2002 to December 2016. The two deep learning models are calibrated on a rolling window of four years. The signals are extracted from a rolling window of $L=30$ days. The $K=0$ factor model corresponds to directly using stock returns instead of residuals for the signal and trading policy.}
\end{table}

The deep learning statistical arbitrage strategies can achieve high average returns in spite of leverage constraints. Our main deep learning models are estimated with a Sharpe ratio objective. As the sum of absolute stock weights is normalized to one, the arbitrage strategies impose an implicit leverage constraint. We show that the average return can be increased while maintaining this leverage constraint. For this purpose we change the objective for the deep learning model to a mean-variance objective. In order to illustrate the effect of the different objective function, we set the risk aversion parameter to $\gamma=1$.

Tables \ref{tab:deep-mean-variance-results} and \ref{tab:deep-mean-variance-results-tests} collect the results for the Sharpe ratio, mean, volatility and significance tests. As expected the Sharpe ratios are slightly lower compared to the corresponding model with Sharpe ratio objective, but the mean returns are substantially increased. The CNN+Transformer model achieves average annual returns around 20\% with PCA and IPCA residuals while the volatility is only around half as large as the one of a market portfolio. The mean returns are statistically highly significant and not spanned by conventional risk factors or a price trend risk premium. The Fourier+FFN model can also obtain high average returns, but those come at the cost of a higher volatility. Overall, we confirm that the more flexible signal extraction function of the CNN+Transformer is crucial for the superior performance.

\begin{table}[t!]
    \centering
    {\footnotesize
    \tcaptab{Significance of Arbitrage Alphas based on Mean-Variance Objective}
    \setlength{\tabcolsep}{4pt} 
    \scalebox{0.9}{
    \begin{tabular}{c||ccccc||ccccc||ccccc}
    \toprule
    \multicolumn{16}{c}{CNN+Trans model}\\
    \midrule
     & \multicolumn{5}{c||}{Fama-French} & \multicolumn{5}{c||}{PCA} & \multicolumn{5}{c}{IPCA} \\
    \cmidrule{2-16}
    K & $\alpha$ & $t_\alpha$ & $R^2$ & $\mu$ & $t_\mu$ & $\alpha$ & $t_\alpha$ & $R^2$ & $\mu$ & $t_\mu$ & $\alpha$ & $t_\alpha$ & $R^2$ & $\mu$ & $t_\mu$ \\
    \midrule
    0       &       5.8\% &        2.2$^{*}$ & 19.6\% &  9.5\% &      3.2$^{**}$ &  5.8\% &        2.2$^{*}$ & 19.6\% &  9.5\% &      3.2$^{**}$ &  5.8\% &        2.2$^{*}$ & 19.6\% &  9.5\% &      3.2$^{**}$ \\
    1       &       9.9\% &       12$^{***}$ &  7.1\% & 10.5\% &      12$^{***}$ & 26.3\% &      8.3$^{***}$ &  1.6\% & 27.3\% &     8.6$^{***}$ & 14.0\% &       11$^{***}$ & 23.5\% & 15.9\% &      11$^{***}$ \\
    3       &       7.5\% &       11$^{***}$ &  5.3\% &  7.8\% &      11$^{***}$ & 22.1\% &      9.1$^{***}$ &  2.2\% & 22.6\% &     9.2$^{***}$ & 16.6\% &       12$^{***}$ & 17.6\% & 17.9\% &      12$^{***}$ \\
    5       &       5.7\% &       11$^{***}$ &  5.3\% &  5.9\% &      12$^{***}$ & 19.0\% &       10$^{***}$ &  3.2\% & 19.6\% &      11$^{***}$ & 16.7\% &       12$^{***}$ & 16.0\% & 18.2\% &      12$^{***}$ \\
    8       &       4.4\% &      9.8$^{***}$ &  3.6\% &  4.6\% &      10$^{***}$ & 16.3\% &       10$^{***}$ &  1.6\% & 16.6\% &      10$^{***}$ & 15.5\% &       12$^{***}$ & 18.3\% & 17.0\% &      12$^{***}$ \\
    10      &          - &           - &     - &     - &          - & 14.8\% &       10$^{***}$ &  1.7\% & 15.3\% &      10$^{***}$ & 15.2\% &       13$^{***}$ & 20.6\% & 16.6\% &      12$^{***}$ \\
    15      &          - &           - &     - &     - &          - &  8.5\% &      8.4$^{***}$ &  0.9\% &  8.7\% &     8.5$^{***}$ & 14.8\% &       13$^{***}$ & 21.6\% & 16.3\% &      13$^{***}$ \\
    \midrule
    \midrule
    \multicolumn{16}{c}{Fourier+FFN model}\\
    \midrule
     & \multicolumn{5}{c||}{Fama-French} & \multicolumn{5}{c||}{PCA} & \multicolumn{5}{c}{IPCA} \\
    \cmidrule{2-16}
    K & $\alpha$ & $t_\alpha$ & $R^2$ & $\mu$ & $t_\mu$ & $\alpha$ & $t_\alpha$ & $R^2$ & $\mu$ & $t_\mu$ & $\alpha$ & $t_\alpha$ & $R^2$ & $\mu$ & $t_\mu$ \\
    \midrule
    0       &       3.2\% &         0.7 & 8.4\% & 5.5\% &        1.1 &  3.2\% &         0.7 & 8.4\% &  5.5\% &        1.1 &  3.2\% &         0.7 & 8.4\% &  5.5\% &        1.1 \\
    1       &       2.8\% &         1.6 & 1.8\% & 2.5\% &        1.5 & 15.4\% &         1.7 & 1.3\% & 16.6\% &        1.9 &  7.9\% &         1.8 & 2.6\% &  9.7\% &       2.2$^{*}$ \\
    3       &       4.1\% &      4.4$^{***}$ & 3.4\% & 4.3\% &     4.5$^{***}$ & 30.3\% &         1.3 & 0.1\% & 32.1\% &        1.3 & 17.4\% &      4.1$^{***}$ & 1.9\% & 17.6\% &     4.1$^{***}$ \\
    5       &       2.9\% &      4.8$^{***}$ & 3.1\% & 3.1\% &     5.0$^{***}$ & 21.0\% &         1.3 & 0.1\% & 22.5\% &        1.4 & 15.9\% &      4.3$^{***}$ & 2.6\% & 17.0\% &     4.5$^{***}$ \\
    8       &       3.5\% &      6.8$^{***}$ & 2.3\% & 3.6\% &     7.0$^{***}$ & 17.4\% &       2.6$^{**}$ & 0.3\% & 17.2\% &      2.6$^{**}$ & 12.9\% &      4.3$^{***}$ & 4.4\% & 14.4\% &     4.7$^{***}$ \\
    10      &          - &           - &    - &    - &          - &  7.1\% &         1.7 & 0.3\% &  7.4\% &        1.8 & 11.7\% &      3.9$^{***}$ & 3.5\% & 12.6\% &     4.1$^{***}$ \\
    15      &          - &           - &    - &    - &          - &  5.5\% &        2.1$^{*}$ & 0.1\% &  5.7\% &       2.2$^{*}$ & 11.3\% &      4.3$^{***}$ & 4.0\% & 12.1\% &     4.5$^{***}$ \\
    \bottomrule
    \end{tabular}
    \label{tab:deep-mean-variance-results-tests}
    }}
    \bnotetab{This table shows the out-of-sample pricing errors $\alpha$ of the arbitrage strategies relative of the Fama-French 8 factor model and their mean returns $\mu$ for the different arbitrage models and different number of factors $K$ that we use to obtain the residuals. We use a mean-variance objective function with risk aversion $\gamma=1$. We run a time-series regression of the out-of-sample returns of the arbitrage strategies on the 8-factor model (Fama-French 5 factors + momentum + short-term reversal + long-term reversal) and report the annualized $\alpha$, accompanying t-statistic value $t_\alpha$, and the $R^2$ of the regression. In addition, we report the annualized mean return $\mu$ along with its accompanying t-statistic $t_\mu$. The hypothesis test are two-sided and stars indicate p-values of 5\% ($^{*}$), 1\% ($^{**}$), and 0.1\% ($^{***}$). All results use the out-of-sample daily returns from January 2002 to December 2016.
    }
\end{table}

\subsection{Unconditional Residual Means}

The unconditional average of residuals is not a profitable strategy and does not provide information about the potential arbitrage profitability contained in the residuals. A natural question to ask is if the residuals themselves have a risk premium component and if trading an equally weighted portfolio of residuals could be profitable. Table \ref{tab:unconditional-mean-results} in the Appendix shows the performance of this simple strategy. If we do not project out any factors ($K=0$), this strategy essentially trades an equally weighted market portfolio. Table \ref{tab:unconditional-mean-results-tests} in the Appendix reports the test statistics relative to the Fama-French 8 factor model, which completely subsumes the market risk premium. Once we regress out at least 3 factors, the equally weighted residuals have a mean return of around 1\% or lower. The low volatility confirms that the residuals are only weakly cross-sectionally dependent and are largely diversified away. The moderately large Sharpe ratios for PCA residuals is a consequence of the near zero volatility. Scaling up the mean returns to a meaningful magnitude would potentially require an unreasonable amount of leverage. Overall, we confirm that residuals need to be combined with a signal and trading policy that takes advantage of the time series patterns in order to achieve a profitable strategy.

IPCA factors are close to uncorrelated with conventional risk factors. The $R^2$ values in Table \ref{tab:unconditional-mean-results-tests} are as expected for the Fama-French factors and, not surprisingly, after regressing out all of those factors, the cross-sectional average of the residuals is essentially orthogonal to those factors. The PCA residuals show a very similar behavior. However, the conditional IPCA model leaves a component in the residuals that it is highly correlated with conventional risk factors. In this sense, the IPCA factors extract a factor model that is quite different from the Fama-French factors. 

Importantly, unconditional means and alphas of asset pricing residuals are a poor measure of arbitrage opportunities. The mean and alphas of residuals that are optimally traded based on their time series patterns have mean returns that can be larger by a factor of 50. This implies more generally, that the unconditional perspective of evaluating asset pricing models could potentially overstate the efficiency of markets and the pricing ability of asset pricing models.

\subsection{Importance of Time-Series Signal}

How important is the flexibility in the signal extraction function relative to the allocation function? So far, we have considered a rigid parametric model for the signal and allocation function and a flexible allocation function but either a pre-specified time-series filter or a data-driven flexible filter. In \ref{tab:aux-model-results} in Appendix we also report the results for two additional model variations, which serve as ablation tests emphasizing the central importance of applying a time-series model to extract a signal extraction function from the data. 

The first model, OU+FFN, uses the same 4-dimensional OU signal as the OU+Threshold policy, but replaces the threshold allocation function with an FFN allocation function. This FFN allocation function has the same architecture as that of the Fourier+FFN policy, except the input is 4-dimensional instead of 30-dimensional. The results show that even despite using a very flexible allocation function, the results are similar or even worse than the simple parametric thresholding rule.  This points to the weakness of the OU signal representation: although the allocation function is a powerful universal approximator, it cannot accomplish much with an information-poor input. If the optimal allocation function given the simple OU signal is well approximated by the parametric thresholding rule, then the nonparametric FFN offers too much flexibility without comparable efficiency, which leads to a noisier estimate of a simple function and hence worse out-of-sample performance. 

The second model does not extract a time-series signal from the residuals, but uses the residuals themselves as signal to a flexible FFN allocation function. As the allocation function uses the same type of network as for the CNN+Transformer, Fourier+FFN or OU+FFN, this setup directly assesses the relevance of using a time-series model for the signal. The FFN model also performs worse than the deep learning models that apply a time-series filter to the residuals. This is a good example to emphasize the importance of a time-series model. While FFNs are flexible in learning low dimensional functional relationships, they are limited in learning a complex dependency model if the training data is limited. For example, the FFN is not sufficiently efficient to learn an FFT-like transformation and hence has a substantially worse performance on the original time-series compared to frequency-transformed time-series.

In summary, the flexible data-driven signal extraction function of the CNN+Transformer model seems to be the critical element for statistical arbitrage. A flexible allocation function is not sufficient to compensate for an uninformative signal.

\subsection{Dependency between Arbitrage Strategies}

The trading strategies for different factor models are only weakly correlated. In Table \ref{tab:corralt}, we report the correlations of the returns of our CNN+Transformer strategies across factor models with 3, 5, and 10 factors, based on the Sharpe ratio objective function strategy.\footnote{The correlations for the mean-variance objective function are similar.} Notably, the correlations between strategies from different factor model families range from roughly 0.2 to 0.45, indicating that strategies for different factor model families can be used as part of a diversification strategy. While the performance of the arbitrage trading for the residuals obtained with different families of factor models is comparable, the factors themselves are different. Hence, even if the arbitrage signal and allocation functions are similar, the resulting strategies can be weakly correlated. The within-family correlations range from 0.4 to 0.85, indicating that the residuals from the same class of factor model capture similar patterns.

\begin{table}[t!]
    \centering
    {\small
    \tcaptab{OOS Annualized Performance of CNN+Trans for 60 Days Lookback Window}
    \begin{tabular}{c||ccc||ccc||ccc}
    \toprule
     & \multicolumn{3}{c||}{Fama-French} 
     & \multicolumn{3}{c||}{PCA} 
     & \multicolumn{3}{c}{IPCA}\\
    \cmidrule{2-10}
    
    
     K &  SR &$\mu$ & $\sigma$ & SR &$\mu$ & $\sigma$ & SR &$\mu$ & $\sigma$  \\
    \midrule
    0  
    & 1.50 & 13.5\% & 9.0\%
    & 1.50 & 13.5\% & 9.0\%
    & 1.50 & 13.5\% & 9.0\%
    \\ 
    
    1  
    & 2.95 & 9.6\% & 3.2\%
    & 2.68 & 15.8\% & 5.9\%
    & 3.14 & 8.8\% & 2.8\%
    \\ 
    
    3  
    & 3.21 & 8.7\% & 2.7\%
    & 3.49 & 16.8\% & 4.8\%
    & 3.84 & 9.6\% & 2.5\%
    \\ 
    
    5  
    & 3.23 & 6.8\% & 2.1\%
    & 3.54 & 16.0\% & 4.5\%
    & 3.90 & 9.2\% & 2.4\%
    \\ 
    
    8 
    & 2.96 &  4.2\% &  1.4\% 
    & 3.02 & 12.5\% & 4.2\%
    & 3.93 & 8.7\% & 2.2\%
    \\
    
    10  
     & - & - & -
    & 2.67 & 9.9\% & 3.7\%
    & 3.98 & 9.2\% & 2.3\%
    \\ 
    
    15  
     & - & - & -
    & 2.36 & 8.1\% & 3.4\%
    & 4.24 & 9.6\% & 2.3\%
    \\ 
    \bottomrule
    \end{tabular}\label{tab:deep-results60lookback}}
\bnotetab{This table shows the out-of-sample annualized Sharpe ratio (SR), mean return ($\mu$), and volatility ($\sigma$) of the CNN+Transformer model for different numbers of risk factors $K$, that we use to obtain the residuals. The signals are extracted from a rolling window of $L=60$ days. We use the daily out-of-sample residuals from January 1998 to December 2016 and evaluate the out-of-sample arbitrage trading from January 2002 to December 2016. The model is calibrated on a rolling window of four years and uses the Sharpe ratio objective function. The $K=0$ factor model corresponds to directly using stock returns instead of residuals for the signal and trading policy. 
}
\end{table}

\subsection{Stability over Time}\label{sec:time-stability}

Our results are robust to length of the local window to extract the trading signal. We re-estimate the CNN+Transformer model on an extended rolling lookback window of $L=60$ days, while keeping the rest of the model structure the same. Tables \ref{tab:deep-results60lookback} and \ref{tab:deep-results60lookback-tests} show that the results are robust to the choice of lookback window. Extending the local window to 60 trading days, which is close to three months, leads to essentially the same performance as using only the most recent $L=30$ trading days to infer the signal. This is further evidence that the arbitrage signal is different from conventional momentum or reversal strategies that incorporate information from longer time periods. As the signal can be inferred from the most recent past, it implies that either the arbitrage signal depends only on the most recent days or that those days are sufficient to infer the relevant time-series structure. In the next section, we provide evidence that the arbitrage trading signals put strong emphasis on most recent two weeks prior to the trading.

\begin{table}[t!]
    \centering
    {\footnotesize
    \tcaptab{Significance of Arbitrage Alphas for 60 Days Lookback Window}
    \setlength{\tabcolsep}{4pt} 
    \scalebox{0.9}{
    \begin{tabular}{c||ccccc||ccccc||ccccc}
    \toprule
    \multicolumn{16}{c}{CNN+Trans Model , Sharpe objective function, $L=60$ days lookback window}\\
    \midrule
     & \multicolumn{5}{c||}{Fama-French} & \multicolumn{5}{c||}{PCA} & \multicolumn{5}{c}{IPCA} \\
    \cmidrule{2-16}
    K & $\alpha$ & $t_\alpha$ & $R^2$ & $\mu$ & $t_\mu$ & $\alpha$ & $t_\alpha$ & $R^2$ & $\mu$ & $t_\mu$ & $\alpha$ & $t_\alpha$ & $R^2$ & $\mu$ & $t_\mu$ \\
    \midrule
    0       &      11.8\% &      5.6$^{***}$ & 19.5\% & 13.5\% &     5.8$^{***}$ & 11.8\% &      5.6$^{***}$ & 19.5\% & 13.5\% &     5.8$^{***}$ & 11.8\% &      5.6$^{***}$ & 19.5\% & 13.5\% &     5.8$^{***}$ \\
    1       &       9.1\% &       11$^{***}$ &  7.2\% &  9.6\% &      11$^{***}$ & 15.5\% &       10$^{***}$ &  1.2\% & 15.8\% &      10$^{***}$ &  8.2\% &       12$^{***}$ & 10.1\% &  8.8\% &      12$^{***}$ \\
    3       &       8.3\% &       12$^{***}$ &  7.1\% &  8.7\% &      12$^{***}$ & 16.5\% &       13$^{***}$ &  2.5\% & 16.8\% &      14$^{***}$ &  9.2\% &       15$^{***}$ &  9.3\% &  9.6\% &      15$^{***}$ \\
    5       &       6.5\% &       12$^{***}$ &  6.0\% &  6.8\% &      13$^{***}$ & 15.6\% &       13$^{***}$ &  2.2\% & 16.0\% &      14$^{***}$ &  8.8\% &       15$^{***}$ & 10.3\% &  9.2\% &      15$^{***}$ \\
    8       &       4.1\% &       11$^{***}$ &  3.2\% &  4.2\% &      11$^{***}$ & 12.2\% &       11$^{***}$ &  1.6\% & 12.5\% &      12$^{***}$ &  8.3\% &       15$^{***}$ &  8.9\% &  8.7\% &      15$^{***}$ \\
    10      &          - &           - &     - &     - &          - &  9.7\% &       10$^{***}$ &  1.0\% &  9.9\% &      10$^{***}$ &  8.8\% &       15$^{***}$ &  8.3\% &  9.2\% &      15$^{***}$ \\
    15      &          - &           - &     - &     - &          - &  8.1\% &      9.1$^{***}$ &  0.7\% &  8.1\% &     9.1$^{***}$ &  9.2\% &       16$^{***}$ &  9.3\% &  9.6\% &      16$^{***}$ \\
    \bottomrule
    \end{tabular}
    \label{tab:deep-results60lookback-tests}
    }}
    \bnotetab{This table shows the out-of-sample pricing errors $\alpha$ of the arbitrage strategies relative of the Fama-French 8 factor model and their mean returns $\mu$ for the CNN+Transformer model and different number of factors $K$ that we use to obtain the residuals. The signals are extracted from a rolling window of $L=60$ days. We run a time-series regression of the out-of-sample returns of the arbitrage strategies on the 8-factor model (Fama-French 5 factors + momentum + short-term reversal + long-term reversal) and report the annualized $\alpha$, accompanying t-statistic value $t_\alpha$, and the $R^2$ of the regression. In addition, we report the annualized mean return $\mu$ along with its accompanying t-statistic $t_\mu$. The hypothesis test are two-sided and stars indicate p-values of 5\% ($^{*}$), 1\% ($^{**}$), and 0.1\% ($^{***}$). All results use the out-of-sample daily returns from January 2002 to December 2016 and are based on a Sharpe ratio objective.
    }
\end{table}

\begin{table}[t!]
    \centering
    {\small
    \tcaptab{OOS Annualized Performance of CNN+Trans for Constant Model}
    \label{tab:deep-resultsNoRetrain}
    \scalebox{1}{
    \begin{tabular}{c||ccc||ccc||ccc}
    \toprule
    \multicolumn{10}{c}{$T_{\text{train}}$ = 4 years}\\
    \midrule
     & \multicolumn{3}{c||}{Fama-French} 
     & \multicolumn{3}{c||}{PCA} 
     & \multicolumn{3}{c}{IPCA}\\
    \cmidrule{2-10}    
     K 
     & SR &$\mu$ & $\sigma$
     & SR &$\mu$ & $\sigma$
     & SR &$\mu$ & $\sigma$   \\
    \midrule
    0  
    & 1.10 & 8.5\% & 7.8\%
    & 1.10 & 8.5\% & 7.8\%
    & 1.10 & 8.5\% & 7.8\%
    \\ 
    
    1  
    & 1.90 & 4.5\% & 2.3\%
    & 0.66 & 5.2\% & 7.9\%
    & 0.94 & 3.1\% & 3.3\%
    \\ 
    
    3  
    & 1.60 & 3.6\% & 2.2\%
    & 1.65 & 8.7\% & 5.3\%
    & 1.82 & 5.3\% & 2.9\%
    \\ 
    
    5  
    & 1.81 & 3.0\% & 1.7\%
    & 1.93 & 9.8\% & 5.1\%
    & 2.09 & 5.4\% & 2.6\%
    \\ 
    
    8       
    & 1.70 & 2.5\% & 1.5\% 
    & 2.04 & 9.6\% & 4.7\%
    & 1.89 & 5.0\% & 2.6\%
    \\
    
    10  
    & - & - & -
    & 2.06 & 9.1\% & 4.4\%
    & 1.77 & 4.7\% & 2.7\%
    \\ 
    
    15  
    & - & - & -
    & 1.82 & 7.0\% & 3.9\%
    & 2.09 & 5.5\% & 2.7\%
    \\ 
    \midrule
    \midrule
    \multicolumn{10}{c}{$T_{\text{train}}$ = 8 years}\\
    \midrule
     & \multicolumn{3}{c||}{Fama-French} 
     & \multicolumn{3}{c||}{PCA} 
     & \multicolumn{3}{c}{IPCA}\\
    \cmidrule{2-10}
     K 
     & SR &$\mu$ & $\sigma$ 
     & SR &$\mu$ & $\sigma$ 
     & SR &$\mu$ & $\sigma$   \\
    \midrule
    0  
    & 1.33 & 12.0\% & 9.0\%
    & 1.33 & 12.0\% & 9.0\%
    & 1.33 & 12.0\% & 9.0\%
    \\ 
    
    1  
    & 2.06 & 5.0\% & 2.4\%
    & 1.81 & 15.2\% & 8.4\%
    & 2.02 & 8.5\% & 4.2\%
    \\ 
    
    3  
    & 2.46 & 5.3\% & 2.2\%
    & 2.04 & 13.1\% & 6.4\%
    & 2.47 & 7.5\% & 3.0\%
    \\ 
    
    5  
    & 1.82 & 3.2\% & 1.8\%
    & 1.91 & 11.9\% & 6.2\%
    & 2.64 & 7.6\% & 2.9\%
    \\ 
    
    8       
    & 1.48 &  2.5\% & 1.7\% 
    & 1.89 & 10.8\% & 5.7\%
    & 2.71 &  8.3\% & 3.1\%
    \\
    
    10  
    & - & - & -
    & 1.82 & 10.0\% & 5.5\%
    & 2.68 & 8.2\% & 3.1\%
    \\ 
    
    15  
    & - & - & -
    & 1.38 & 6.2\% & 4.5\%
    & 2.70 & 7.8\% & 2.9\%
    \\ 
    \bottomrule
    \end{tabular}
    }}
\bnotetab{This table shows the out-of-sample annualized Sharpe ratio (SR), mean return ($\mu$), and volatility ($\sigma$) of the CNN+Transformer model for different numbers of risk factors $K$. We estimate the model on only once on the first $T_{\text{train}}$ days and keep it constant on the remaining test set. We use the daily out-of-sample residuals from January 1998 to December 2016 and evaluate the out-of-sample arbitrage trading from January 1998 $+T_{\text{train}}$ to December 2016. The signals are extracted from a rolling window of $L=30$ days and we use the Sharpe ratio objective function. 
}
\end{table}

\begin{table}[t!]
\centering
    {\footnotesize
    \tcaptab{Significance of Arbitrage Alphas for Constant Model}
    \setlength{\tabcolsep}{4pt} 
    \scalebox{0.9}{
    \begin{tabular}{c||ccccc||ccccc||ccccc}
    \toprule
    \multicolumn{16}{c}{CNN+Trans model, Sharpe objective function, $T_{\text{train}}=4$ years}\\
    \midrule
     & \multicolumn{5}{c||}{Fama-French} & \multicolumn{5}{c||}{PCA} & \multicolumn{5}{c}{IPCA} \\
    \cmidrule{2-16}
    K & $\alpha$ & $t_\alpha$ & $R^2$ & $\mu$ & $t_\mu$ & $\alpha$ & $t_\alpha$ & $R^2$ & $\mu$ & $t_\mu$ & $\alpha$ & $t_\alpha$ & $R^2$ & $\mu$ & $t_\mu$ \\
    \midrule
    0       &       8.4\% &      4.2$^{***}$ & 3.0\% & 8.5\% &     4.3$^{***}$ &  8.4\% &      4.2$^{***}$ & 3.0\% & 8.5\% &     4.3$^{***}$ &  8.4\% &      4.2$^{***}$ &  3.0\% & 8.5\% &     4.3$^{***}$ \\
    1       &       4.0\% &      6.8$^{***}$ & 5.9\% & 4.5\% &     7.3$^{***}$ &  4.1\% &        2.0$^{*}$ & 4.5\% & 5.2\% &       2.5$^{*}$ &  3.1\% &      3.7$^{***}$ &  1.6\% & 3.1\% &     3.6$^{***}$ \\
    3       &       3.2\% &      5.7$^{***}$ & 4.9\% & 3.6\% &     6.2$^{***}$ &  8.2\% &      6.1$^{***}$ & 2.7\% & 8.7\% &     6.4$^{***}$ &  5.3\% &      7.4$^{***}$ & 11.7\% & 5.3\% &     7.0$^{***}$ \\
    5       &       2.8\% &      6.6$^{***}$ & 4.3\% & 3.0\% &     7.0$^{***}$ &  9.3\% &      7.1$^{***}$ & 1.8\% & 9.8\% &     7.5$^{***}$ &  5.5\% &      8.6$^{***}$ &  8.3\% & 5.4\% &     8.1$^{***}$ \\
    8       &       2.3\% &      6.1$^{***}$ & 5.1\% & 2.5\% &     6.6$^{***}$ &  9.0\% &      7.5$^{***}$ & 2.2\% & 9.6\% &     7.9$^{***}$ &  5.0\% &      7.7$^{***}$ &  8.2\% & 5.0\% &     7.3$^{***}$ \\
    10      &          - &           - &    - &    - &          - &  8.6\% &      7.5$^{***}$ & 1.9\% & 9.1\% &     8.0$^{***}$ &  5.1\% &      8.0$^{***}$ & 16.6\% & 4.7\% &     6.9$^{***}$ \\
    15      &          - &           - &    - &    - &          - &  6.8\% &      6.8$^{***}$ & 1.0\% & 7.0\% &     7.1$^{***}$ &  5.8\% &      9.3$^{***}$ & 17.6\% & 5.5\% &     8.1$^{***}$ \\
    \midrule
    \midrule
    \multicolumn{16}{c}{CNN+Trans model, Sharpe objective function, $T_{\text{train}}=8$ years}\\
    \midrule
     & \multicolumn{5}{c||}{Fama-French} & \multicolumn{5}{c||}{PCA} & \multicolumn{5}{c}{IPCA} \\
    \cmidrule{2-16}
    K & $\alpha$ & $t_\alpha$ & $R^2$ & $\mu$ & $t_\mu$ & $\alpha$ & $t_\alpha$ & $R^2$ & $\mu$ & $t_\mu$ & $\alpha$ & $t_\alpha$ & $R^2$ & $\mu$ & $t_\mu$ \\
    \midrule
    0       &      10.1\% &      4.1$^{***}$ & 18.1\% & 12.0\% &     4.4$^{***}$ & 10.1\% &      4.1$^{***}$ & 18.1\% & 12.0\% &     4.4$^{***}$ & 10.1\% &      4.1$^{***}$ & 18.1\% & 12.0\% &     4.4$^{***}$ \\
    1       &       4.4\% &      6.5$^{***}$ & 14.3\% &  5.0\% &     6.8$^{***}$ & 14.5\% &      5.8$^{***}$ &  2.5\% & 15.2\% &     6.0$^{***}$ &  7.0\% &      6.6$^{***}$ & 30.6\% &  8.5\% &     6.7$^{***}$ \\
    3       &       4.9\% &      7.9$^{***}$ & 11.6\% &  5.3\% &     8.2$^{***}$ & 12.8\% &      6.7$^{***}$ &  2.7\% & 13.1\% &     6.8$^{***}$ &  7.0\% &      7.9$^{***}$ &  8.2\% &  7.5\% &     8.2$^{***}$ \\
    5       &       2.9\% &      5.8$^{***}$ & 12.3\% &  3.2\% &     6.0$^{***}$ & 11.6\% &      6.2$^{***}$ &  1.6\% & 11.9\% &     6.3$^{***}$ &  7.1\% &      8.7$^{***}$ & 12.1\% &  7.6\% &     8.7$^{***}$ \\
    8       &       2.3\% &      4.7$^{***}$ &  5.4\% &  2.5\% &     4.9$^{***}$ & 10.2\% &      6.0$^{***}$ &  3.1\% & 10.8\% &     6.3$^{***}$ &  7.7\% &      9.0$^{***}$ & 14.6\% &  8.3\% &     9.0$^{***}$ \\
    10      &          - &           - &     - &     - &          - &  9.4\% &      5.7$^{***}$ &  2.6\% & 10.0\% &     6.0$^{***}$ &  7.7\% &      8.9$^{***}$ & 11.3\% &  8.2\% &     8.9$^{***}$ \\
    15      &          - &           - &     - &     - &          - &  6.0\% &      4.4$^{***}$ &  0.9\% &  6.2\% &     4.6$^{***}$ &  7.4\% &      8.9$^{***}$ & 11.2\% &  7.8\% &     8.9$^{***}$ \\
    \bottomrule
    \end{tabular}
    \label{tab:deep-resultsNoRetrain-tests}
    }}
    \bnotetab{This table shows the out-of-sample pricing errors $\alpha$ of the arbitrage strategies relative of the Fama-French 8 factor model and their mean returns $\mu$ for the CNN+Transformer model and different number of factors $K$. We estimate the model on only once on the first $T_{\text{train}}$ days and keep it constant on the remaining test set. We use the daily out-of-sample residuals from January 1998 to December 2016 and evaluate the out-of-sample arbitrage trading from January 1998 $+T_{\text{train}}$ to December 2016. The signals are extracted from a rolling window of $L=30$ days and we use the Sharpe ratio objective function. We run a time-series regression of the out-of-sample returns of the arbitrage strategies on the 8-factor model (Fama-French 5 factors + momentum + short-term reversal + long-term reversal) and report the annualized $\alpha$, accompanying t-statistic value $t_\alpha$, and the $R^2$ of the regression. In addition, we report the annualized mean return $\mu$ along with its accompanying t-statistic $t_\mu$. The hypothesis test are two-sided and stars indicate p-values of 5\% ($^{*}$), 1\% ($^{**}$), and 0.1\% ($^{***}$).
 }
\end{table}

A constant-in-time signal and allocation function captures a large fraction of the arbitrage information. We re-estimate the CNN+Transformer model with a constant model instead of the rolling window calibration. Our main models are estimated on a rolling window of four years, which allows the models to adopt to changing economic conditions. Here we use either the first $T_{\text{train}}=4$ years (1,000 trading days) or $T_{\text{train}}=8$ years (2,000 trading days) to estimate the signal and allocation function, and then keep those functions constant for the remaining out-of-sample trading period. The results are reported in Tables \ref{tab:deep-resultsNoRetrain} and \ref{tab:deep-resultsNoRetrain-tests}. As expected the performance decreases relative to a time-varying model with re-estimation, which suggests that there is some degree of time-variation in the signal and allocation function. The longer training window of 8 years results in slightly higher Sharpe ratios than the 4 year window, as the model has more data and more variety in the market environment to learn the arbitrage information. Importantly, the constant CNN+Transformer still substantially outperforms the other benchmark models, Fourier+FFN and OU+Threshold, even if those are estimated on a rolling window. We conclude that the constant signal and allocation function for the CNN+Transformer model already capture a substantial amount of arbitrage information. Therefore, the constant functions serve as a meaningful model to analyze in more detail in Section \ref{sec:interpretation}.

\subsection{Market Frictions and Transaction Costs}\label{sec:market-frictions-results}

Our deep learning arbitrage strategies remain profitable in the presence of realistic trading frictions. In practice, trading costs associated with high turnover or large short-selling positions can diminish the profitability of arbitrage trading. In order to ensure that our model produces economically meaningful results, we extend it to the setting in which both transaction costs and holding costs are accounted for. We do not model market frictions associated with market impact, as in our empirical analysis we restrict the asset universe to stocks with large market capitalization, which are especially liquid.

In our market-friction extension, the daily returns $R_t$ of the strategy now have constant linear penalties associated with the daily turnover and the proportion of short trades. These penalties quantify proportional transaction costs, which are used to model trading fees, size of the bid-ask spread, etc., and holding costs, which are used to model short borrow rate fees charged by a brokerage. In particular, we incorporate a subset of the market friction models proposed by \cite{boyd-mpo}, which are commonly used in the statistical arbitrage literature.\footnote{See for example \cite{avelee}, \cite{papayeo} and \cite{Krauss}.} Mathematically, we subtract the market-friction costs
\begin{equation*}
\text{cost}(\wret_{t-1},\wret_{t-2}) = 0.0005 \|\wret_{t-1}-\wret_{t-2}\|_{L^1}+0.0001 \|\min(\wret_{t-1},0)\|_{L^1}
\end{equation*}
from the portfolio returns and use these net portfolio returns in the optimization problem of section \ref{sec:arbitragetrading}, where  $\wret_{t-1}\in\R^{N_{t-1}}$ is the strategy's allocation weight vector at time $t-1$. The first penalty term represents a slippage/transaction cost of 5 basis points per transaction, whereas the second one is a holding cost of 1 basis point per short position. Both costs are universal for all times and all stocks. This corresponds to a modification of the objective function in the training and evaluation parts of our algorithm. We use this model for the sake of illustration and simplicity given that in our empirical study we trade a universe of highly liquid US stocks, but more complicated models\footnote{For example, those considering time and stock-dependent transaction costs or market impact of the trades on the stock prices.} may be included in the computations without any significant structural changes.

\begin{table}[t!]
\centering
{\small
\tcaptab{OOS Performance of CNN+Trans with Trading Frictions}
\label{tab:deep-resultsFrictions}
\scalebox{0.95}{
\begin{tabular}{c||ccc||ccc}
\toprule
\multicolumn{7}{c}{IPCA factor model}\\
\midrule
  & \multicolumn{3}{c||}{Sharpe ratio} & \multicolumn{3}{c}{Mean-variance}\\
\cmidrule{2-7}


 K & SR &$\mu$ & $\sigma$ & SR &$\mu$ & $\sigma$  \\
\midrule
0  
& 0.52 & 8.5\% & 16.3\%
& 0.22 & 2.6\% & 11.9\%
\\ 

1  
& 0.85 & 5.9\% & 6.9\%
& 0.86 & 5.5\% & 6.4\%
\\ 

3  
& 1.24 & 6.6\% & 5.4\%
& 1.16 & 6.9\% & 5.9\%
\\ 

5  
& 1.11 & 5.5\% & 5.0\%
& 1.02 & 5.3\% & 5.3\%
\\ 

10  
& 0.98 & 5.1\% & 5.2\%
& 1.04 & 5.4\% & 5.2\%
\\ 

15  
& 0.94 & 4.8\% & 5.1\%
& 1.02 & 5.1\% & 5.0\%
\\ 
\bottomrule
\end{tabular}}
}
    \bnotetab{This table shows the out-of-sample annualized Sharpe ratio (SR), mean return ($\mu$), and volatility ($\sigma$) for the CNN+Transformer model with trading frictions on IPCA residuals. We use the daily out-of-sample residuals from January 1998 to December 2016 and evaluate the out-of-sample arbitrage trading from January 2002 to December 2016. The models are calibrated on a rolling window of four years and use either the Sharpe ratio or mean-variance objective function with trading costs  ($\text{cost}(\wret_{t-1},\wret_{t-2}) = 0.0005 \|\wret_{t-1}-\wret_{t-2}\|_{L^1}+0.0001 \|\min(\wret_{t-1},0)\|_{L^1}$). The signals are extracted from a rolling window of $L=30$ days.} 
\end{table}

\begin{figure}[t!]
  \tcapfig{Turnover of CNN+Transformer Model with and without Trading Friction Objective}
\label{fig:turnover_deep}
  \centering
    \begin{subfigure}[t]{0.475\textwidth}
  \centering
    \includegraphics[trim={0 0 0 1.2cm},clip,width=1\linewidth]{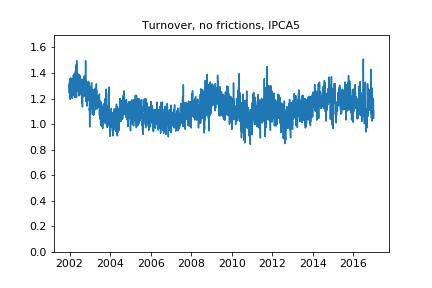}
    \caption{No Trading Friction Penalty}
    \end{subfigure}
     \begin{subfigure}[t]{0.475\textwidth}
  \centering
    \includegraphics[trim={0 0 0 1.2cm},clip,width=1\linewidth]{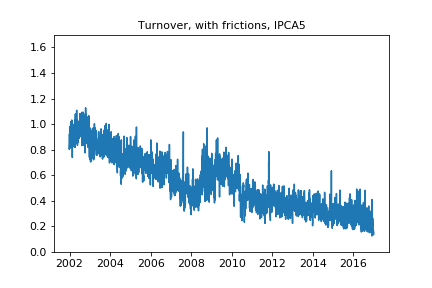}
    \caption{With Trading Friction Penalty}
    \end{subfigure}
     \bnotefig{These figures show the daily turnover of CNN+Transformer model with and without trading friction objective on the representative IPCA 5-factor residuals for the out-of-sample trading period between January 2002 and December 2016. The models are calibrated on a rolling window of four years and use the Sharpe ratio objective function with or without trading costs ($\text{cost}(\wret_{t-1},\wret_{t-2}) = 0.0005 \|\wret_{t-1}-\wret_{t-2}\|_{L^1}+0.0001 \|\min(\wret_{t-1},0)\|_{L^1}$). We define turnover as the $\ell_1$ norm of the difference between allocation weight vectors at consecutive times, i.e. $||\wret_{t-1}-\wret_{t-2}||_{L^1}$.}
\end{figure}

\begin{figure}[t!]
  \tcapfig{Proportion of Short Allocation Weights of CNN+Transformer Model with and without Trading Friction Objective}
  \label{fig:short_deep}
  \centering
  \begin{subfigure}[t]{0.475\textwidth}
  \centering
    \includegraphics[trim={0 0 0 1.2cm},clip,width=1\linewidth]{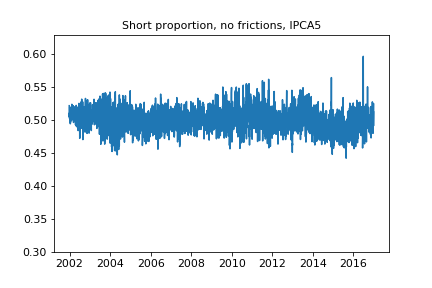}
    \caption{No Trading Friction Penalty}
    \end{subfigure}
  \begin{subfigure}[t]{0.475\textwidth}
  \centering
    \includegraphics[trim={0 0 0 1.2cm},clip,width=1\linewidth]{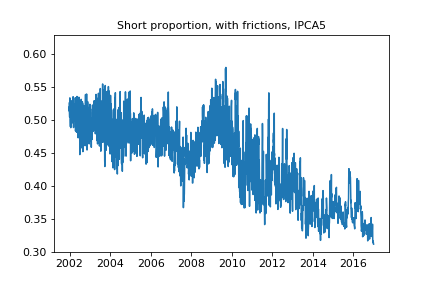}
    \caption{With Trading Friction Penalty}
  \end{subfigure}
    \bnotefig{These figures show the daily fraction of short trades of the CNN+Transformer strategies with and without trading friction objective on the representative IPCA 5-factor residuals for the out-of-sample trading period between January 2002 and December 2016. Each plot shows the absolute value of the sum of negative weights $\|\min(\wret_{t-1},0)\|_{L^1}$ relative to the sum of absolute values of all weights, which is normalized to $\|\wret_{t-1}\|_1=1$. The models are calibrated on a rolling window of four years and use the Sharpe ratio objective function with or without trading costs ($\text{cost}(\wret_{t-1},\wret_{t-2}) = 0.0005 \|\wret_{t-1}-\wret_{t-2}\|_{L^1}+0.0001 \|\min(\wret_{t-1},0)\|_{L^1}$).
    }
\end{figure}

Table \ref{tab:deep-resultsFrictions} displays the Sharpe ratios, average returns, and volatility of our CNN+Transformer model under market frictions for IPCA residuals. The results for PCA residuals are collected in the Appendix in Table \ref{tab:deep-resultsFrictionsPCA} with very similar findings. We exclude the Fama-French factor model from the analysis with market frictions, as we take the traded factors from Kenneth French Data Library as given, which are based on a larger stock universe with different trading costs and, hence, would not be directly comparable to the IPCA and PCA results.\footnote{Regardless, each factor corresponds to a portfolio of traded assets, and thus the residuals of this model could be traded in a number of a number of ways under suitable extensions. For example, we could include ETFs which try to track a value or size premium, project these latent factors onto our asset universe, or approximate each factor with a number of sparse subset of assets in our asset universe as in \cite{ruoxuan-pelger-interpretable}. However, these changes constitute differences that would make the results incomparable to the PCA and IPCA results.} As expected the Sharpe ratios are lower and range from 0.94 to 1.24 for a reasonable number of IPCA factors. The Sharpe ratio and mean-variance objective have the desired effects, but lead to overall very similar results. Importantly, the arbitrage strategies retain their economic significance even in the presence of trading costs.

These results present a lower bound on the profitability under trading frictions, as we have made four simplifying assumptions. First, in the current implementation the factor composition cannot be changed due to trading costs. A possible extension could construct the latent risk factors by including the trading friction objective. For example, the sparse representation of latent factors as in \cite{pelgerxiongsparse2020} would reduce trading costs. Second, because the policy with frictions is recursive, we are conducting an approximate training process to maintain parallelization given our computational resources and the large volume of data, but this may lead to suboptimal optimization results. However, it would be possible to conduct an exact sequential training process at the cost of more computation. Third, our modified architecture with the market-friction objective is given by the simplest modification to our architecture without frictions, but it is possible that the optimal transaction and holding cost-minimizing strategy has a more complicated functional form or is not Markovian and requires additional previous allocations. Last but not least, we keep the hyperparameters of our main analysis, but we could potentially improve the performance by employing hyperparameter tuning.

The effect of trading frictions is time-varying and our model can exploit particularly profitable arbitrage time periods by increasing trading and short positions. In Figure \ref{fig:turnover_deep} we analyze the daily turnover of a representative CNN+Transformer strategy based on IPCA 5-factor residuals and a Sharpe ratio objective. Broadly, we see that our model with trading friction penalty is able to adapt by decreasing daily turnover. However, our model seems to reduce turnover based on trading opportunities. During the times of high market volatility such as 2007--2009, arbitrage trading could be potentially be more profitable, which our model takes advantage of. On the other hand, during the later years of the calm bull market from 2011--2015, strategies with less turnover could maintain profitability. This pattern is confirmed in Figure \ref{fig:short_deep} which shows the daily proportion of allocation weights, which are short stocks in our universe. As expected the holding cost friction model reduces the overall proportion of short trades. Interestingly, our model is able to intelligently choose time periods during which it can maximize performance by taking positions with higher short proportion, such as the market turmoil at the end of 2015 and the financial crisis of 2008. Effectively, this indicates that the CNN+Transformer trading policy has learned to avoid holding and transaction costs by generally modifying the original strategy's allocations to be less short-biased on average, and to more appropriately enter short-dominant positions during relevant subperiods. 

\subsection{Portfolio Weight Concentration}

In this section we study the portfolio weight concentration of successful statistical arbitrage trading. We show that our arbitrage portfolios are well-diversified, but we can still achieve a large fraction of the profitability when restricted to a sparse set of assets. In this and the following sections, the benchmark model is the CNN+Transformer based on IPCA 5-factor residuals and a Sharpe ratio objective. The results hold qualitatively for alternative factor specifications. We study the out-of-sample portfolio weights $\wret_t$ for individual stocks. 

     \begin{figure}[t!]
       \tcapfig{Distribution of Portfolio Weights}
  \label{fig:hist}
     \begin{subfigure}[t]{.48\textwidth}
         \centering
    \includegraphics[trim={0 0 0 1.2cm},clip,width=1\linewidth]{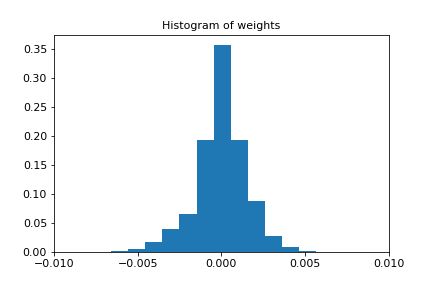}
          \caption{Distribution of stock portfolio weights}
     \end{subfigure}
     \begin{subfigure}[t]{.48\textwidth}
         \centering
       \includegraphics[trim={0 0 0 1.2cm},clip,width=1\linewidth]{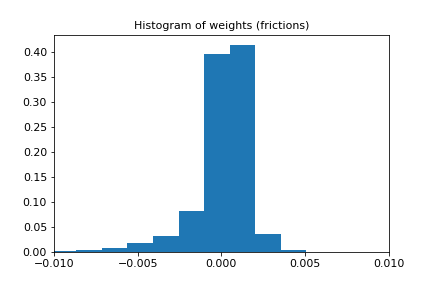}
          \caption{Distribution of stock portfolio weights with trading friction penalty}
     \end{subfigure}
         \bnotefig{These figures show histograms of the distribution of stock weights $\wret_t$ aggregated over time. Subplot (a) shows the out-of-sample weights for our empirical benchmark model, which is the CNN+Transformer model based on IPCA 5-factor residuals. Subplot (b) estimates the baseline model with trading friction penalty.  The out-of-sample trading period is between January 2002 and December 2016.
 }
     \end{figure}

Figure \ref{fig:hist} shows that the optimal portfolio weights are well-diversified and do not rely on excessively large weights on individual stocks. The left subplot the shows the histogram of the distribution of stock weights $\wret_t$ aggregated over time. The weights are approximately normally distributed and centered at zero. Hence, most weights concentrate around values in $[-0.5\%, +0.5\%]$, which implies a relatively well-diversified portfolio. Second, the portfolios correspond to long-short positions with approximately similar weights on each of the two legs. The right subplot estimates the baseline model with the trading friction penalty from Section \ref{sec:market-frictions-results}. While a trading friction penalty increases the kurtosis, which implies sparser weights, we still do not observe excessive weights on individual stocks. The trading friction penalty also reduces short sales, which is consistent with Figure \ref{fig:short_deep}.

Our statistical arbitrage trading policy does not target specific industries. Figure \ref{fig:industry} in the Appendix shows the rolling industry concentration of portfolio weights standardized by the population industry concentration. The fraction invested in a specific industry follows very closely the sample proportion of stocks in the corresponding industry. While there is some minor variation in the industry concentration over time, it deviates less than 20\% from the population concentration. This suggests that stocks in all industries offer statistical arbitrage opportunities.

     \begin{figure}[t!]
       \tcapfig{Performance of Sparse Portfolios}
       \vspace{-0.4cm}
  \label{fig:sparse}
  \begin{center}
  PANEL A: Largest Stock Portfolio Weights \\
  \end{center}
     \begin{subfigure}[t]{.32\textwidth}
         \centering
    \includegraphics[trim={0 0 0 1.2cm},clip,width=1\linewidth]{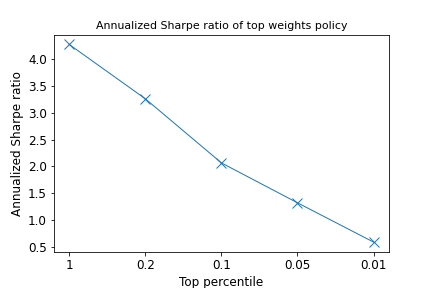}
          \caption{Sharpe ratio}
        \end{subfigure}
               \begin{subfigure}[t]{.32\textwidth}
         \centering
    \includegraphics[trim={0 0 0 1.2cm},clip,width=1\linewidth]{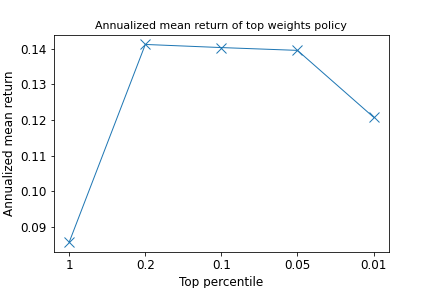}
          \caption{Mean Return}
     \end{subfigure}
          \begin{subfigure}[t]{.32\textwidth}
         \centering
    \includegraphics[trim={0 0 0 1.2cm},clip,width=1\linewidth]{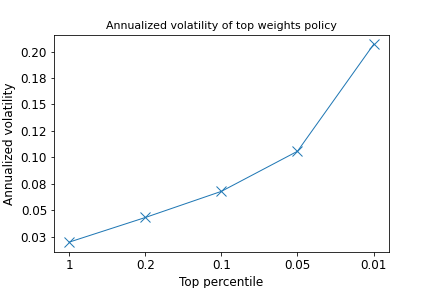}
          \caption{Volatility}
     \end{subfigure}
     \begin{center}
   PANEL B: Largest Residual Portfolio Weights \\
   \end{center}
     \begin{subfigure}[t]{.32\textwidth}
         \centering
    \includegraphics[trim={0 0 0 1.2cm},clip,width=1\linewidth]{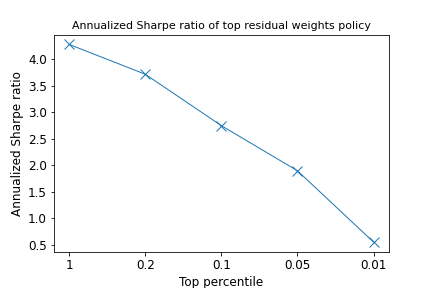}
          \caption{Sharpe ratio}
      \end{subfigure}
               \begin{subfigure}[t]{.32\textwidth}
         \centering
    \includegraphics[trim={0 0 0 1.2cm},clip,width=1\linewidth]{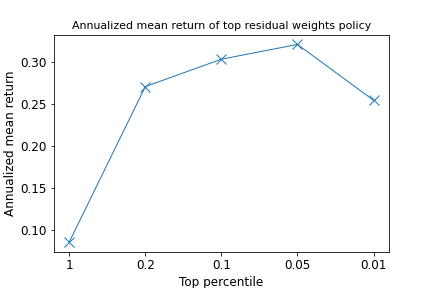}
          \caption{Mean Return}
     \end{subfigure}
          \begin{subfigure}[t]{.32\textwidth}
         \centering
    \includegraphics[trim={0 0 0 1.2cm},clip,width=1\linewidth]{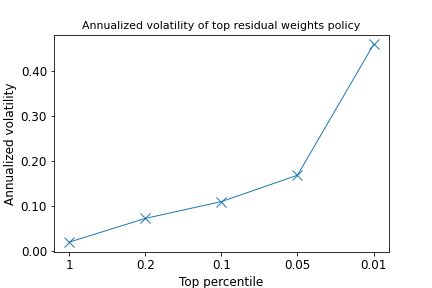}
          \caption{Volatility}
     \end{subfigure}         
         \bnotefig{These figures show the annualized out-of-sample Sharpe ratio, mean return and volatility of arbitrage strategies based on selecting only the largest portfolio weights in absolute value. Panel A selects the proportion $p$ of the most extreme stock weights $\wret_t$ for trading, while panel B selects the proportion $p$ of the most extreme residual weights $\w_t$ for the trading policy. The baseline model is the CNN+Transformer model based on IPCA 5-factor residuals for the out-of-sample trading period between January 2002 and December 2016. We consider the full model $p=1$ and the fraction $p=0.01,0.05,0.1$ and $0.2$.
          }
     \end{figure}

     \begin{figure}[t!]
       \tcapfig{Cumulative Returns of Sparse Portfolios}
  \label{fig:cumsparse}
     \begin{subfigure}[t]{.48\textwidth}
         \centering
    \includegraphics[trim={0 0 0 1.2cm},clip,width=1\linewidth]{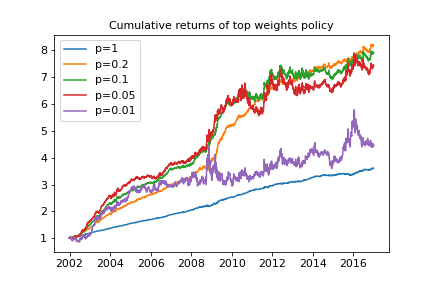}
          \caption{Cumulative Returns of Extreme Stock Portfolio Weights}
     \end{subfigure}
     \begin{subfigure}[t]{.48\textwidth}
         \centering
       \includegraphics[trim={0 0 0 1.2cm},clip,width=1\linewidth]{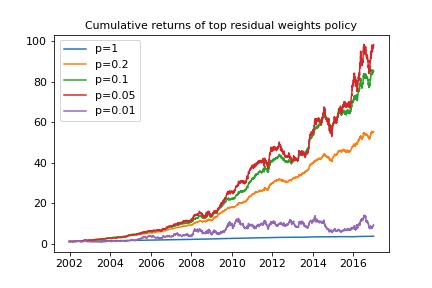}
          \caption{Cumulative Returns of Extreme Residual Portfolio Weights}
     \end{subfigure}
         \bnotefig{These figures show the cumulative returns of sparse portfolio weights in the stock and residual weight vectors. Subplot (a) selects the proportion $p$ of the stock weights $\wret_t$ with the largest absolute value for trading, while subplot (b) selects the proportion $p$ of the residual weights $\w_t$ with the largest absolute value for the trading policy. The baseline model is the CNN+Transformer model based on IPCA 5-factor residuals for the out-of-sample trading period between January 2002 and December 2016. We consider the full model $p=1$ and the fraction $p=0.01,0.05,0.1$ and $0.2$.
          }
     \end{figure}

While our statistical arbitrage portfolios are well-diversified, a subset of a few stocks can already achieve most of their performance. This finding is important for optimization under trading frictions. We construct sparse trading policies by selecting only the proportion $p$ of the stock weights $\wret_t$ with the largest absolute value for trading. We compare the full model with sparse models, which invest only in $p=1\%$, $5\%$, $10\%$ or $20\%$ of the stocks. Note that this represents only a lower bound for the performance of sparse portfolios, as we are not including the sparsity constraint in the optimization.

Figure \ref{fig:sparse} in panel A shows the annualized out-of-sample Sharpe ratio, mean return and volatility of arbitrage strategies based only the most extreme portfolio weights. 
Portfolios with less stocks are less diversified and as expected have a higher volatility. However, focussing on the extreme weights can increase the mean return of the portfolios. The Sharpe ratio of sparse portfolios is generally decreasing in the degree of sparsity, as the loss in diversification dominates the effect of higher expected returns. However, a sparse portfolio can capture a substantial amount of the profitability. We achieve out-of-sample Sharpe ratios above two with only 10\% of the stocks. These are lower bounds as we use the model that is optimized for non-sparse weights, and select only the extreme weights. It is noteworthy, that using only 5\% of the stocks, which corresponds to roughly 25 stocks in our sample, achieves mean returns of 14\%.  However, the performance deteriorates when our portfolio consists of only 1\%, corresponding to around 5 stocks. The cumulative return time series in Figure \ref{fig:cumsparse} further illustrate the different effect of sparsity on the mean and variance of the strategies.  

In our second analysis, we construct sparse residual portfolio weights $\w_t$. Similar to the sparse stock weights, we select the proportion $p$ of the most extreme residual weights $\w_t$ for the trading policy. A sparse stock portfolio weight $\wret_t$ combines a sparse factor representation and a sparse trading policy in residuals. By studying the largest portfolio weights in residuals, we can separate the effect of sparse factors from a sparse trading policy in residuals. This is relevant as some factors could be approximated by a sparse set of traded assets, for example ETFs. Panel B in Figure \ref{fig:sparse} shows the out-of-sample Sharpe ratio, mean return and volatility for sparse residual portfolio weights $\w_t$. The Sharpe ratio effects are very similar, as targeting the most extreme residual portfolios results in larger mean returns, while it increases the variance. However, a sparse residual portfolio can achieve around twice the expected return of a sparse stock portfolio.

\subsection{Complexity of Arbitrage Trading}

Our CNN+Transformer solution can discover complex trading signals and policies. As a robustness check, we study how much simple reversal strategies can earn. Given the residuals from our baseline IPCA-5-factor model, we construct long-short trading strategies for different time lags. Such high-minus-low strategies have been motivated in \cite{he_et_al2022}. In more detail, the portfolio weights of the simple trading strategy are long-short portfolios, that buy the 20\% lowest residuals and sell the 20\% highest residuals from $L$ periods in the past.  

     \begin{figure}[b!]
       \tcapfig{Simple Reversal Trading}
  \label{fig:simplereversal}
     \begin{subfigure}[t]{.32\textwidth}
         \centering
    \includegraphics[trim={0 0 0 1.2cm},clip,width=1\linewidth]{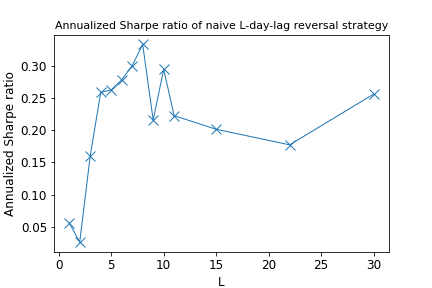}
          \caption{Sharpe ratio}
        \end{subfigure}
               \begin{subfigure}[t]{.32\textwidth}
         \centering
    \includegraphics[trim={0 0 0 1.2cm},clip,width=1\linewidth]{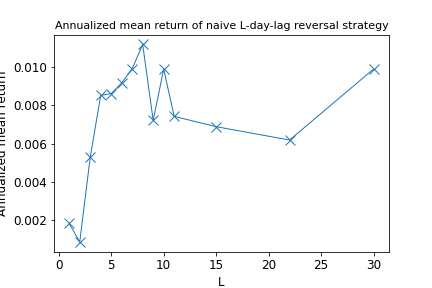}
          \caption{Mean Return}
     \end{subfigure}
          \begin{subfigure}[t]{.32\textwidth}
         \centering
    \includegraphics[trim={0 0 0 1.2cm},clip,width=1\linewidth]{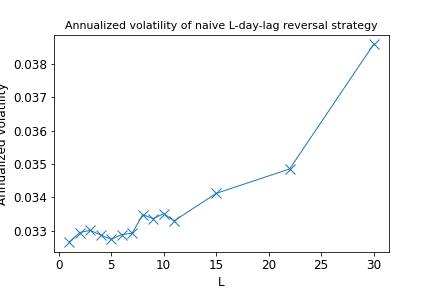}
          \caption{Volatility}
     \end{subfigure}
    \bnotefig{These figures show the annualized out-of-sample Sharpe ratio, mean return and volatility of simple reversal strategies based on IPCA 5-factor residuals. The portfolio weights of the simple trading strategy are long-short portfolios, that buy the 20\% lowest residuals and sell the 20\% highest residuals for $L$ periods in the past. The out-of-sample trading period is between January 2002 and December 2016.
          }
     \end{figure}

Figure \ref{fig:simplereversal} shows the out-of-sample Sharpe ratio, mean return and volatility of simple reversal strategies. First, these simple reversal strategies yield positive returns and Sharpe ratios. The variance is increasing for longer lags. The reversal returns increase substantially with a lag of at least one week (five trading days). The Sharpe ratios are primarily driven by the low variance and achieve Sharpe ratios of up to 0.3 for lags between one and two weeks. 

Simple reversal strategies result in substantially lower mean returns and Sharpe ratios than our complex arbitrage strategies. This confirms that simple ad-hoc approaches are not sufficient to leverage mispricing and successful arbitrage trading is more complex than simple reversal patterns. The findings also suggest that there is profitability in longer holding periods for signals that are further in the past. We will study this aspect in depth in the next section.

\subsection{Persistence of Arbitrage}

We study arbitrage trading for different holding periods and show that statistical arbitrage signals are persistent. Given the out-of-sample portfolio weights $\wret_t$, we document the portfolio performance for longer holding periods. As in the previous sections, our benchmark model is the CNN+Transformer based on IPCA 5-factor residuals and a Sharpe ratio objective.

First, we consider longer holding periods, that can overlap. An investment with weights $\wret_t$ is held for $B$ trading days, ranging from 1 to 30 trading days. The trading with longer holding periods is overlapping and the fraction $1/B$ is invested based on new weights every trading day. Assuming log returns, this is equivalent to the portfolio weights
 \begin{align*}
     w^{R,\text{$B$-days}}_t = \frac{1}{B} \sum_{l=0}^{B-1} w^R_{t-l}.
 \end{align*}

\begin{figure}[t!]
       \tcapfig{Performance for Longer Holding Periods}
         \label{fig:holding}
         \begin{center}
  PANEL A: Overlapping Trading \\
  \end{center}
     \begin{subfigure}[t]{.32\textwidth}
         \centering
    \includegraphics[trim={0 0 0 0.7cm},clip,width=1\linewidth]{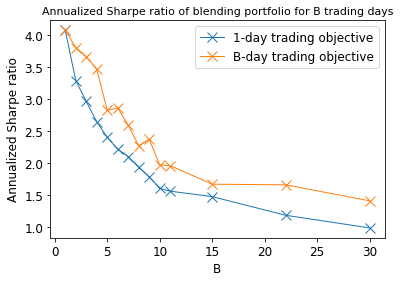}
          \caption{Sharpe ratio}
        \end{subfigure}
               \begin{subfigure}[t]{.32\textwidth}
         \centering
    \includegraphics[trim={0 0 0 0.7cm},clip,width=1\linewidth]{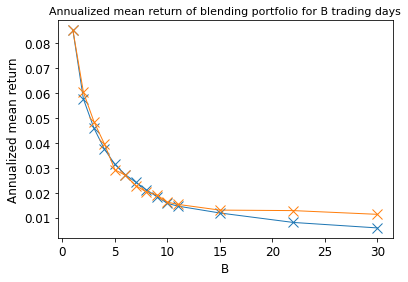}
          \caption{Mean Return}
     \end{subfigure}
          \begin{subfigure}[t]{.32\textwidth}
         \centering
    \includegraphics[trim={0 0 0 0.7cm},clip,width=1\linewidth]{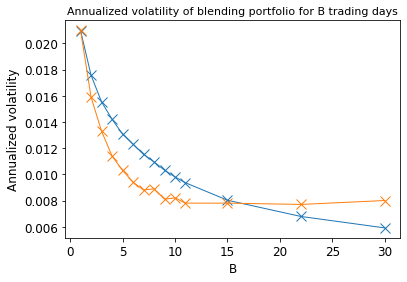}
          \caption{Volatility}
     \end{subfigure}
     \begin{center}
       PANEL B: Non-overlapping Trading \\
  \end{center}
     \begin{subfigure}[t]{.32\textwidth}
         \centering
    \includegraphics[trim={0 0 0 0.7cm},clip,width=1\linewidth]{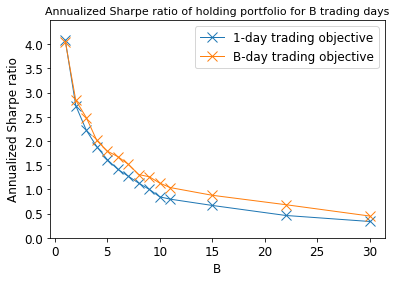}
          \caption{Sharpe ratio}
        \end{subfigure}
               \begin{subfigure}[t]{.32\textwidth}
         \centering
    \includegraphics[trim={0 0 0 0.7cm},clip,width=1\linewidth]{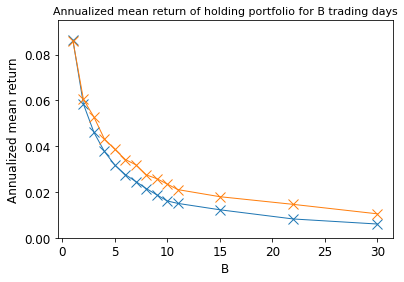}
          \caption{Mean Return}
     \end{subfigure}
          \begin{subfigure}[t]{.32\textwidth}
         \centering
    \includegraphics[trim={0 0 0 0.7cm},clip,width=1\linewidth]{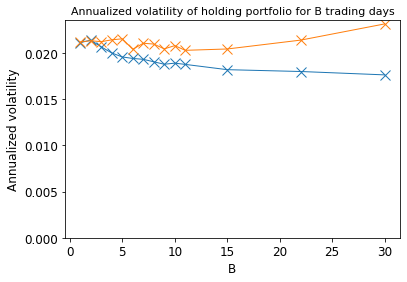}
          \caption{Volatility}
     \end{subfigure}
    \bnotefig{These figures show the annualized out-of-sample Sharpe ratio, mean return and volatility for different holding periods for our empirical benchmark model, which is the CNN+Transformer model based on IPCA 5-factor residuals. An investment with weights $\wret_t$ is held for $B$ trading days, ranging from 1 to 30 trading days. The blue line indicates a model that is estimated with a daily trading objective (that is, our baseline model), while the orange line displays a model that is estimated with a B-holding day objective. In panel A the trading with longer holding periods is overlapping and the fraction $1/B$ is invested based on new weights every trading day. In panel B, we estimate for a holding period of $B$ days, the performance of the $B$ different trading strategies starting at different days and report the average results. Hence, panel B reports results for non-overlapping portfolios. The out-of-sample trading period is between January 2002 and December 2016.
          }
     \end{figure}

Panel A in Figure \ref{fig:holding} shows the annualized out-of-sample Sharpe ratio, mean return and volatility for different overlapping holding periods. Daily trading indeed generates the highest payoff in terms of mean returns and Sharpe ratios. The half-life in terms of Sharpe ratios is around seven trading days. In fact, the Sharpe ratio is still above one with a one month (22 business days) holding period, when using signals estimated with a 1-day trading objective. As expected, the lower trading frequency can reduce the volatility. This shows that statistical arbitrage is long-lived and persists for several days and even weeks. We also include the results, when we change the estimation to optimize the Sharpe ratio of the overlapping multi-horizon holding periods. It is possible to achieve out-of-sample Sharpe ratios around 1.5 with a one month holding period, when directly optimizing for longer holding periods. The mean returns for a strategy based on the B-day trading objective decline less than for a 1-day trading objective. For longer horizons this comes at the cost of a slightly higher variance. For shorter horizons, the $B$-day trading objective reduces the variance by taking advantage of additional diversification.
     
Overlapping trading generates diversification effects between strategies. It is a valid measure of how fast signals are dying out, but it is still implicitly based on daily trading.
Optimizing for a longer holding period can take advantage of this diversification effects and increase the mean return without strongly increasing the variance. This is one reason why we can maintain Sharpe ratios of 1.5 for one month holding periods.

An alternative analysis actually trades only every $B$ days without overlap. As there are $B$ different starting days, this yields $B$ possible implementations of a $B$-day holding strategy. We report the average performance for each of the possible $B$ starting days, that is, we average the performance metrics, but do not create a portfolio of overlapping returns. Panel B in Figure \ref{fig:holding} shows the corresponding results. By construction, the mean returns for weights based on the 1-day trading objective are identical to overlapping trading. The difference between the two analyses comes from the variance. The overlapping trading generates large diversification benefits, which are not achieved with the separate non-overlapping trading. As a result the Sharpe ratios for non-overlapping trading are lower. However, even with a one month (22 business days) holding period, we still achieve an out-of-sample Sharpe ratio of around 0.5. There is little improvement when optimizing the trading for a longer holding period with non-overlapping trading.

Overall, we have two main findings. First, arbitrageurs do indeed correct prices quickly. Depending on the analysis around half of the Sharpe ratio vanishes after one or two weeks. Arbitrageurs do engage in arbitrage trading as signals do eventually die out. Second, a large fraction of the statistical arbitrage signal is persistent. Even with holding periods of over one month, the signals can be highly profitable. This has important economic implications. The persistence in arbitrage signals could be explained by the limited capacity of arbitrageurs, that is, arbitrageurs might have insufficient capital to fully exploit temporal mispricing quickly. An alternative explanation could relate to strategic trading. Arbitrageurs might try to take advantage of temporal mispricing, while at the same time limit their market impact to avoid revealing their detected signal. A third explanation could relate to behavioral trading. In the presence of noise traders or uninformed traders, the adjustment of prices also depends on how quickly the non-arbitrageurs adjust their expectations. The persistence of arbitrage has also practical implications. It explains our findings from Section \ref{sec:market-frictions-results}, why statistical arbitrage can also be exploited in the presence of transaction costs by avoiding frequent trading.

\subsection{Estimated Structure}\label{sec:interpretation}

What are the patterns that our CNN+Transformer model can learn and exploit? In order to answer this question, we analyze the different building blocks of our benchmark model and show their structure on representative and informative residuals inputs. Our goal is to ascertain, characterize, and explain the role that the convolutional features and attention heads play in the determination of the final allocation weight and recognition of salient time series patterns. The benchmark model for this section is the CNN+Transformer based on IPCA 5-factor residuals and a Sharpe ratio objective. The model is calibrated on the first 8 years of data and kept constant, which allows us to study the signal and allocation function.

\begin{figure}[t!]
    \tcaptab{Examples of Allocation and Returns of CNN+Transformer Strategy}
\label{fig:example-trading-residuals}
    \begin{subfigure}[t]{0.5\textwidth}
        \centering
        \includegraphics[trim={0 0 0 1.2cm},clip,width=1\linewidth]{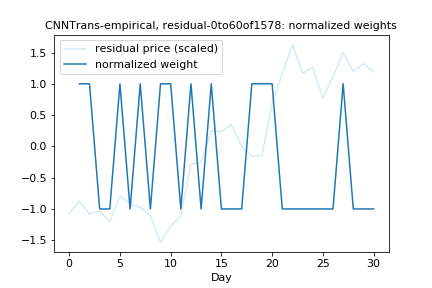}
    \end{subfigure}
    \begin{subfigure}[t]{0.5\textwidth}
        \centering
        \includegraphics[trim={0 0 0 1.1cm},clip,width=1\linewidth]{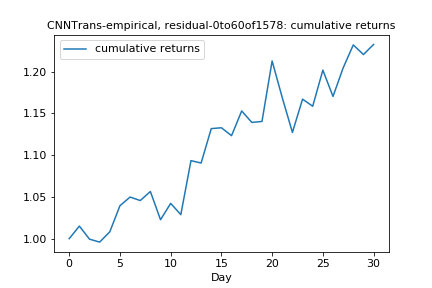}
    \end{subfigure}
    \begin{subfigure}[t]{0.5\textwidth}
        \centering
        \includegraphics[trim={0 0 0 1.2cm},clip,width=1\linewidth]{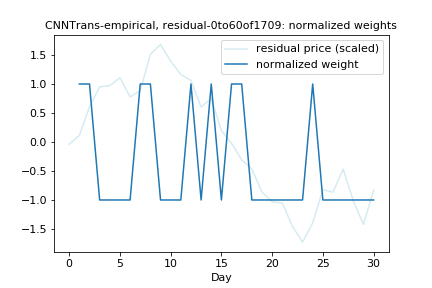}
    \end{subfigure}
    \begin{subfigure}[t]{0.5\textwidth}
        \centering
        \includegraphics[trim={0 0 0 1.2cm},clip,width=1\linewidth]{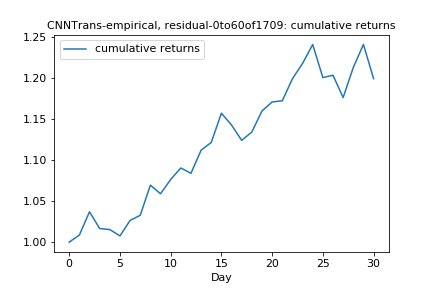}
    \end{subfigure}
    \bnotetab{These plots display representative examples of the CNN+Transformer out-of-sample arbitrage trading on a sample of residuals from the IPCA 5-factor model. The left subplots show the normalized cumulative returns of the residuals and the normalized allocation weight, which the specific residual has in the trading strategy. The right subplots illustrates the payoff of trading the specific residual with the displayed allocation function. The model is calibrated on the first 8 years of data and kept constant.} 
\end{figure}

As an illustrative example, Figure \ref{fig:example-trading-residuals} shows the allocation and return on representative residuals. The left subplot displays an out-of-sample time-series of cumulative returns of a randomly selected residual and its value in the allocation function. We normalize the allocation weight to have an absolute value of one, that is, for this illustrative example we only trade this particular residual. The right subplots depicts the payoff of trading the specific residual with the displayed allocation function. The first residual shows mean-reversion patterns, which are successfully detected and exploited by the function $ {\bm w}^{\epsilon |\text{CNN+Trans}}$. The second residual has a downward trend, which is also correctly detected and taken advantage of by the model.\footnote{Note that our in our empirical study the model trades in all residuals and is not limited to trade only in one residual. Hence, the empirical performance is substantially better as shown in Figure \ref{fig:deep-cum-rets}.} These examples suggest that the CNN+Transformer model can learn mean-reversion and trend patterns. Figures \ref{fig:weights-and-signals} and \ref{fig:example-trading-residuals-appendix} are further examples with the same takeaways.


\begin{figure}[t!]
  \tcapfig{Local Basic Patterns of Benchmark Model}\label{fig:interpretation-basic-patterns}
  \begin{subfigure}[t]{.245\textwidth}
  \centering
    \includegraphics[width=1\linewidth]{PlotsInterpretation/basis-pattern-1.png}
    \caption{Basic pattern 1}
   \end{subfigure}
  \begin{subfigure}[t]{.245\textwidth}
  \centering
    \includegraphics[width=1\linewidth]{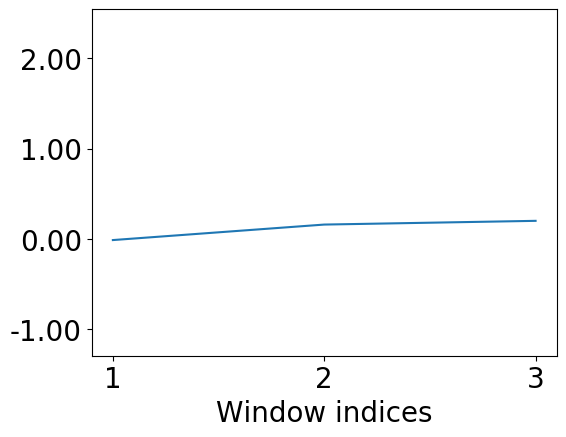}
    \caption{Basic pattern 2}
   \end{subfigure}
  \begin{subfigure}[t]{.245\textwidth}
  \centering
    \includegraphics[width=1\linewidth]{PlotsInterpretation/basis-pattern-3.png}
    \caption{Basic pattern 3}
   \end{subfigure}
  \begin{subfigure}[t]{.245\textwidth}
  \centering
    \includegraphics[width=1\linewidth]{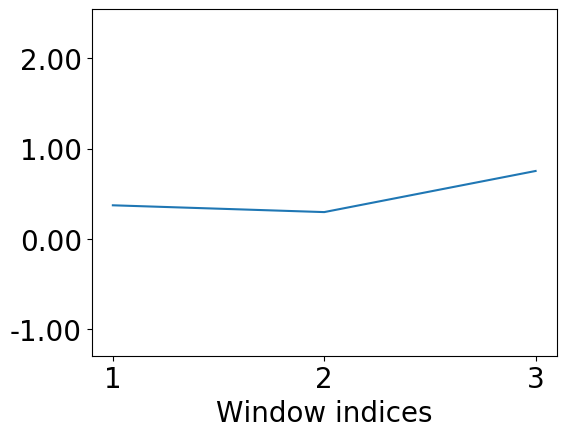}
    \caption{Basic pattern 4}
   \end{subfigure}
  \begin{subfigure}[t]{.245\textwidth}
  \centering
    \includegraphics[width=1\linewidth]{PlotsInterpretation/basis-pattern-5.png}
    \caption{Basic pattern 5}
   \end{subfigure}
  \begin{subfigure}[t]{.245\textwidth}
  \centering
    \includegraphics[width=1\linewidth]{PlotsInterpretation/basis-pattern-6.png}
    \caption{Basic pattern 6}
   \end{subfigure}
  \begin{subfigure}[t]{.245\textwidth}
  \centering
    \includegraphics[width=1\linewidth]{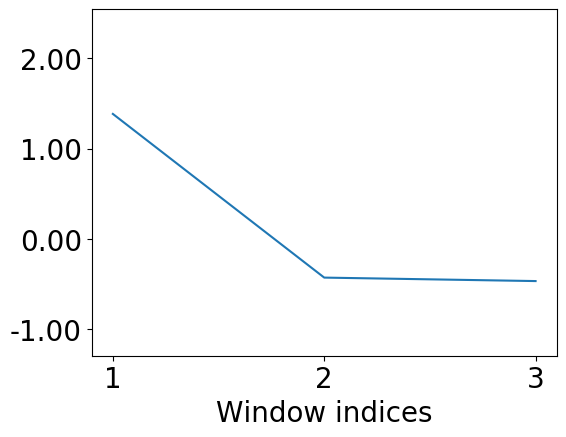}
    \caption{Basic pattern 7}
   \end{subfigure}
  \begin{subfigure}[t]{.245\textwidth}
  \centering
    \includegraphics[width=1\linewidth]{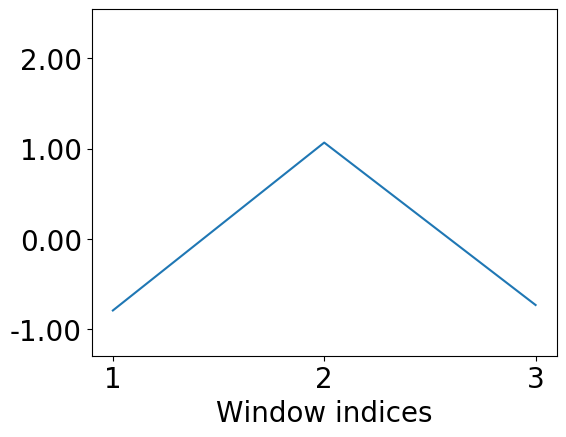}
    \caption{Basic pattern 8}
   \end{subfigure}
\bnotefig{These figures show the $D=8$ local filters of the CNN estimated for the benchmark model in our empirical analysis. These are projections of our higher dimensional nonlinear filter from a 2-layer CNN into two-dimensional linear filters. These building blocks are labeled ``basic patterns''. The benchmark model is the CNN+Transformer model based on IPCA 5-factor residuals. We estimate the model on only once on the first $T_{\text{train}}$=8 years based on the Sharpe ratio objective.}
\end{figure}

\begin{figure}[t!]
    \tcaptab{Example Attention Weights for Sinusoidal Residual Inputs}\label{fig:simulated-attention-weights}
\begin{subfigure}[t]{\textwidth}
        \centering
        \includegraphics[trim={0 0 0 1.2cm},clip,width=1\linewidth]{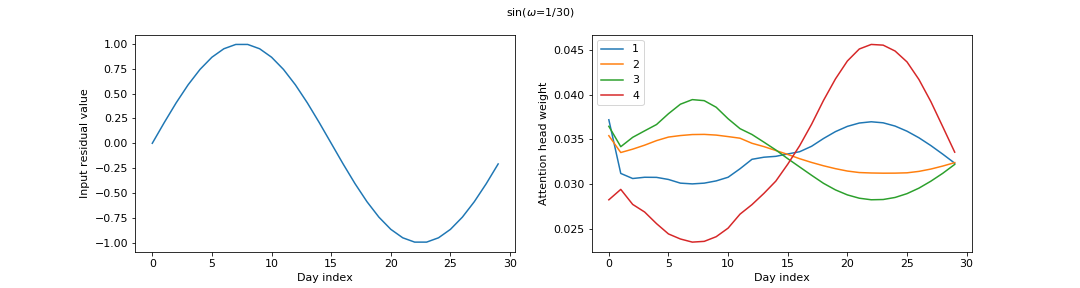}
    \end{subfigure}
    \subcaption{Input residual and attention head weights for $x_l=\sin \left(2 \pi \frac{l}{30}\right)$}
    \begin{subfigure}[t]{\textwidth}
        \centering
        \includegraphics[trim={0 0 0 1.2cm},clip,width=1\linewidth]{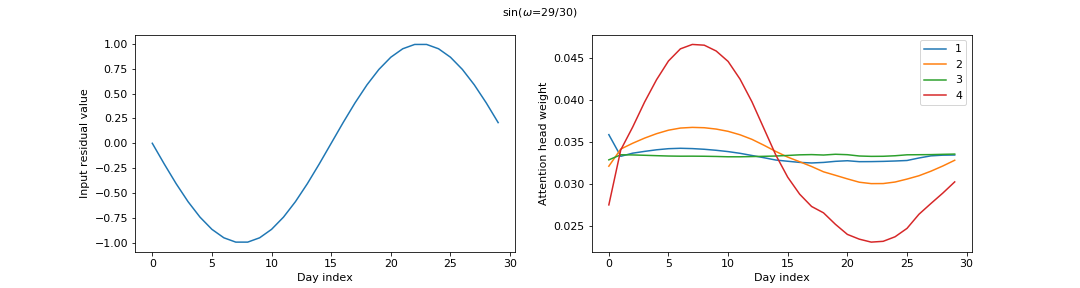}
    \end{subfigure}  
    \subcaption{Input residual and attention head weights for for $x_l=\sin \left(2 \pi \frac{l+15}{30} \right)$}
      \bnotefig{\scriptsize These plots show the attention head weights of the CNN+Transformer benchmark model for simulated sinusoidal residual input time series. Both sine functions have a cycle of 30 days and the second is shifted by 15 days. The right subplot shows the attention weights for the $H=4$ attention heads for the specific residuals. The empirical benchmark model is the CNN+Transformer model based on IPCA 5-factor residuals. We estimate the model only once on the first $T_{\text{train}}$=8 years based on the Sharpe ratio objective.}  
\end{figure}

\begin{figure}[t!]
  \tcapfig{CNN+Transformer Model Structure for Representative Residual}\label{fig:windowIPCA5}
  \begin{subfigure}[t]{.33\textwidth}
  \centering
    \includegraphics[width=1\linewidth]{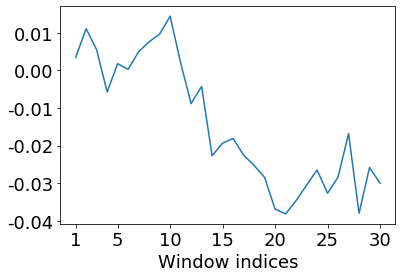}
    \caption{Cumulative residual}
    \end{subfigure}
    \begin{subfigure}[t]{.33\textwidth}
  \centering
    \includegraphics[width=1\linewidth]{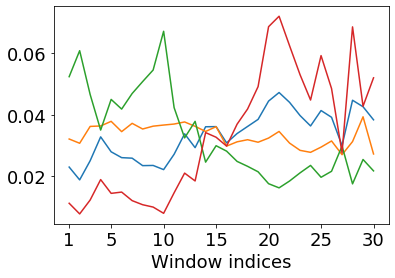}
    \caption{Attention weights per head}
  \end{subfigure}
  \begin{subfigure}[t]{.33\textwidth}
  \centering
    \includegraphics[width=1\linewidth]{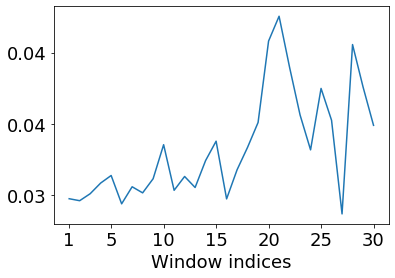}
    \caption{Average attention weights}
  \end{subfigure}
 
  \begin{subfigure}[t]{.245\textwidth}
  \centering
    \includegraphics[width=1\linewidth]{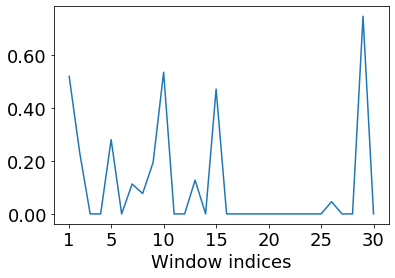}
    \caption{CNN activations for\\ basic pattern 1}
    \end{subfigure}
    \begin{subfigure}[t]{.245\textwidth}
  \centering
    \includegraphics[width=1\linewidth]{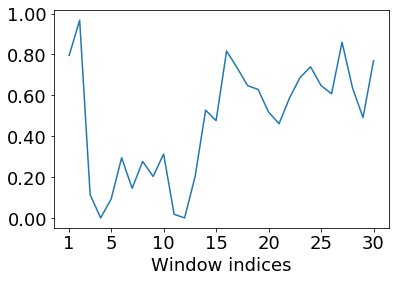}
    \caption{CNN activations for\\ basic pattern 2}
    \end{subfigure}
    \begin{subfigure}[t]{.245\textwidth}
  \centering
    \includegraphics[width=1\linewidth]{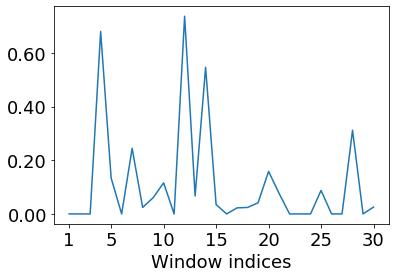}
    \caption{CNN activations for\\ basic pattern 3}
    \end{subfigure}
    \begin{subfigure}[t]{.245\textwidth}
  \centering
    \includegraphics[width=1\linewidth]{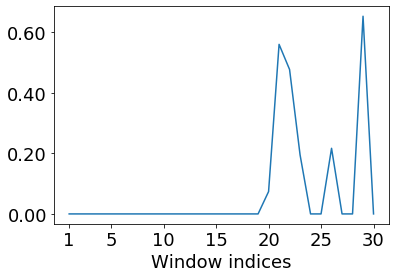}
    \caption{CNN activations for\\ basic pattern 4}
    \end{subfigure}
    
     \begin{subfigure}[t]{.245\textwidth}
  \centering
    \includegraphics[width=1\linewidth]{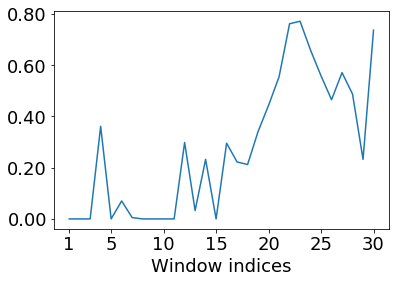}
    \caption{CNN activations for\\ basic pattern 5}
    \end{subfigure}
    \begin{subfigure}[t]{.245\textwidth}
  \centering
    \includegraphics[width=1\linewidth]{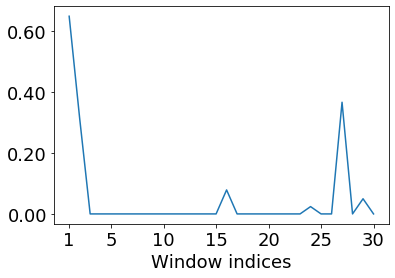}
    \caption{CNN activations for\\ basic pattern 6}
    \end{subfigure}
    \begin{subfigure}[t]{.245\textwidth}
  \centering
    \includegraphics[width=1\linewidth]{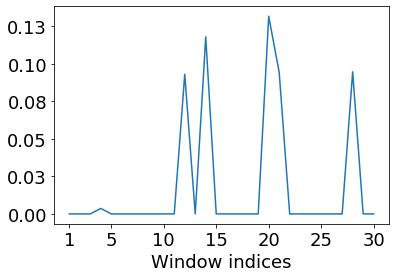}
    \caption{CNN activations for\\ basic pattern 7}
    \end{subfigure}
    \begin{subfigure}[t]{.245\textwidth}
  \centering
    \includegraphics[width=1\linewidth]{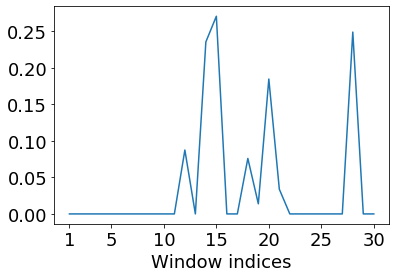}
    \caption{CNN activations for\\ basic pattern 8}
    \end{subfigure}
     \bnotefig{These figures illustrate the different components of the CNN+Transformer benchmark model evaluated for a randomly selected, representative empirical residual. The cumulative residual returns, which are the input to the model, are plotted in (a). The convolutional activations (d)--(k) quantify the exposure of the residual time-series to local basis filters. Subplot (b) displays the attention weights for the $H=4$ attention heads, which represent global dependency patterns. Subplot (c) shows the average of these four attention head weights. The empirical benchmark model is the CNN+Transformer model based on IPCA 5-factor residuals. We estimate the model on only once on the first $T_{\text{train}}$=8 years based on the Sharpe ratio objective.}
\end{figure}

Next, we ``dissect'' the CNN+Transformer model to understand what type of functions it can estimate. Our analysis begins with the eight basic convolutional patterns learned by the convolutional layers of our network, which are displayed in Figure \ref{fig:interpretation-basic-patterns}. The CNN represents a given time-series as a matrix of exposures to local basic patterns. As explained in Section \ref{sec:theoryCNN}, these local filters are more complicated than simple local linear filters, but we can project our CNN filters into two-dimensional orthogonal linear filters, which are more interpretable. These local patterns are the building blocks to construct global patterns. We see that these basic patterns display a variety of salient price behavior which are considered to be important. Basic patterns 4 and 6 capture local upward trends, basic patterns 3 and 7 track local downward trends and basic patterns 1, 5 and 8 learn reversion patterns. However, the basis patterns do not include very spiked, sharp changes. Overall, the building blocks seem to be sufficient to construct any smooth trend and mean-reversion pattern.

We can understand the global patterns learned by the transformer by studying the attention function. The attention functions ${\bm \alpha_i}(.,.)$ of each attention head $i=1,...,H=4$ capture the dependencies between the local patterns. Our arbitrage signal can be interpreted as ``loadings'' to these ``attention factors''. We use the same $H$ attention functions for all residuals, but in order to visualize them, we evaluate them for a given residual time-series, which yield the attention weights per head:
\begin{align*}
\alpha_{i,j} = {\bm \alpha_i} \left( \tilde x_L, \tilde x_j \right) \qquad \text{for $i=1,...,H$}.
\end{align*}
As our signal only depends on the final cumulative return projection $h^{\text{proj}}_L$, the attention weights $\alpha_{i,j}$ for $i=1,..,H$ and $j=1,...,L$ contain all the ``global factor'' information. Hence, we will plot the $H \times L$ dimensional attention weights of the attention heads to understand which global patterns are activated by specific time-series.

Figure \ref{fig:simulated-attention-weights} plots the attention head weights for simulated sinusoidal residual input time-series. Note that the attention head weights discover the sinusoidal pattern although the model was estimated on the empirical data and not specifically trained for this simulated input. The different attention heads capture different patterns. The fourth attention head displayed in red has the strongest activation and capture high-frequency mean reversion patterns. These attention head weights are positive for negative realizations. We will label these fourth attention head weights a ``negative reversal'' pattern. The third attention head weights depicted by the green curve co-move with the mean-reversion patterns of the original time-series, that is they are positive for high values, but seem to be only activated if this positive ``hilltop'' appears at the beginning of the time-series. If the mean-reversion cycle achieves its positive values at the end, the third attention head is not activated. We will label these third attention head weights the ``early reversal'' pattern. The first attention head in blue seems to be a ``dampened'' version of the fourth red attention head. Figure \ref{fig:simulated-attention-weightsAppendix} in the Appendix shows additional simulated input time-series that confirm this interpretation. In summary, the different attention head weights can be assigned to specific global patterns.

In Figure \ref{fig:windowIPCA5}, we plot the different components of the CNN+Transformer model evaluated on a randomly selected, representative 30-day empirical residual. The cumulative residual in subfigure (a) is the input to the CNN. This $L=30$ dimensional vector is represented by the CNN in terms of its ``exposure'' to local basic pattern. The subfigures (d)--(k) show this $D \times L$ dimensional representation, which is the output of the CNN. As we have $D=8$ local filters, we obtain eight time-series that display the activation to each filter. For example, basic pattern 1 is associated with a ``reversal kink'' in subfigure \ref{fig:interpretation-basic-patterns}(a) and hence has the strongest activation to this basic pattern on day 28, when the residual has a downward spike. This $D \times L$ matrix of exposures to local patterns is the input to the transformer. The attention head weights in subfigure (b) connect the local patterns to a global pattern. The fourth attention head weight in red has its highest values during the ``bottom'' of the residual movements, confirming our previous intuition that this attention head activates during bad times. The third attention head weight in green spikes during the ``top'' at the beginning of the residual time-series, which is in line with our interpretation as an early reversal pattern. The first attention head in blue is a dampened version of the red attention head. The average over the four attention head weights depicted in subfigure (c) suggests that the heads attend on average more closely to the latter half of the time series.

\begin{figure}[h!]
  \tcapfig{CNN+Transformer Model Structure for Representative Residual Over Time}\label{fig:fullInterpretation}
  \begin{subfigure}[t]{.33\textwidth}
  \centering
    \includegraphics[width=1\linewidth]{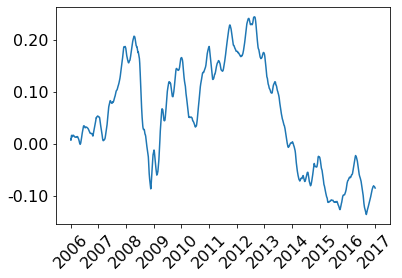}
    \caption{Cumulative residuals}
    \end{subfigure}
    \begin{subfigure}[t]{.33\textwidth}
  \centering
    \includegraphics[width=1\linewidth]{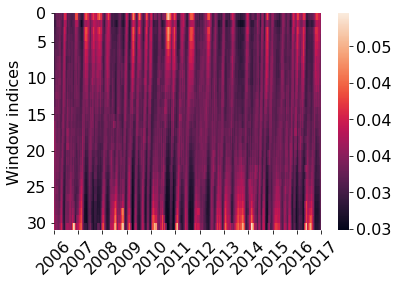}
    \caption{Average attention weights}
  \end{subfigure}
  \begin{subfigure}[t]{.33\textwidth}
  \centering
    \includegraphics[width=1\linewidth]{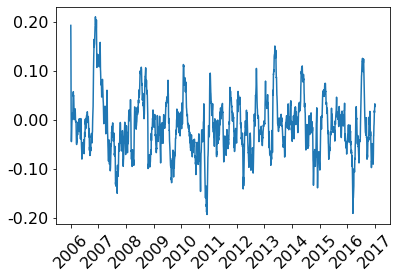}
    \caption{Allocation weights}
  \end{subfigure}
    \begin{subfigure}[t]{.245\textwidth}
  \centering
    \includegraphics[width=1\linewidth]{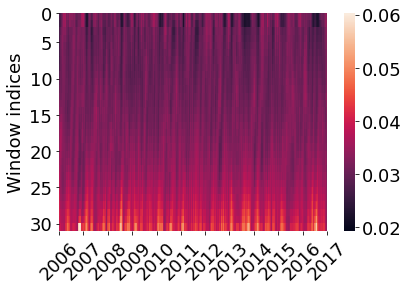}
    \caption{Attention weights for head 1}
    \end{subfigure}
    \begin{subfigure}[t]{.245\textwidth}
  \centering
    \includegraphics[width=1\linewidth]{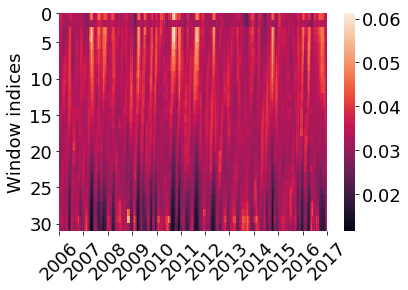}
    \caption{Attention weights for head 2}
    \end{subfigure}
    \begin{subfigure}[t]{.245\textwidth}
  \centering
    \includegraphics[width=1\linewidth]{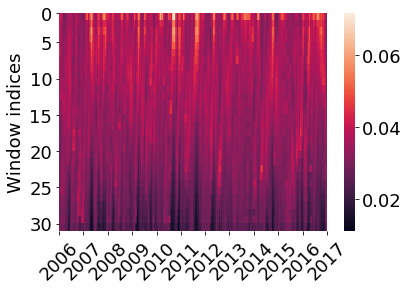}
    \caption{Attention weights for head 3}
    \end{subfigure}
    \begin{subfigure}[t]{.245\textwidth}
  \centering
    \includegraphics[width=1\linewidth]{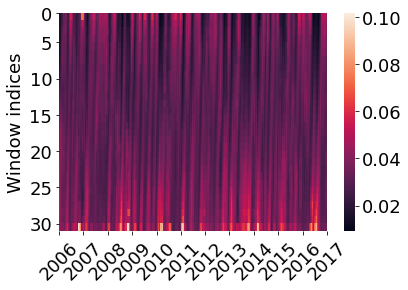}
    \caption{Attention weights for head 4}
    \end{subfigure}
     \bnotefig{These figures illustrate the out-of-sample behavior from 2006 to 2016 of the CNN+Transformer benchmark model for a single residual time-series. The cumulative residual returns are plotted in (a), and the suggested allocation weights before cross-sectional normalization are plotted in (c). The attention head weights (d)--(g) quantify the activation for each attention head over time. Subplot (b) shows the average of these weights over the four heads for different times. All time-series have been smoothed using a simple moving average with a 30-day window for better presentation. The empirical benchmark model is the CNN+Transformer model based on IPCA 5-factor residuals. We estimate the model on only once on the first $T_{\text{train}}$=8 years based on the Sharpe ratio objective.}
\end{figure}

In Figure \ref{fig:fullInterpretation} we generalize the analysis of Figure \ref{fig:windowIPCA5} to study the model structure of the benchmark model over time. While Figure \ref{fig:windowIPCA5} represents a ``snapshot'' for one point in time, we now display the allocation weights and attention head weights of a single representative residual for different times. Subfigure \ref{fig:fullInterpretation}(a) shows the cumulative residual time-series. For a specific point in time we use the lagged $L=30$ days as an input to obtain the allocation weights and attention head weights for that time. The out-of-sample allocation weights correctly change the directions to exploit the patterns in the residual time-series. The attention head weights over time offer additional insights into the structure of the global patterns. Each vertical slice from window index 1 to window index 30 displays the normalized attention weights for the time point under the slice. The third attention head, which was displayed in green in Figures \ref{fig:simulated-attention-weights} and \ref{fig:windowIPCA5} has the largest values during ``up-patterns'' of the residual, for example for 2007, 2010 and 2012. Importantly, the attention weights focus on the early days within the 30-day window. This confirms our previous interpretation as an ``early reversal'' pattern. Attention head four, which was previously represented as a red line, has the highest values during down-times, such as 2009, 2014 and the middle of 2016. In contrast to attention head 3, this head focuses on the immediate past within the local window. Attention head 1, which is a dampened down-version, focuses more uniformly on all the values within the local window.

\begin{figure}[h!]
  \tcapfig{Variable Importance for Allocation Weight}
  \label{fig:absolute-gradient}
  \begin{subfigure}[t]{.5\textwidth}
  \centering
    \includegraphics[trim={0 0 0 1.1cm},clip,width=1\linewidth]{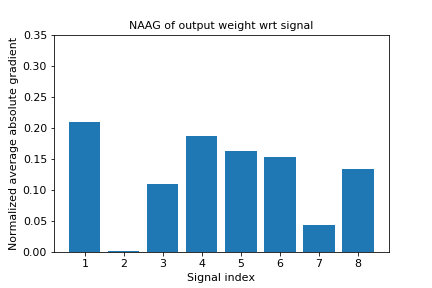}
    \caption{Importance of Local Basic Patterns}
  \end{subfigure}
    \begin{subfigure}[t]{.5\textwidth}
  \centering
    \includegraphics[trim={0 0 0 1.1cm},clip,width=1\linewidth]{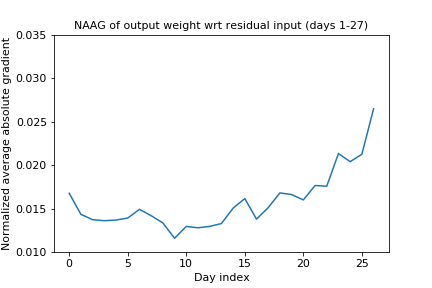}
    \caption{Importance of Residual Days}
  \end{subfigure}
     \bnotefig{These figures show the normalized average absolute gradient (NAAG) of the allocation weight with respect to various inputs to intermediate layers in the CNN+Transformer benchmark network. A higher NAAG indicates a higher importance. Subplot (a) quantifies the importance of the $D=8$ different convolutional filters, that is, we display the gradient with respect to the output of the convolutional network, which is the input to the self-attention layer. In (b), we report the importance of the first 27 days of the input residual time series. Each average absolute gradient is normalized by dividing each element by the sum of all elements. The empirical benchmark model is the CNN+Transformer model based on IPCA 5-factor residuals. We estimate the model on only once on the first $T_{\text{train}}$=8 years based on the Sharpe ratio objective.}
\end{figure}

The average attention weights in (b) illustrate the asymmetric response of the transformer network. During uptrends, it focuses on the residual prices which are further in the past part of the window to decide which to position to take; however, during downtrends, it focuses on the most recent cumulative residual prices in the lookback window, which indicates that it is taking into account the latest data in order to decide what position to take. This indicates that our CNN+Transformer policy network has learned to act swiftly during downtrends, and more slowly during uptrends. This shows that our model learns in particular the commonly repeated wisdom that ``markets take escalators up and elevators down''. This asymmetric policy is a key benefit of the attention-based model, which cannot easily be replicated by the parametric Ornstein-Uhlenbeck or fixed basis pattern benchmark models we compare against. The convolutional subnetwork's patterns provide translation invariant information about what kind of trend is present within each 3-day subwindow of the 30-day cumulative residual price lookback window, which allows the transformer subnetwork to form a stable attention function that results in this unique policy.

Figure \ref{fig:absolute-gradient} sheds further light on which days and patterns are important. The figure shows the normalized average absolute gradient (NAAG) of the allocation weight with respect to various inputs to intermediate layers in the CNN+Transformer benchmark network. A higher NAAG indicates a higher importance. Subplot (a) quantifies the importance of the $D=8$ different basic patterns. We observe that the flat basic pattern 2 has a negligible weight, while basis patterns that are needed for trend or reversal patterns have high importance. In (b), we report the importance of the first 27 days of the input residual time series.\footnote{As the attention head weights are determined relative to the last local 3-day window, that subwindow has mechanically a larger weight and is not comparable to the other 27 days.} Crucially, all previous days matter, which emphasizes that the trading allocation depends on the past dynamics. The most recent 14 days seem to get on average more attention for the trading decisions. However, as indicated in Figure \ref{fig:fullInterpretation}, the importance of the days seems to be asymmetric for different global patterns.

\section{Conclusion}\label{sec:Conclusion}

In this paper, we introduce a unifying conceptual framework to compare different statistical arbitrage approaches based on the decomposition into (1) arbitrage portfolio generation, (2) signal extraction and (3) allocation decision. We develop a novel deep learning statistical arbitrage approach. It uses conditional latent factors to generate arbitrage portfolios. The signal is estimated with a CNN+Transformer, which combines global dependency patterns with local filters. The allocation is estimated with a nonparametric FFN based on a global trading objective.

We conduct a comprehensive empirical out-of-sample study on U.S. equities and demonstrate the potential of machine learning methods in arbitrage trading. Our CNN+Transformer substantially outperforms all benchmark approaches. The implied trading strategies are not spanned by conventional risk factors, including price trend factors, and survive realistic transaction and holding costs. Our model provides insights into optimal trading policies which are based on asymmetric trend and reversion patterns. In particular, our study shows that the trading signal extraction is the most challenging and separating element among different statistical arbitrage approaches. 

Our findings contribute to the debate on efficiency of markets. We quantify the scope of profits that arbitrageurs can achieve in equity markets. Importantly, the substantial profitability of our arbitrage strategies is not inconsistent with equilibrium asset pricing, following similar arguments as in \cite{Gatev}. It could rather be viewed as empirical evidence about how efficiency is maintained in practice. We document non-declining profitability of arbitrage trading over time, which suggests that the profits are compensation for arbitrageurs to enforce the law of one price. 

\singlespacing
\bibliographystyle{ecta}
{\small
\bibliography{bib}
}
\onehalfspacing

\newpage
\appendix
\renewcommand{\theequation}{A.\arabic{equation}}%
\renewcommand{\thefigure}{A.\arabic{figure}} \setcounter{figure}{0}
\renewcommand{\thetable}{A.\Roman{table}} \setcounter{table}{0}

\section{Data}

\subsection{List of the Firm-Specific Characteristics}

\begin{table}[htb]
\tcaptab{~\textbf{Firm Characteristics by Category}}
\bigskip
\label{tab:category}
\footnotesize
\begin{tabular}{lllllll}

\toprule
     & \multicolumn{2}{l}{{\ul \textbf{Past Returns}}}                                                        &  &      & \multicolumn{2}{l}{{\ul \textbf{Value}}}                                                                       \\
(1)  & r2\_1                        & Short-term momentum                                   &  & (26) & A2ME                             & Assets to market cap                                     \\
(2)  & r12\_2                       & Momentum                                              &  & (27) & BEME                             & Book to Market Ratio                                     \\
(3)  & r12\_7                       & Intermediate momentum                                 &  & (28) & C                                & Ratio of cash and short-term \\
  &                        &                                &  & &                                 & investments to total assets \\
(4)  & r36\_13                      & Long-term momentum                                    &  & (29) & CF                               & Free Cash Flow to Book Value                             \\
(5)  & ST\_Rev                      & Short-term reversal                                   &  & (30) & CF2P                             & Cashflow to price                                        \\
(6)  & LT\_Rev                      & Long-term reversal                                    &  & (31) & D2P                              & Dividend Yield                                           \\
     &                              &                                                       &  & (32) & E2P                              & Earnings to price                                        \\
     & \multicolumn{2}{l}{{\ul \textbf{Investment}}}                                                      &  & (33) & Q                                & Tobin's Q                                                \\
(7)  & Investment                   & Investment                                            &  & (34) & S2P                              & Sales to price                                           \\
(8)  & NOA                          & Net operating assets                                  &  & (35) & Lev                              & Leverage                                                 \\
(9)  & DPI2A                        & Change in property, plants, and              &  &      &                                  &                                                          \\
  &                         &  equipment             &  &      &                                  &                                                          \\
(10) & NI                           & Net Share Issues                                      &  &      & \multicolumn{2}{l}{{\ul \textbf{Trading Frictions}}}                                                          \\
     &                              &                                                       &  & (36) & AT                               & Total Assets                                             \\
     & \multicolumn{2}{l}{{\ul \textbf{Profitability}}}                                                       &  & (37) & Beta                             & CAPM Beta                                                \\
(11) & PROF                         & Profitability                                         &  & (38) & IdioVol                          & Idiosyncratic volatility                                 \\
(12) & ATO                          & Net sales over lagged net operating assets            &  & (39) & LME                              & Size                                                     \\
(13) & CTO                          & Capital turnover                                      &  & (40) & LTurnover                        & Turnover                                                 \\
(14) & FC2Y                         & Fixed costs to sales                                  &  & (41) & MktBeta                          & Market Beta                                              \\
(15) & OP                           & Operating profitability                               &  & (42) & Rel2High                         & Closeness to past year high                              \\
(16) & PM                           & Profit margin                                         &  & (43) & Resid\_Var                       & Residual Variance                                        \\
(17) & RNA                          & Return on net operating assets                        &  & (44) & Spread                           & Bid-ask spread                                           \\
(18) & ROA                          & Return on assets                                      &  & (45) & SUV                              & Standard unexplained volume                              \\
(19) & ROE                          & Return on equity                                      &  & (46) & Variance                         & Variance                                                 \\
(20) & SGA2S                        & Selling, general and administrative  &  &      &                                  &                                                          \\
 &                         &  expenses to sales &  &      &                                  &                                                          \\
(21) & D2A                          & Capital intensity                                     &  &      &                                  &                                                          \\
     &                              &                                                       &  &      &                                  &                                                          \\
     & \multicolumn{2}{l}{{\ul \textbf{Intangibles}}}                                                          &  &      &                                  &                                                          \\
(22) & AC                           & Accrual                                               &  &      &                                  &                                                          \\
(23) & OA                           & Operating accruals                                    &  &      &                                  &                                                          \\
(24) & OL                           & Operating leverage                                    &  &      &                                  &                                                          \\
(25) & PCM                          & Price to cost margin                                  &  &      &                                  &                                                     \\
\bottomrule    
\end{tabular}
\bnotetab{This table shows the 46 firm-specific characteristics sorted into six categories. More details on the construction are in the Internet Appendix of \cite{DeepMarkus}.} 
\end{table}

\section{Implementation of Different Models}

\subsection{Feedforward Neural Network (FFN)}\label{app:ffn}

\begin{figure}[h!]
  \centering
  \caption{Feedforward Network Architecture}\label{fig:FFT+FFN}
    \includegraphics[width=1\linewidth]{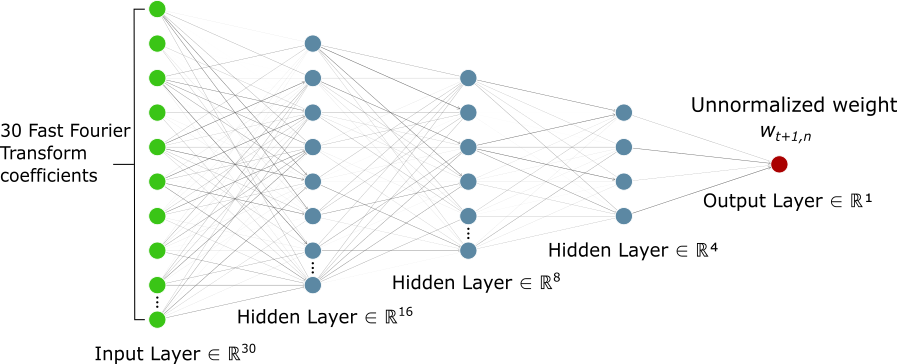}
\end{figure}

In the Fourier+FFN and FFN models, we utilize a feedforward network with $L^{\text{FFN}}$ layers as illustrated in Figure \ref{fig:FFT+FFN}. Each hidden layer takes the output from the previous layer and transforms it into an output as 
\begin{align*}
x^{(l)} &= \text{ReLU}\left( {W^{\text{FFN},(l-1)\top}} x^{(l-1)} + w_0^{\text{FFN},(l-1)} \right) = \text{ReLU}\left( w_0^{\text{FFN},(l-1)} + \sum_{k=1}^{K^{(l-1)}} w_k^{\text{FFN},(l-1)} x^{(l-1)}_k \right)
\\
y&= W^{\text{FFN},(L^{\text{FFN}})\top} x^{(L)} + w_0^{\text{FFN},(L^{\text{FFN}})}
\end{align*}
with hidden layer outputs $x^{(l)}=(x_1^{(l)},...,x_{K^{(l)}}^{(l)}) \in \mathbbm R^{K^{(l)}}$, parameters $W^{(\text{FFN},l)}=(w_1^{\text{FFN},(l)},...,w_{K^{(l)}}^{\text{FFN},(l)}) \in \mathbbm R^{K^{(l)} \times K^{(l-1)}}$ for $l=0,...,L^{\text{FFN}}-1$ and $W^{\text{FFN},(L^{\text{FFN}})} \in \mathbbm R^{K^{(L)}}$, and where $
\text{ReLU}(x_k)=\max(x_k,0). $

\subsection{Ornstein-Uhlenbeck Model}\label{app:OU}

Following \cite{avelee} and \cite{papayeo} we model $X_t$ as an Ornstein-Uhlenbeck (OU) process  
 \begin{align*}
 d X_t = \kappa \left( \mu  - X_t \right) dt + \sigma d B_t
 \end{align*}   
 for a Brownian motion $B_t$. As the analytical solution of the above stochastic differential equation is 
$$X_{t+\Delta t}=(1-e^{-\kappa\Delta t})\mu+e^{-\kappa\Delta t}X_t+ \sigma\int_t^{t+\Delta t}e^{-\kappa (t+\Delta t -s)} dB_s$$
for any $\Delta t$, we can without loss of generality set $\Delta t=1$, and estimate the parameters $\kappa,\mu$ and $\sigma$ from the AR(1) model
$$X_{t+1}=a+bX_t+e_t,$$
where each $e_t$ is a normal, independent and identically distributed random variable with mean 0. The parameters are estimated with a standard linear regression, which yields
\begin{align*}
\hat \kappa = -\frac{\log(\hat b)}{\Delta t},\quad \hat \mu = \frac{\hat a}{1-\hat b},\quad  \frac{\hat \sigma}{\sqrt{2 \hat \kappa}}  =\sqrt{\frac{\hat \sigma_e^2}{1-\hat b^2}}.
\end{align*}
The strategy depends on the ratio $\frac{X_L - \hat \mu}{\hat \sigma \sqrt{2 \hat \kappa}}$. Note that this is only defined for $b<1$ which is equivalent to parameter restrictions that the OU process is mean-reverting. The trading policy depends on the thresholds $c_{\text{thresh}}$ and $c_{\text{crit}}$, which are hyperparameters. These hyperparameters are selected on the validation data from the candidate values $c_{\text{thresh}}\in \{1,1.25,1.5\}$ and $c_{\text{crit}} \in \{0.25,0.5,0.75\}$. Our benchmark model has the values $c_{\text{thresh}}=1.25$ and $c_{\text{crit}}=0.25$, which coincides with the optimal values  in \cite{avelee} and \cite{papayeo}.
 

\subsection{Convolutional Neural Network with Transformer}\label{app:CNNtrans}

\subsubsection{Convolutional Neural Network}\label{app:CNN}

In our empirical application, we consider a 2-layered convolutional network with some standard technical additions. The network takes as input a window $x^{(0)}=x \in \R^L$ of $L$ consecutive daily cumulative returns or log prices of a residual, and outputs the feature matrix $\tilde{x}\in\R^{L\times F}$ given by computing the following quantities for $l=1,\ldots,L, d=1\ldots, D$

    \begin{equation}\label{eq:Conv1}
        y^{(0)}_{l,d} = b^{(0)}_d + \sum_{m=1}^{\ds} W^{(0)}_{d,m}x^{(0)}_{l-m+1},
        \quad
        x^{(1)}_{l,d} = \text{ReLU}\left( 
            \frac{y^{(0)}_{1,d} - \mu^{(0)}_d}{\sigma^{(0)}_d}
        \right).
    \end{equation}
    \begin{equation}\label{eq:Conv2}
        y^{(1)}_{l,d} = b^{(1)}_d+ \sum_{m=1}^{\ds}\sum_{j=1}^D W^{(1)}_{d,j,m}x^{(1)}_{l-m+1,j},
        \quad 
         x^{(2)}_{l,d}=\text{ReLU}\left( 
         \frac{y^{(1)}_{l,d}-\mu^{(1)}_d}{\sigma^{(1)}_d}
         \right),
    \end{equation}
    \begin{equation}\label{eq:residualConnection}
        \tilde{x}_{l,d} = x^{(2)}_{l,d} +  x^{(0)}_{l},
    \end{equation}
    
    where 
    \begin{equation*}
    \mu^{(i)}_k = \frac{1}{L}\sum_{l=1}^L y^{(i)}_{l,k},
    \quad
    \sigma^{(i)}_k = \sqrt{\frac{1}{L}\sum_{l=1}^L\left(y^{(i)}_{l,k} -\mu^{(i)}_k\right)^2}.
    \end{equation*}
    and $b^{(0)},b^{(1)}\in\R^D,$ $W^{(0)}\in\R^{D\times \ds}$ and $W^{(1)}\in\R^{D\times D \times \ds}$ are parameters to be estimated. Compared with the simple convolutional network introduced in the main text, the previous equations incorporate three standard technical improvements commonly used in deep learning practice. First, they include ``bias terms'' $b^{(i)}$ in the first part of equations \ref{eq:Conv1} and \ref{eq:Conv2} to allow for more flexible modeling. Second, they include so-called ``instance normalization'' before each activation function to speed up the optimization and avoid vanishing gradients caused by the saturation of the $\text{ReLU}$ activations. Third, they include a ``residual connection'' in equation \ref{eq:residualConnection} to facilitate gradient propagation during training.

\subsubsection{Transformer Network}\label{app:Transformer}

The benchmark model in our empirical application is a one-layer transformer, following the implementation of the seminal paper of \cite{attention}. First, the sequence of features $\tilde{x}\in\R^{L\times D}$ is projected onto $D/H$-dimensional subspaces (called the ``attention heads'') for an integer $H$ dividing $D$, obtaining, for $1\leq i\leq H$, 
\begin{equation*}
 V_i = \tilde{x}W^V_i+b^V_i\in\R^{L\times D/H},\quad K_i = \tilde{x}W^K_i+b^K_i\in\R^{L\times D/H},\quad  Q_i = \tilde{x}W^Q_i+b^Q_i \in\R^{L\times D/H},
\end{equation*}
where $W^V_i,W^K_i,W^Q_i\in\R^{D \times F/H}$, $b^V_i,b^K_i,b^Q_i\in\R^{D/H}$ are parameters to be estimated.
Next, each projection $V_i$ is processed temporally obtaining the hidden states $h_i\in \R^{L\times D/H}$, with $$
     h_{i,l} = \sum_{j=1}^Lw_{l,j,i}V_{i,j}\in\R^{D/H}, \quad w_{l,j,i}=\frac{\exp(K_{i,l}\cdot Q_{i,j})}{\sum_{m=1}^L\exp(K_{i,l}\cdot Q_{i,m} )}\in[0,1]   \quad \text{$l=1,...,L$ and $i=1,...,D/H$}.  $$
These states are then concatenated and linearly combined, obtaining the last hidden state
$$h=\mathrm{Concat}(h_1, ..., h_h)W^O +b^O\in\R^{L\times D},$$
where $W^O\in\R^{F\times F},b^O\in\R^F$ are parameters to be estimated.

Finally, $h$ is normalized and processed time-wise though a 2-layered feedforward network as described in detail in the original paper (\cite{attention}). The number of hidden units in the intermediate layer is a technical hyperparameter that we call HDN in section \ref{sec:hyperparameters}. This network also has dropout regularization with hyperparameter called DRP in section \ref{sec:hyperparameters}.

\subsection{Network Estimation Details}\label{sec:learning}    

As explained in Section \ref{sec:arbitragetrading}, we estimate the parameters of the models with neural networks by solving the optimization problems introduced in equation (\ref{doubleOptimization}) or in equation (\ref{eq:opt-problem}) of section \ref{sec:arbitragetrading}, depending on the model and the objective function. 
In all cases, this is done by replacing the mean and variance by their annualized sample counterpart over a training set, and by finding the optimal network parameters with stochastic gradient descent using PyTorch's Adam optimizer and the optimization hyperparameters learning rate and number of optimization epochs described in detail in section \ref{sec:hyperparameters}.

As mentioned in Section \ref{sec:implementation}, our main results use rolling windows of 1,000 days as training sets. The networks are reestimated every 125 days to strike a balance between computational efficiency and adaptation to changing economic conditions, and the strategies' returns are always obtained out-of-sample. Additionally, to be able to train our model over these long windows without running into memory issues, we split each training window into temporal ``batches'', as is commonly done in deep learning applications. Each batch contains the returns and residuals for all the stocks in a subwindow of $125$ days of the original training window, with the subwindows being consecutive and non-overlapping (i.e., for a training window of 1000 days, we split it into the subwindow containing the first 125 days, the subwindow containing the days between the 126th day and the 250th day, etc.). The optimization process is applied successively to each batch, completing the full sequence of batches before starting a new optimization iteration or epoch.

In the implementation of our optimization procedure under market frictions, we found it useful to include the last allocation as an additional input to the allocation function $\wfunc$, as the inclusion of the cost term makes the objective function depend on it. However, the inclusion of the previous allocations in either the objective function or the architecture of the model complicates the parallelization of the training and evaluation computations, because after this change the model requires the output of previous lookback window in order to compute the output of the current window. 
To allow training to remain parallelized, which is desirable for reasonable computational speed given the volume of data of our empirical application, in our implementation of the training function in each epoch, we take the previous allocations from the output of the previous epoch and use them as a pre-computed approximation of the allocations for the current epoch. This approximation converges in our empirical experiments and allows us to maintain parallelization, but may produce suboptimal results. For evaluation purposes, however, everything is computed exactly and with no approximations using a sequential approach.

Throughout section \ref{sec:empirical}, all presented results have been computed with PyTorch 1.5 and have been parallelized across 8 NVIDIA GeForce GTX Titan V GPUs, on a server with two Intel Xeon E5-2698v3 32-core CPUs and 1 TB of RAM. The full rationale for the hyperparameter choices are described in detail in section \ref{sec:hyperparameters}, but for a CNN+Transformer model with a lookback window of 30 days, 8 convolutional filters with a filter size of 2, 4 attention heads, 125-day reestimation using a rolling lookback window of 1000, it takes our deep model approximately 7 hours to be periodically estimated and run in our 19 years of daily out-of-sample data with our universe of on average $\sim$550 stocks per month.

\section{Additional Empirical Results}

\subsection{Robustness to Hyperparameter Selection}\label{sec:hyperparameters}

In this subsection, we describe our hyperparameter selection procedure and explore additional hyperparameter choices to show that the performance of our strategies is extremely robust to our choices. These results complement the time stability checks we exhibited in Section \ref{sec:time-stability}. To decide which hyperparameters we would select for use in our network, we  fixed a validation dataset as follows: we took the first 1000 trading days of our data set of residuals (all trading days from January 1, 1998 through December 31, 2001) of the 5-factor IPCA-based model, which is estimated with a 20-year rolling window. Because it is solely used for training in our rolling train/test procedures used to compute strategy returns, this data is completely in-sample, and thus completely avoids look-ahead bias which would influence any of our out-of-sample trading results in the main text. We started with a reasonable set for our hyperparameters, and tested also additional points adjacent to these sets.\footnote{Note the computation over a large set of hyperparameters is computationally infeasible, which requires us to restrict the set to reasonable values.} For each model represented by a point on the grid, we trained the model using the Sharpe ratio objective on the first 750 days of the 1000 trading days, and evaluated it by its out-of-sample Sharpe ratio on the last 250 days of the 1000 trading days. We tested 16 combinations of hyperparameters, which are illustrated in Table \ref{tab:hyperparameter}. The results of our test on the last 250 days of our validation data are displayed in Table \ref{tab:grid-search}. 

The results in Table \ref{tab:grid-search} show that all Sharpe ratios fall within a tight range of values, which is roughly $[3.5, 4.2]$. Means and volatilities concentrate similarly, falling within $[13\%, 17.8\%]$ and $[3.6\%, 4.3\%]$. Computation of 95\% bootstrapped confidence intervals for mean return shows that all models' confidence intervals contain the interval $[10\%, 20\%]$, with volatilies similarly contained. Hence, these models are statistically not distinguishable. Given the statistical insignificance of the differences in performance of these models, we chose the model displayed in Table \ref{tab:hyperparameter}, which is the most parsimonious one, that is it has the smallest number of parameters, and hence benefits low GPU memory usage and ease of interpretability.


\begin{table}[t!]
\centering
\tcaptab{~Hyperparameter options for the network in the empirical analysis}
\label{tab:hyperparameter}
{\small
\begin{tabular}{r|l|c|c}
\toprule
Notation & Hyperparameters & Candidates & Chosen \\
\midrule
$D$ & Number of filters in the convolutional network  & 8, 16 &  8\\
ATT & Number of attention heads & 2, 4 & 4 \\
HDN & Number of hidden units in the transformer's linear layer & 2D, 3D & 2D \\
DRP & Dropout rate in the transformer & 0.25, 0.5 &0.25 \\
$D_{size}$ & Filter size in the convolutional network & 2 & 2 \\
LKB & Number of days in the residual lookback window & 30 & 30\\
WDW & Number of days in the rolling training window & 1000 & 1000 \\
RTFQ & Number of days of the retraining frequency & 125 & 125 \\
BTCH & Batch size, in days & 125 & 125 \\
LR & Learning rate & 0.001 & 0.001 \\
EPCH & Number of optimization epochs & 100 & 100 \\
OPT & Optimization method & Adam & Adam\\
\bottomrule
\end{tabular}
\bnotetab{ This table shows the parameters for our network architecture with respect to the Sharpe ratio on our validation data and the candidates we tried In DRP, we follow the convention that the dropout rate $p$ is the proportion of units which are removed.}
}
\end{table}

\begin{table}[t!]
\centering
\tcaptab{~Performance of candidate models on the last year of the validation data set}
\label{tab:grid-search}
{\small
\begin{tabular}{cccc|ccc}
\toprule
 $D$ &  ATT &  HDN &  DRP & SR &  $\mu$ &  $\sigma$ \\
\midrule
  8 &    2 &  2 &     0.25 &   3.81 & 16.3\% & 4.3\% \\
  8 &    2 &  2 &     0.50 &   3.92 & 16.0\% & 4.1\% \\
  8 &    2 &  3 &     0.25 &   3.79 & 16.2\% & 4.3\% \\
  8 &    2 &  3 &     0.50 &   4.00 & 16.4\% & 4.1\% \\
  8 &    4 &  2 &     0.25 &   3.81 & 15.6\% & 4.1\% \\
  8 &    4 &  2 &     0.50 &   4.13 & 17.8\% & 4.3\% \\
  8 &    4 &  3 &     0.25 &   3.82 & 15.6\% & 4.1\% \\
  8 &    4 &  3 &     0.50 &   4.16 & 17.4\% & 4.2\% \\
 16 &    2 &  2 &     0.25 &   4.00 & 14.8\% & 3.7\% \\
 16 &    2 &  2 &     0.50 &   4.06 & 16.2\% & 4.0\% \\
 16 &    2 &  3 &     0.25 &   4.11 & 14.9\% & 3.6\% \\
 16 &    2 &  3 &     0.50 &   4.06 & 16.6\% & 4.1\% \\
 16 &    4 &  2 &     0.25 &   3.93 & 15.6\% & 4.0\% \\
 16 &    4 &  2 &     0.50 &   3.66 & 13.9\% & 3.8\% \\
 16 &    4 &  3 &     0.25 &   4.18 & 16.8\% & 4.0\% \\
 16 &    4 &  3 &     0.50 &   3.51 & 13.0\% & 3.7\% \\
\bottomrule
\end{tabular}
\bnotetab{ This table shows the model performance with respect to the Sharpe ratio, mean, and volatility on our validation data set for the candidate models implied by Table \ref{tab:hyperparameter}. The models are trained on the first three years of the validation data set (1998--2000) and tested on the last year (2001). In DRP, we follow the convention that the dropout rate $p$ is the proportion of units which are removed.}
}
\end{table}

To ensure that our results are stable across several choices of hyperparameters, we study the results of four additional models with perturbed hyperparameters. This complements our robustness results of Section \ref{sec:time-stability} regarding the size of the lookback window and the retraining frequency. The four additional networks and their hyperparameter configurations are listed in Table \ref{tab:robustness-results}, with Network 1 being the network studied throughout this empirical section. Network 2 corresponds to more filters and commensurately more hidden units to consume them, and higher dropout rates to more strongly regularize the additional parameters. Network 3 halves the number of attention heads from our original specification. Networks 4 and 5 modify the size of the rolling training window from 1000 trading days to 1250 and 750 trading days, respectively, which corresponds closely to three and five calendar years. These additional hyperparameter configurations constitute local perturbations in hyperparameter space, to which our strategies' performance are relatively robust.

\begin{table}[t!]
\tcaptab{Alternative best performing models on the data from 2002--2016}\label{tab:robustness-results}
\small
\centering
\begin{tabular}{c|cccccccc}
\toprule
Model & FLNB & FLSZ & ATT & HDN & DRP & LKB & WDW \\
\midrule
Network 1 & [1,8] & 2 & 4 & 16 & 0.25 & 30 & 1000 \\
Network 2 & [1,16] & 2 & 4 & 32 & 0.5 & 30 & 1000 \\
Network 3  & [1,8] & 2 & 2 & 16 & 0.25 & 30 & 1000 \\
Network 4 & [1,8] & 2 & 4 & 16 & 0.25  & 30 & 1250 \\
Network 5 & [1,8] & 2 & 4 & 16 & 0.25  & 30 & 750 \\
\bottomrule
\end{tabular}
\bnotetab{This table reports four of the best performing models for our network architecture with respect to the Sharpe ratio on our data from 2002--2016 and the candidates described in Table \ref{tab:hyperparameter}. Our original network, which is studied throughout this section, is labeled as Network 1.}
\end{table}

\begin{table}[t!]
{\small
\tcaptab{Performance of the alternative models on our benchmark residual datasets, 2002--2016}\label{tab:RobustnessResults}
\begin{center}\scalebox{1}{
\begin{tabular}{c||ccc||ccc||ccc}
\toprule
 & \multicolumn{3}{c||}{Fama-French 5} & \multicolumn{3}{c||}{PCA 5} & \multicolumn{3}{c}{IPCA 5}\\
\cmidrule{2-10}
 Model & SR & $\mu$ & $\sigma$ &  SR & $\mu$ & $\sigma$ & SR & $\mu$ & $\sigma$   \\
\midrule
Network 1
& 3.21 & 4.6\% & 1.4\%
& 3.36 & 14.3\% & 4.2\%
& 4.16 & 8.7\% & 2.1\%
\\ 
Network  2
& 3.16 & 4.6\% & 1.4\%
& 3.26 & 13.9\% & 4.3\%
& 4.35 & 8.4\% & 1.9\%\\
Network  3
& 3.30 & 4.8\% & 1.4\%
& 3.17 & 13.4\% & 4.2\%
& 4.00 & 8.4\% & 2.1\%
\\ 
Network  4
& 2.93 & 4.1\% & 1.4\%
& 2.74 & 11.7\% & 4.3\%
& 3.96 & 7.9\% & 2.0\%
\\
Network  5
& 3.13 & 4.9\% & 1.6\%
& 3.52 & 15.0\% & 4.3\%
& 3.77 & 8.6\% & 2.3\%
\\
\bottomrule
\end{tabular}}\end{center}}
\bnotetab{This table shows the average annualized returns, volatilities and Sharpe ratios of our alternative models from Table \ref{tab:robustness-results} on our three benchmark residual datasets, trained with the Sharpe ratio objective function.}
\end{table}

In Table \ref{tab:RobustnessResults}, we report the results of these models on a representative subset of 5-factor models, which are now evaluated on the full out-of-sample data. We see that the Sharpe ratios are broadly similar across all three different perturbations of network architecture hyperparameters (i.e., number of filters, number of attention heads, and dropout rate). The small range of values induced by these choices shows that our network performs similarly over a variety of sensible network parameters, and highlights the efficacy of our reasonable choice of convolutional, attentional, and feedforward subnetworks which specialize in finding small temporal patterns, arranging these patterns throughout time, and deciding on allocations based on these arranged patterns. 


For the allocation function feedforward network (FFN) utilized for the Fourier+FFN model, we choose a reasonable architecture based on deep learning conventions and have verified that the results are robust to this choice. Because the input of the network are the $L=30$ coefficients of the Fourier decomposition of each residual window $(X^{(n,t)}_l)_{1\leq l \leq L}$ and the output is the corresponding allocation weight $w_{n,t}\in\R$, we follow standard deep learning practices and consider 3 hidden layers with dimensions 16,8,4 regularized with a dropout rate of 0.25. We use the ReLU activation function, and train using the same procedure outlined in \ref{sec:learning}, with the same batch size, learning rate, number of optimization epochs, and optimization method as in Table \ref{tab:hyperparameter}.

\newpage

\subsection{Interpretation}\label{app:interpretation}

\begin{figure}[H]
    \tcaptab{Illustrative Example of Allocation Weights and Signals for Different Methods}\label{fig:weights-and-signals}
    \begin{subfigure}[t]{.333\textwidth}
        \centering
        \includegraphics[trim={0 0 0 1.2cm},clip,width=1\linewidth]{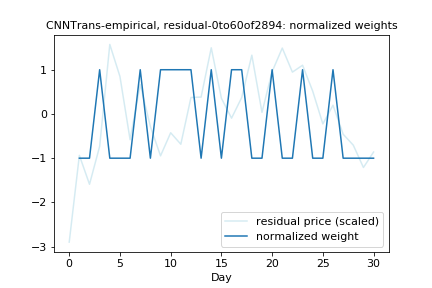}
        \subcaption{Cumulative residuals $x_l$ and allocation weight $w_l^{\epsilon |\text{CNN+Trans}}$}
    \end{subfigure}
    \begin{subfigure}[t]{.333\textwidth}
        \centering
        \includegraphics[trim={0 0 0 1.2cm},clip,width=1\linewidth]{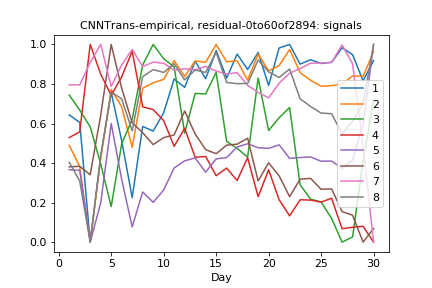}
        \subcaption{Signal $\theta_l^{\text{CNN+Trans}}$}
    \end{subfigure}
    \begin{subfigure}[t]{.333\textwidth}
        \centering
        \includegraphics[trim={0 0 0 1.2cm},clip,width=1\linewidth]{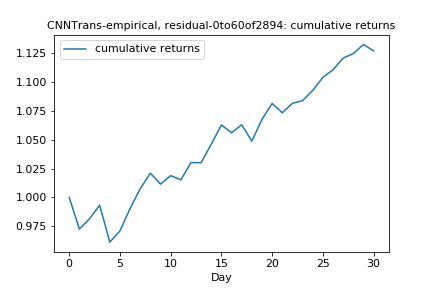}
        \subcaption{Cumulative returns of CNN+Trans strategy}
    \end{subfigure}
    \begin{subfigure}[t]{.333\textwidth}
        \centering
        \includegraphics[trim={0 0 0 1.2cm},clip,width=1\linewidth]{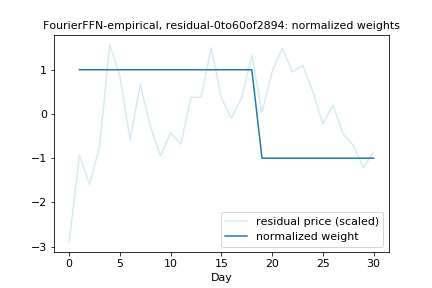}
        \subcaption{Cumulative residual $x_l$ and allocation weight $w_l^{\epsilon |\text{FFT}}$}
    \end{subfigure}
    \begin{subfigure}[t]{.333\textwidth}
        \centering
        \includegraphics[trim={0 0 0 1.1cm},clip,width=1\linewidth]{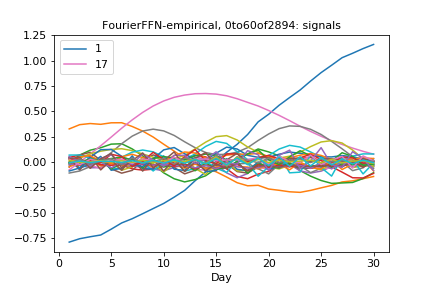}
        \subcaption{Signal $\theta_l^{\text{FFT}}$}
    \end{subfigure}
    \begin{subfigure}[t]{.333\textwidth}
        \centering
        \includegraphics[trim={0 0 0 1.1cm},clip,width=1\linewidth]{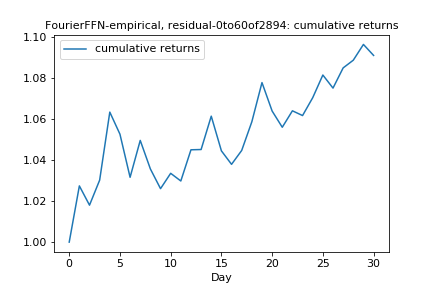}
        \subcaption{Cumulative returns of Fourier+FFN strategy}
    \end{subfigure}

    \begin{subfigure}[t]{.333\textwidth}
        \centering
        \includegraphics[trim={0 0 0 1.2cm},clip,width=1\linewidth]{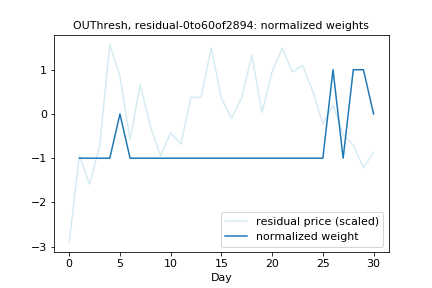}
        \subcaption{Cumulative residual $x_l$ and allocation weight $w_l^{\epsilon |\text{OU}}$}
    \end{subfigure}
    \begin{subfigure}[t]{.333\textwidth}
        \centering
        \includegraphics[trim={0 0 0 1.2cm},clip,width=1\linewidth]{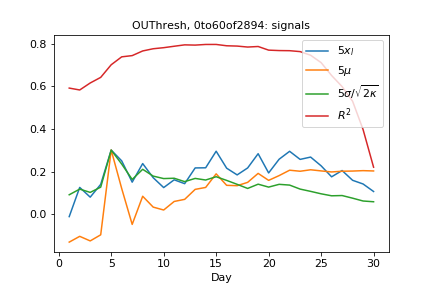}
        \subcaption{Signal $\theta_l^{\text{OU}}$}
    \end{subfigure}
    \begin{subfigure}[t]{.333\textwidth}
        \centering
        \includegraphics[trim={0 0 0 1.2cm},clip,width=1\linewidth]{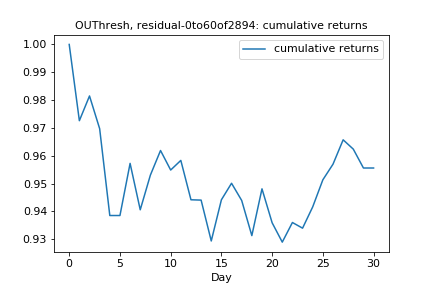}
        \subcaption{Cumulative returns of OU+Thresh strategy}
    \end{subfigure}
    \bnotetab{These plots are an illustrative example of the allocation weights and signals of the Ornstein-Uhlenbeck with Threshold (OU+Thres), Fast Fourier Transform (FFT) with Feedforward Neural Network (FFN), and Convolutional Neural Network (CNN) with Transformer models for a specific cumulative residual. The models are estimated on the empirical data, and the residual is a representative empirical example. In more detail, we consider the residuals from five IPCA factors and estimate the benchmark models as explained in Section \ref{sec:interpretation}. The left subplots display the cumulative residual process along with the out-of-sample allocation weights $w_l^{\epsilon|\cdot}$ that each model assigns to this specific residual. In this example, we consider trading only this specific residual and hence normalize the weights to $\{-1,0,1\}$. The middle column plots show the time-series of estimated out-of-sample signals for each model, by applying the $\theta_l^{\cdot}$ arbitrage signal function to the previous $L$ cumulative returns of the residual. The right column plots display the out-of-sample cumulative returns of trading this particular residual based on the corresponding allocation weights. We use a rolling lookback window of $L=30$ days to estimate the signal and allocation, which we evaluate for the out-of-sample trading on the next 30 days. The plots only show the out-of-sample period. The evaluation of this illustrative example is a simplification of the general model that we use in our empirical main analysis, where we trade all residuals and map them back into the original stock returns.} 
\end{figure}    

\begin{figure}[H]
    \tcaptab{Additional Examples of Allocation Weights and Signals}\label{fig:example-trading-residuals-appendix}
    \begin{subfigure}[t]{.333\textwidth}
        \centering
        \includegraphics[trim={0 0 0 1.2cm},clip,width=1\linewidth]{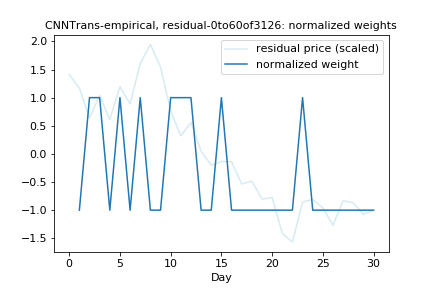}
        \subcaption{Cumulative residuals $x_l$ and allocation weight $w_l^{\epsilon |\text{CNN+Trans}}$}
    \end{subfigure}
    \begin{subfigure}[t]{.333\textwidth}
        \centering
        \includegraphics[trim={0 0 0 1.2cm},clip,width=1\linewidth]{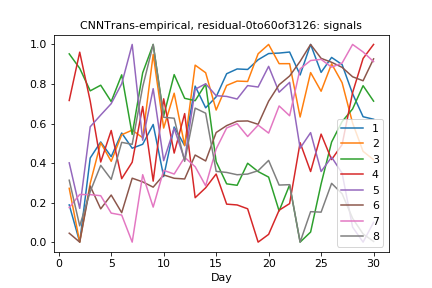}
        \subcaption{Signal $\theta_l^{\text{CNN+Trans}}$}
    \end{subfigure}
    \begin{subfigure}[t]{.333\textwidth}
        \centering
        \includegraphics[trim={0 0 0 1.2cm},clip,width=1\linewidth]{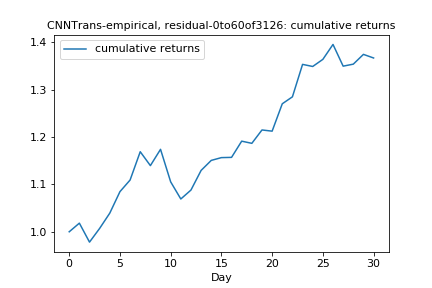}
        \subcaption{Cumulative returns of CNN+Trans strategy}
    \end{subfigure}
    
    \begin{subfigure}[t]{.333\textwidth}
        \centering
        \includegraphics[trim={0 0 0 1.2cm},clip,width=1\linewidth]{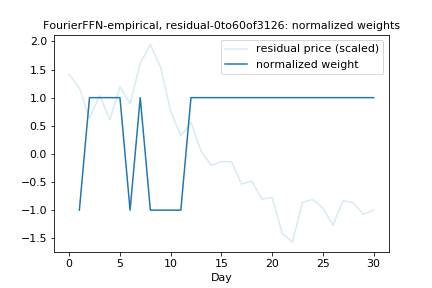}
        \subcaption{Cumulative residual $x_l$ and allocation weight $w_l^{\epsilon |\text{FFT}}$}
    \end{subfigure}
    \begin{subfigure}[t]{.333\textwidth}
        \centering
        \includegraphics[trim={0 0 0 1.1cm},clip,width=1\linewidth]{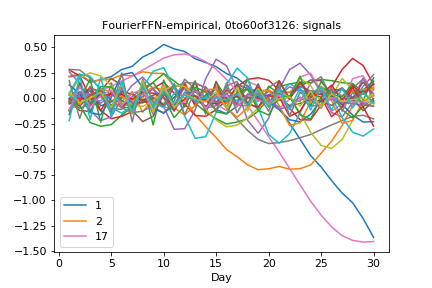}
        \subcaption{Signal $\theta_l^{\text{FFT}}$}
    \end{subfigure}
    \begin{subfigure}[t]{.333\textwidth}
        \centering
        \includegraphics[trim={0 0 0 1.1cm},clip,width=1\linewidth]{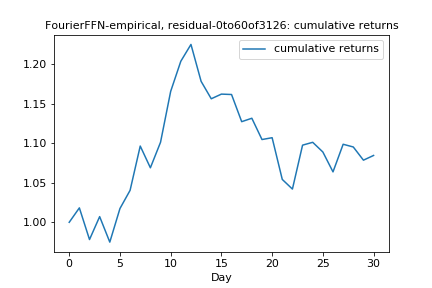}
        \subcaption{Cumulative returns of Fourier+FFN strategy}
    \end{subfigure}

    \begin{subfigure}[t]{.333\textwidth}
        \centering
        \includegraphics[trim={0 0 0 1.2cm},clip,width=1\linewidth]{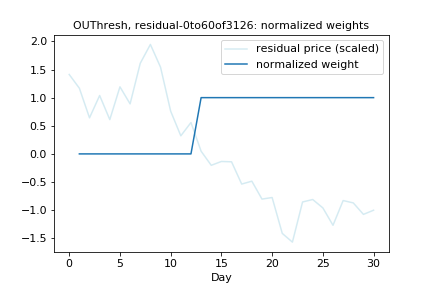}
        \subcaption{Cumulative residual $x_l$ and allocation weight $w_l^{\epsilon |\text{OU}}$}
    \end{subfigure}
    \begin{subfigure}[t]{.333\textwidth}
        \centering
        \includegraphics[trim={0 0 0 1.2cm},clip,width=1\linewidth]{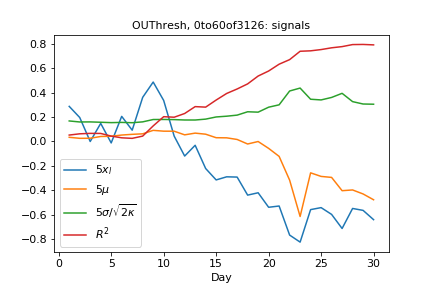}
        \subcaption{Signal $\theta_l^{\text{OU}}$}
    \end{subfigure}
    \begin{subfigure}[t]{.333\textwidth}
        \centering
        \includegraphics[trim={0 0 0 1.2cm},clip,width=1\linewidth]{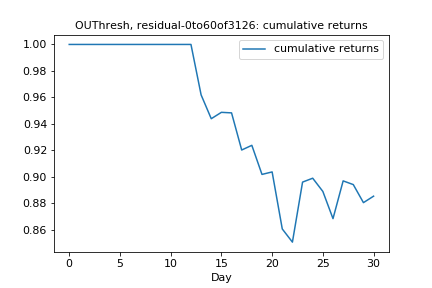}
        \subcaption{Cumulative returns of OU+Thres strategy}
    \end{subfigure}
    \bnotetab{These plots are an illustrative example of the allocation weights and signals of the Ornstein-Uhlenbeck with Threshold (OU+Thres), Fast Fourier Transform (FFT) with Feedforward Neural Network (FFN), and Convolutional Neural Network (CNN) with Transformer models for a specific cumulative residual. The models are estimated on the empirical data, and the residual is a representative empirical example. In more detail, we consider the residuals from five IPCA factors and estimate the benchmark models as explained in Section \ref{sec:interpretation}. The left subplots display the cumulative residual process along with the out-of-sample allocation weights $w_l^{\epsilon|\cdot}$ that each model assigns to this specific residual. In this example, we consider trading only this specific residual and hence normalize the weights to $\{-1,0,1\}$. The middle column plots show the time-series of estimated out-of-sample signals for each model, by applying the $\theta_l^{\cdot}$ arbitrage signal function to the previous $L$ cumulative returns of the residual. The right column plots display the out-of-sample cumulative returns of trading this particular residual based on the corresponding allocation weights. We use a rolling lookback window of $L=30$ days to estimate the signal and allocation, which we evaluate for the out-of-sample on the next 30 days. The plots only show the out-of-sample period. The evaluation of this illustrative example is a simplification of the general model that we use in our empirical main analysis, where we trade all residuals and map them back into the original stock returns.} 
\end{figure}

\begin{figure}[H]
    \tcaptab{Example Attention Weights for Sinusoidal Residual Inputs}
    \label{fig:simulated-attention-weightsAppendix}
    
    \begin{subfigure}[t]{\textwidth}
        \centering
        \includegraphics[trim={0 0 0 1.2cm},clip,width=1\linewidth]{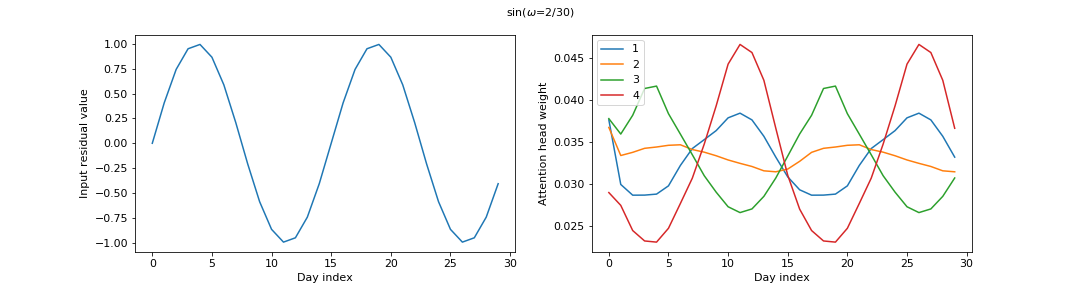}
    \end{subfigure}
    \subcaption{Input residual and attention head weights for $\omega_{\sin}=2/30$}
    
    \begin{subfigure}[t]{\textwidth}
        \centering
        \includegraphics[trim={0 0 0 1.2cm},clip,width=1\linewidth]{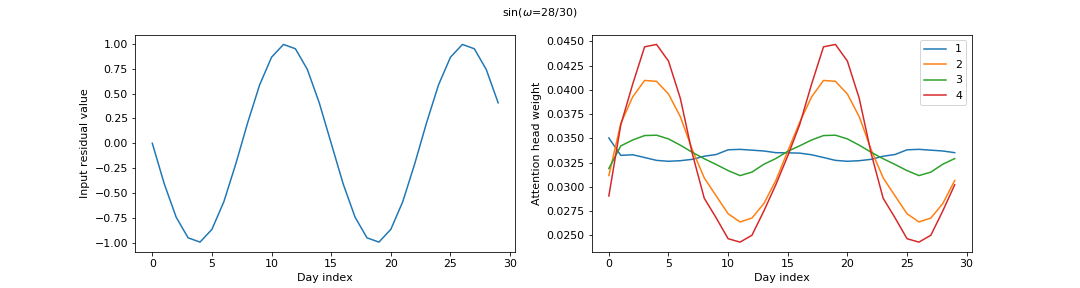}
    \end{subfigure}
    \subcaption{Input residual and attention head weights for $\omega_{\sin}=28/30$}

%
%
%
    \begin{subfigure}[t]{\textwidth}
        \centering
        \includegraphics[trim={0 0 0 1.2cm},clip,width=1\linewidth]{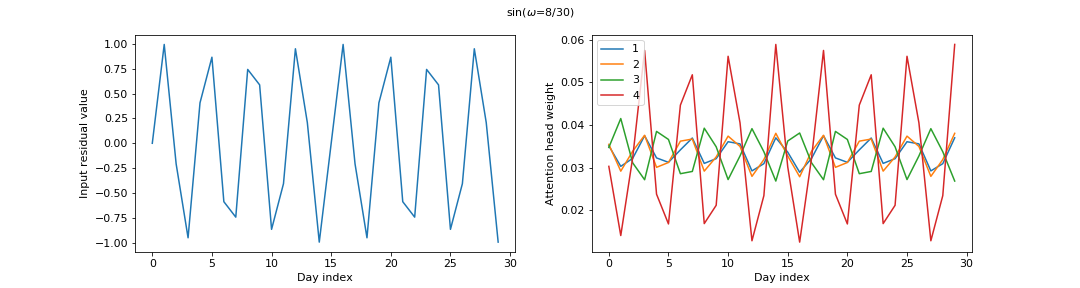}
    \end{subfigure}
    \subcaption{Input residual and attention head weights for $\omega_{\sin}=8/30$}
    
    
    \begin{subfigure}[t]{\textwidth}
        \centering
        \includegraphics[trim={0 0 0 1.2cm},clip,width=1\linewidth]{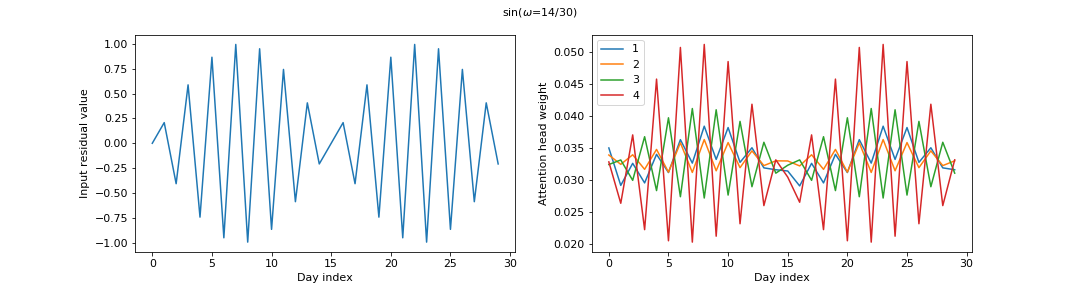}
    \end{subfigure}
    \subcaption{Input residual and attention head weights for $\omega_{\sin}=14/30$}
    
    
    \bnotefig{\scriptsize These plots show the attention head weights of the CNN+Transformer benchmark model for simulated sinusoidal residual input time series. The inputs are $x_l=\sin(2\pi\omega_{\sin} l)$, for various $\omega_{\sin}$ and $l \in \{0,...,29\}$. The right subplot shows the attention weights for the $H=4$ attention heads for the specific residuals. The empirical benchmark model is the CNN+Transformer model based on IPCA 5-factor residuals. We estimate the model on only once on the first $T_{\text{train}}$=8 years based on the Sharpe ratio objective.}
\end{figure}

\subsection{Unconditional Residual Means}

\begin{table}[H]
    \centering
    {\small
    \tcaptab{OOS Annualized Performance of Unconditional Average Residuals}
    \scalebox{1.0}{
    \begin{tabular}{c||ccc||ccc||ccc}
    \toprule
    \multicolumn{10}{c}{Equally Weighted Residuals}\\
    \midrule
     & \multicolumn{3}{c||}{Fama-French} & \multicolumn{3}{c||}{PCA} & \multicolumn{3}{c}{IPCA}\\
    \cmidrule{2-10}
    
     K & SR &$\mu$ & $\sigma$ & SR & $\mu$ & $\sigma$ & SR & $\mu$ &  $\sigma$  \\
    \midrule
    0       &       0.52 &  11.2\% &  21.4\% &       0.52 &  11.2\% &  21.4\% &       0.52 &  11.2\% &  21.4\% \\
    1       &       0.39 &   1.9\% &   4.8\% &  -0.23 &  -0.4\% &  1.5\% &   0.76 &  3.2\% &  4.2\% \\
    3       &       0.18 &   0.7\% &   3.7\% &   0.34 &   0.3\% &  0.9\% &   0.76 &  2.0\% &  2.7\% \\
    5       &       0.22 &   0.8\% &   3.5\% &   0.93 &   0.7\% &  0.7\% &   0.63 &  1.4\% &  2.3\% \\
    8       &      -0.17 &  -0.5\% &   2.9\% &   1.04 &   0.6\% &  0.5\% &   0.66 &  1.4\% &  2.2\% \\
    10      &          - &      - &      - &   0.90 &   0.4\% &  0.5\% &   0.65 &  1.3\% &  2.1\% \\
    15      &          - &      - &      - &   1.08 &   0.4\% &  0.4\% &   0.62 &  1.3\% &  2.0\% \\
    \bottomrule
    \end{tabular}}
    \label{tab:unconditional-mean-results}
    }
    \bnotetab{This table shows the out-of-sample annualized Sharpe ratio (SR), mean return ($\mu$), and volatility ($\sigma$) of equally weighted residuals. We evaluate the out-of-sample arbitrage trading from January 2002 to December 2016. The $K=0$ factor model corresponds to directly using stock returns instead of residuals for the signal and trading policy.} 
\end{table}

\begin{table}[H]
    \centering
    {\footnotesize
    \tcaptab{Significance of Arbitrage Alphas Based on Unconditional Average Residuals}
    \setlength{\tabcolsep}{4pt} 
    \scalebox{0.9}{
    \begin{tabular}{c||ccccc||ccccc||ccccc}
    \toprule
    \multicolumn{16}{c}{Equally Weighted Residuals}\\
    \midrule
     & \multicolumn{5}{c||}{Fama-French} & \multicolumn{5}{c||}{PCA} & \multicolumn{5}{c}{IPCA} \\
    \cmidrule{2-16}
    K & $\alpha$ & $t_\alpha$ & $R^2$ & $\mu$ & $t_\mu$ & $\alpha$ & $t_\alpha$ & $R^2$ & $\mu$ & $t_\mu$ & $\alpha$ & $t_\alpha$ & $R^2$ & $\mu$ & $t_\mu$ \\
    \midrule
    0       &       1.4\% &         1.4 & 97.0\% & 11.2\% &         2.0$^{*}$ &       1.4\% &         1.4 & 97.0\% & 11.2\% &         2.0$^{*}$ &       1.4\% &         1.4 & 97.0\% & 11.2\% &         2.0$^{*}$ \\
    1       &       0.4\% &        0.4 & 36.6\% &  1.9\% &        1.5 &  0.0\% &       0.0 & 25.8\% & -0.4\% &       -0.9 &  0.4\% &         1.1 & 85.0\% & 3.2\% &      2.9$^{**}$ \\
    3       &       0.4\% &         0.4 &  9.6\% &  0.7\% &        0.7 &  0.4\% &         1.9 & 13.1\% &  0.3\% &        1.3 &  0.9\% &       3.3$^{**}$ & 84.1\% & 2.0\% &      2.9$^{**}$ \\
    5       &       0.2\% &        0.2 &  7.0\% &  0.8\% &       0.9 &  0.7\% &      4.2$^{***}$ &  5.9\% &  0.7\% &     3.6$^{***}$ &  0.4\% &          2.0$^{*}$ & 89.4\% & 1.4\% &       2.4$^{*}$ \\
    8       &      -0.6\% &       -0.8 &  0.7\% & -0.5\% &      -0.7 &  0.6\% &      4.5$^{***}$ &  4.1\% &  0.6\% &       4.0$^{***}$ &  0.4\% &        2.1$^{*}$ & 89.3\% & 1.4\% &       2.5$^{*}$ \\
    10      &          - &           - &     - &     - &          - &  0.5\% &      3.8$^{***}$ &  3.0\% &  0.4\% &     3.5$^{***}$ &  0.3\% &         1.9 & 89.4\% & 1.3\% &       2.5$^{*}$ \\
    15      &          - &           - &     - &     - &          - &  0.4\% &      4.3$^{***}$ &  2.0\% &  0.4\% &     4.2$^{***}$ &  0.3\% &         1.6 & 89.0\% & 1.3\% &       2.4$^{*}$ \\
    \bottomrule
    \end{tabular}
    \label{tab:unconditional-mean-results-tests}
    }}
    \bnotetab{This table shows the out-of-sample pricing errors $\alpha$ of cross-sectionally equally weighted residuals relative of the Fama-French 8 factor model and their mean returns $\mu$ for the different arbitrage models and different number of factors $K$ that we use to obtain the residuals. We run a time-series regression of the out-of-sample returns of the arbitrage strategies on the 8-factor model (Fama-French 5 factors + momentum + short-term reversal + long-term reversal) and report the annualized $\alpha$, accompanying t-statistic value $t_\alpha$, and the $R^2$ of the regression. In addition, we report the annualized mean return $\mu$ along with its accompanying t-statistic $t_\mu$. The hypothesis test are two-sided and stars indicate p-values of 5\% ($^{*}$), 1\% ($^{**}$), and 0.1\% ($^{***}$). All results use the out-of-sample daily returns from January 2002 to December 2016.
    }
\end{table}

\subsection{Dependency between Arbitrage Strategies}

\begin{center}
\begin{table}[H]
\tcaptab{Correlations between the Returns of the CNN+Transformer Arbitrage Strategies}\label{tab:corralt}
{\small
\sisetup{table-format=-1.2}   
\centering
\scalebox{0.95}{\begin{tabular}{c|cccccccc}
\toprule
             & Fama-French 3    & PCA 3  & IPCA 3  & Fama-French 5  & PCA 5  & IPCA 5 & PCA 10  & IPCA 10 \\
\midrule
Fama-French 3  
 & 1.00
 & 0.32
 & 0.44
 & 0.62
 & 0.25
 & 0.43
 & 0.21
 & 0.44
\\ 

PCA 3  
 & 0.32
 & 1.00
 & 0.32
 & 0.34
 & 0.62
 & 0.35
 & 0.41
 & 0.36
\\ 

IPCA 3  
 & 0.44
 & 0.32
 & 1.00
 & 0.37
 & 0.28
 & 0.81
 & 0.21
 & 0.75
\\ 

Fama-French 5  
 & 0.62
 & 0.34
 & 0.37
 & 1.00
 & 0.28
 & 0.39
 & 0.23
 & 0.40
\\ 

PCA 5  
 & 0.25
 & 0.62
 & 0.28
 & 0.28
 & 1.00
 & 0.29
 & 0.47
 & 0.31
\\ 

IPCA 5  
 & 0.43
 & 0.35
 & 0.81
 & 0.39
 & 0.29
 & 1.00
 & 0.23
 & 0.84
\\ 

PCA 10  
 & 0.21
 & 0.41
 & 0.21
 & 0.23
 & 0.47
 & 0.23
 & 1.00
 & 0.25
\\ 

IPCA 10  
 & 0.44
 & 0.36
 & 0.75
 & 0.40
 & 0.31
 & 0.84
 & 0.25
 & 1.00
\\ 

\bottomrule
\end{tabular}}
}
\bnotetab{This table reports the correlations of our CNN+Transformer strategies for some representative choices of the factor models. The correlations are calculated with returns of the out-of-sample arbitrage trading from January 2002 to December 2016. The models are calibrated on a rolling window of four years and use the Sharpe ratio objective function. The signals are extracted from a rolling window of $L=30$ days.}
\end{table}
\end{center}

\subsection{Time-Series Signal}\label{app:robustness}
In this appendix, we report the OOS returns of strategies using alternative models for the ablation tests in Section \ref{sec:empirical}. For the FFN feedforward network, we use the same architecture, hyperparameters, optimization settings, etc. as in the Fourier+FFN model utilized throughout the empirical results section and described in Appendix \ref{sec:hyperparameters}. For the OU+FFN model, because the input is the low-dimensional OU signal in $\R^4$, we consider a 3 hidden layer with dimensions 4,4,4 regularized with a dropout rate of 0.25. We use the sigmoid activation function, and estimate it using the same procedure outlined in section \ref{sec:learning}, with the same batch size, learning rate, number of optimization epochs, and optimization method as in Table \ref{tab:hyperparameter}.

\begin{table}[h!]
\begin{center}
    \setlength{\tabcolsep}{4pt}
    \footnotesize
    \tcaptab{ OOS Annualized Performance Based on Sharpe Ratio Objective}
    \label{tab:aux-model-results}
    \scalebox{1.0}{
    \begin{tabular}{c||c||ccc||ccc||ccc}
    \multicolumn{11}{c}{\vspace{-5pt}}\\
    \toprule
     & Factors
     & \multicolumn{3}{c||}{Fama-French} 
     & \multicolumn{3}{c||}{PCA} 
     & \multicolumn{3}{c}{IPCA}\\
    \cmidrule{1-11}
    
     Model
     &K
     & SR &$\mu$ & $\sigma$ 
     & SR &$\mu$ & $\sigma$ 
     & SR &$\mu$ & $\sigma$   \\
    \midrule
    \midrule
    & 0       &       0.50 &  10.6\% &  21.3\% &       0.50 &  10.6\% &  21.3\% &       0.50 &  10.6\% &  21.3\% \\
    & 1       &       0.34 &   0.8\% &   2.3\% &   0.05 &  0.7\% &  11.9\% &   0.60 &  4.8\% &  8.0\% \\
    OU & 3       &       0.16 &   0.2\% &   1.4\% &   0.44 &  3.4\% &   7.8\% &   0.70 &  4.6\% &  6.6\% \\
    + & 5       &       0.17 &   0.2\% &   1.2\% &   0.68 &  4.7\% &   7.0\% &   0.66 &  4.2\% &  6.3\% \\
    FFN & 8       &      -0.34 &  -0.3\% &   1.0\% &      0.51 &     3.1\% &      6.0\% &      0.60 &     3.9\% &     6.2\% \\
    & 10      &          - &      - &      - &   0.26 &  1.3\% &   5.0\% &   0.56 &  3.5\% &  6.2\% \\
    & 15      &          - &      - &      - &   0.31 &  1.4\% &   4.3\% &   0.54 &  3.3\% &  6.1\% \\
    
        \midrule
%
        \midrule
& 0       &       0.57 &  8.8\% &  15.3\% &       0.57 &  8.8\% &  15.3\% &       0.57 &  8.8\% &  15.3\% \\
& 1       &       0.60 &  2.0\% &   3.3\% &   0.53 &  6.2\% &  11.7\% &   1.07 &  6.5\% &  6.1\% \\
& 3       &       1.02 &  2.6\% &   2.6\% &   1.15 &  8.2\% &   7.2\% &   1.50 &  7.6\% &  5.0\% \\
FFN & 5       &       1.32 &  2.3\% &   1.7\% &   1.42 &  9.8\% &   6.9\% &   1.55 &  7.3\% &  4.7\% \\
& 8       &       1.31 &  2.1\% &   1.6\% &      1.05 &     6.4\% &      6.2\% &      1.52 &     7.2\% &     4.7\% \\
& 10      &          - &     - &      - &   0.70 &  3.5\% &   5.0\% &   1.48 &  7.0\% &  4.7\% \\
& 15      &          - &     - &      - &   0.51 &  2.4\% &   4.8\% &   1.68 &  7.5\% &  4.5\% \\
    \bottomrule
    \end{tabular}}
      \bnotetab{This table shows the out-of-sample annualized Sharpe ratio (SR), mean return ($\mu$), and volatility ($\sigma$) of our three statistical arbitrage models for different numbers of risk factors $K$, that we use to obtain the residuals. We use the daily out-of-sample residuals from January 1998 to December 2016 and evaluate the out-of-sample arbitrage trading from January 2002 to December 2016. OU+FFN denotes a parametric Ornstein-Uhlenbeck model to extract the signal, but a flexible feedforward neural network to estimate the allocation function. FFN takes the residuals directly as signals and estimates an allocation function with a feedforward neural network. The deep learning models are calibrated on a rolling window of four years and use the Sharpe ratio objective function. The signals are extracted from a rolling window of $L=30$ days. The $K=0$ factor model corresponds to directly using stock returns instead of residuals for the signal and trading policy. 
} 
    \end{center} 
    \end{table}

    \newpage

\subsection{Trading Friction Results for PCA Residuals}

\begin{table}[h!]
\centering
{\small
\tcaptab{OOS Performance of CNN+Trans with Trading Frictions}\label{tab:deep-resultsFrictionsPCA}
\scalebox{0.95}{
\begin{tabular}{c||ccc||ccc}
\toprule
\multicolumn{7}{c}{PCA factor model}\\
\midrule
  & \multicolumn{3}{c||}{Sharpe ratio} & \multicolumn{3}{c}{Mean-variance}\\
\cmidrule{2-7}


 K & SR &$\mu$ & $\sigma$& SR &$\mu$ & $\sigma$   \\
\midrule
0  
& 0.52 & 8.5\% & 16.3\%
& 0.22 & 2.6\% & 11.9\%
\\ 

1  
& 0.88 & 7.3\% & 8.4\%
& 0.79 & 9.0\% & 11.4\%
\\ 

3  
& 0.90 & 5.7\% & 6.3\%
& 0.62 & 4.7\% & 7.6\%
\\ 

5  
& 0.81 & 4.5\% & 5.6\%
& 0.68 & 4.4\% & 6.4\%
\\ 

10  
& -0.08 & -0.4\% & 4.8\%
& -0.08 & -0.4\% & 4.6\%
\\ 

15  
& -0.87 & -3.7\% & 4.3\%
& -0.96 & -3.5\% & 3.7\%
\\ 
\bottomrule
\end{tabular}}
}
    \bnotetab{This table shows the out-of-sample annualized Sharpe ratio (SR), mean return ($\mu$), and volatility ($\sigma$) for the CNN+Transformer model with trading frictions on PCA residuals. We use the daily out-of-sample residuals from January 1998 to December 2016 and evaluate the out-of-sample arbitrage trading from January 2002 to December 2016. The models are calibrated on a rolling window of four years and use either the Sharpe ratio or mean-variance objective function with trading costs $\text{cost}(\wret_{t-1},\wret_{t-2}) = 0.0005 \|\wret_{t-1} - \wret_{t-2}\|_{L^1} + 0.0001 \|\min(\wret_{t-1},0)\|_{L^1}$. The signals are extracted from a rolling window of $L=30$ days.} 
\end{table}

\subsection{Portfolio Concentration}

     \begin{figure}[H]
            \tcapfig{Industry Concentration of Portfolio Weights}
  \label{fig:industry}
         \centering
    \includegraphics[trim={0 1.5cm 0 2.3cm},clip,width=1\linewidth]{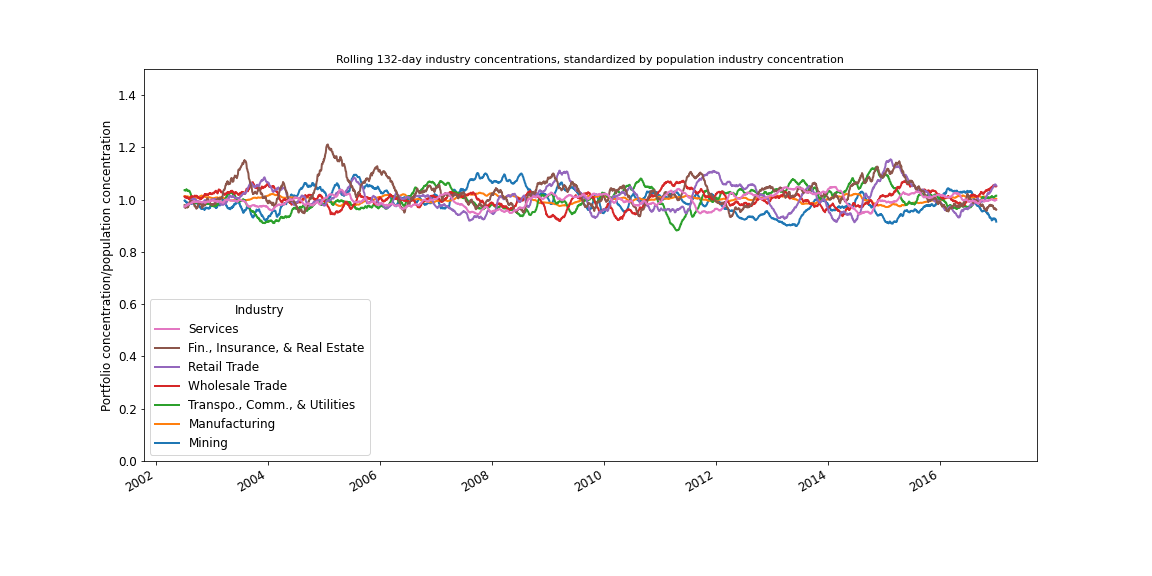}
         \bnotefig{This figure shows the rolling 132-day industry concentration of portfolio weights standardized by the population industry concentration. The stock portfolio weights $\wret_t$ are for the empirical benchmark CNN+Transformer model based on IPCA 5-factor residuals and for the out-of-sample trading period between January 2002 and December 2016. We use standard SIC industry classifications.
 }
     \end{figure}

\end{document}